\pdfoutput = 1
\documentclass[twocolumn]{article}
\usepackage[numbers]{natbib}
\usepackage{balance}
\usepackage{style}
\usepackage{gensymb}
\usepackage{mhchem}
\usepackage{xurl}
\usepackage{times}
\usepackage[english]{babel}
\usepackage{blindtext}
\usepackage{mathtools}
\usepackage{graphicx}
\usepackage{amsmath, amsthm, amssymb, bbm, bm}
\usepackage{enumerate}
\usepackage{float}      
\usepackage{subcaption}  
\usepackage{wrapfig}
\usepackage{algpseudocode}
\usepackage[margin=0cm]{caption}
\usepackage{tabularx}
\usepackage{multirow}
\usepackage{hhline}
\usepackage{makecell}
\usepackage{placeins}  
\usepackage[x11names, usenames, dvipsnames, svgnames, table]{xcolor}
\definecolor{firebrick}{rgb}{.698,.133,.133}
\definecolor{mybluelight}{rgb}{0.9, 0.9, 1.}
\definecolor{myorangelight}{rgb}{1., 0.9, 0.9}
\usepackage[utf8]{inputenc} 
\usepackage[T1]{fontenc}    
\usepackage{url}            
\usepackage{booktabs, colortbl}       
\usepackage{amsfonts}       
\usepackage{nicefrac}       
\usepackage{microtype}      
\usepackage{csquotes}
\usepackage{latexsym}
\usepackage{pifont}
\usepackage{tocbasic}

\let\oldnl\nl
\newcommand{\nonl}{\renewcommand{\nl}{\let\nl\oldnl}}

\usepackage{paralist}

\usepackage{xspace}
\usepackage{soul}
\usepackage{dsfont}
\usepackage{stmaryrd}
\usepackage[textwidth=15mm]{todonotes}
\usepackage{dirtytalk}
\usepackage{pbox}
\usepackage{cprotect}
\usepackage{bm}

\usepackage{verbatim}
\usepackage{textcomp}
\usepackage[normalem]{ulem}
\usepackage{etextools}
\usepackage[inline]{enumitem}

\usepackage[colorlinks=true,allcolors=firebrick,bookmarks=false]{hyperref}

\usepackage[toc,page,header]{appendix}
\usepackage{minitoc}

\renewcommand \thepart{}

\definecolor{darkergreen}{RGB}{21, 152, 56}
\definecolor{red2}{RGB}{252, 54, 65}
\definecolor{Gray}{gray}{0.85}
\newcolumntype{g}{>{\columncolor{Gray}}c}

\usepackage{xr-hyper}
\usepackage{algorithm}
\usepackage{algpseudocode}
\usepackage{tikz}
\usepackage{pgfplots}
\usepackage{pgfplotstable}

\let\OLDthebibliography\thebibliography
\renewcommand\thebibliography[1]{
  \OLDthebibliography{#1}
  \setlength{\parskip}{0pt}
  \setlength{\itemsep}{0pt plus 0.3ex}
}

\newcommand\sfdadep{\texttt{SFDA-SPF}\xspace}
\newcommand\sfdade{\texttt{SFDA-DE}\xspace}
\newcommand\cdcl{\texttt{CDCL}\xspace}
\newcommand\erl{\texttt{ERL}\xspace}
\newcommand\rgv{\texttt{RGV}\xspace}
\newcommand\shot{\texttt{SHOT}\xspace}

\newcommand\pixelcam{\texttt{PixelCAM}\xspace}
\newcommand\sat{\texttt{SAT}\xspace}
\newcommand\deepmil{\texttt{DeepMIL}\xspace}
\newcommand\negev{\texttt{NEGEV}\xspace}

\newcommand\pxap{\texttt{PxAP}\xspace}

\newcommand\cl{\texttt{CL}\xspace}

\newcommand\glas{\texttt{GlaS}\xspace}

\newcommand\camsixteen{\texttt{CAMELYON16}\xspace}
\newcommand\camseventeen{\texttt{CAMELYON17}\xspace}

\newcommand\camseventeenthree{\texttt{CAMELYON17-3}\xspace}
\newcommand\camseventeenfour{\texttt{CAMELYON17-4}\xspace}
\newcommand\camsix{\texttt{C16}\xspace}
\newcommand\camseven{\texttt{C17}\xspace}
\newcommand\camsevenzero{\texttt{C17-0}\xspace}
\newcommand\camsevenone{\texttt{C17-1}\xspace}
\newcommand\camseventwo{\texttt{C17-2}\xspace}
\newcommand\camseventhree{\texttt{C17-3}\xspace}
\newcommand\camsevenfour{\texttt{C17-4}\xspace}

\newcommand\ebhi{\texttt{EBHI}\xspace}
\newcommand\ebhisource{\texttt{EBHI-S}\xspace}
\newcommand\ebhitarget{\texttt{EBHI-T}\xspace}

\newcommand{\second}[1]{\underline{#1}}

\newtheorem{theorem}{Theorem}
\newtheorem{proposition}[theorem]{Proposition}

\newtheorem{corollary}[theorem]{Corollary}

\theoremstyle{definition}

\DeclarePairedDelimiterX{\divx}[2]{(}{)}{%
  #1\;\delimsize\|\;#2%
}

\DeclarePairedDelimiter{\abs}{\lvert}{\rvert}

\makeatletter
\@namedef{ver@everyshi.sty}{}
\newcommand{\removelatexerror}{\let\@latex@error\@gobble}

\newlength{\NFwidth}
\NewDocumentCommand{\NFline}{O{l}m}{\footnotesize\makebox[\NFwidth][#1]{#2}}

\NewDocumentCommand{\NFentry}{sm}{%
  \makebox[.5\NFwidth][l]{\normalsize
    \IfBooleanT{#1}{\makebox[0pt][r]{\textbullet\ }}%
    #2}\ignorespaces}

\title{Selective Prediction Forgetting for Adaptation of Weakly Supervised Localization of Histology Images}

\renewcommand\footnotemark{}
\author{Alexis~Guichemerre${^{1}}$, \quad 
\textbf{Banafsheh Karimian}${^{1}}$, \quad
\textbf{Soufiane~Belharbi}${^1}$, \quad
\textbf{Natacha~Gillet}${^1}$, \quad
\textbf{Nicolas Thome}${^{3}}$, \quad \\
\textbf{Pourya Shamsolmoali}${^{1}}$, 
\textbf{Mohammadhadi Shateri}${^1}$, \quad
\textbf{Luke McCaffrey}${^2}$, \quad
\textbf{Eric~Granger}${^1}$\vspace{0.05in}\\
$^1$LIVIA, Dept. of Systems Engineering, ETS Montreal, Canada \vspace{0.02in}\\
$^2$Goodman Cancer Research Centre, Dept. of Oncology, McGill University, Montreal, Canada
\vspace{0.02in}\\
$^3$Sorbonne Université, Paris, France\\
$^4$University of York, York, United Kingdom\\
\vspace{0.02in}\\
{\tt\scriptsize\{soufiane.belharbi, mohammadhadi.shateri, eric.granger\}@etsmtl.ca
}
\\
{\tt\scriptsize \{alexis.guichemerre.1\}@ens.etsmtl.ca
}
\\
}

\newcommand{\ignore}[1]{}

\begin{document}
\doparttoc 
\faketableofcontents 

\thepart{} 
\parttoc 

\maketitle\thispagestyle{fancy}
\maketitle
\rhead{\color{gray} \small Guichemerre et al. \;  [Unde Review 2026]}

\begin{abstract}
Under strong cross-domain shifts, weakly supervised object localization (WSOL) models often lose their capacity to discriminate and develop a strong prediction bias toward a subset of classes. As a result, the majority of target samples are assigned to these dominant classes.  Since source-free (unsupervised) domain adaptation (SFDA) methods typically rely on pseudo-labels, inaccurate pseudo-label estimation causes suboptimal adaptation and poor classification and localization performance.
This issue is particularly challenging in computational histopathology, where 
shifts in organ-specific morphology, staining protocols, and acquisition conditions can significantly impact class-discriminative visual cues. 
This paper introduces SFDA with selective prediction forgetting (\sfdadep), a method designed to mitigate excessive concentration of predictions for dominant classes during target-domain adaptation. \sfdadep identifies prediction-dominant classes based on the model's target predictions and selectively revises uncertain samples assigned to them. Importantly, this procedure does not require knowledge of the true target class distribution.  In contrast with previous SFDA methods, \sfdadep introduces a “forget set” composed of uncertain samples from dominant classes near their decision boundaries, enabling the classification boundaries to shift away from biased predictions. A second “retain set” is used to anchor certain and accurate samples. To forget the predictions of dominant classes, \sfdadep introduces a complementary optimization objective that reduces their probability while allowing alternative class assignments to emerge. This iterative optimization process reassesses target images for forget and retain sets at regular intervals. In addition to improving image-level pseudo-label estimation, the pixel-level classifier is jointly optimized, further enhancing localization.  
Experiments on cross-organ and cross-center histopathology benchmarks (\glas, \camsixteen, and \camseventeen) show that \sfdadep consistently outperforms state-of-the-art SFDA methods in both classification and localization, including across different levels of target-domain class imbalance.
\end{abstract}

\section{Introduction}  \label{sec:intro} 

Computer-aided diagnostic systems based on deep learning are increasingly explored in digital pathology to support large-scale cancer screening and grading~\citep{komura2018machine,gurcan2009histopathological,rony23,pixelcam,clipit}. However, most state-of-the-art approaches rely on densely annotated datasets, where pixel-level annotations of tumor regions remain labor-intensive and require expert pathologists~\citep{rony23, sfdawsol, pixelcam}. Deep WSOL models address this limitation by jointly predicting image-level diagnoses and localizing discriminative tissue regions using only image-level class labels for training~\citep{cam,dips, rony23,pixelcam}.

\begin{figure}[!t]
    \centering
    \includegraphics[width=0.99\linewidth]{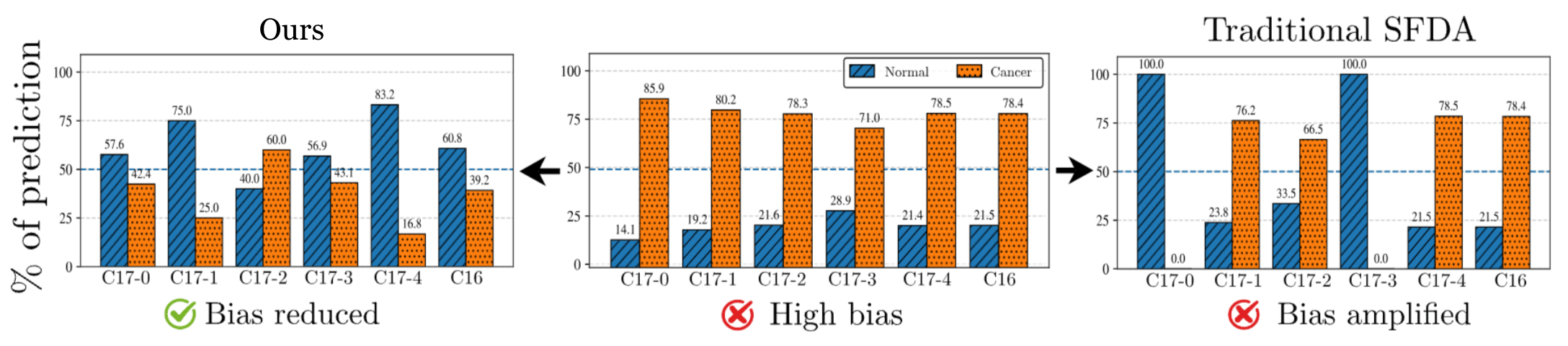}
    \caption{
    (a) \sfdadep reduces predictive bias across centers.
    (b) Source-only inference shows a strong predictive bias toward the
    cancer class when transferring from \glas (source) to \camsixteen
    and \camseventeen (targets) with PixelCAM.
    (c) In contrast, traditional SFDA methods, such as \sfdade, amplify the bias and often collapse toward a dominant class.
    }
    \label{fig:performance}
\end{figure}

Despite this advantage, the performance of WSOL models degrades under domain shifts, which are common in histopathology due to variations in staining protocols, scanner characteristics, tissue processing, inter-center acquisition, and organ-specific tissue architecture~\citep{sfdawsol,pixelcam,chensgcd}. Empirical analysis (Fig.~\ref{fig:performance}) shows that WSOL performance drops substantially under cross-organ shifts and dataset-scale imbalance~\citep{sfdawsol,pixelcam}. When a WSOL model trained on a small source dataset (possibly from a different organ) is deployed on a large target dataset collected at another center, both classification accuracy and spatial localization quality degrade~\citep{sfdawsol}. In contrast, models trained on large datasets generalize more consistently across domains. 
This transfer pattern suggests that WSOL degradation under severe shifts can be influenced by differences in organ-specific tissue structure and dataset size between the source and target domains. Additionally, appearance variations remain a major source of domain shift. In digital pathology, spatial drift makes predictions less reliable and harder to interpret~\citep{sfdawsol}.

SFDA provides a practical framework that adapts a source-trained model using only unlabeled target data~\citep{fg_masoumeh,wacv_zeeshan}, relying mainly on pseudo-labels generated from source-model predictions~\citep{liang2021shot,ding2022sourcefree,sfdawsol}. Under strong target shift and the high visual inter-class similarity, WSOL models can lose their discriminative power, leading them to assign most samples to the same dominant classes~\citep{sfdawsol,pixelcam}. Since standard SFDA methods do not explicitly detect and correct this prediction bias~\citep{fang2024source}, the resulting skewed pseudo-label distribution tends to reinforce dominant-class over-prediction during adaptation~\citep{sfdawsol}, leading to poor performance.  A common approach to reduce bias in SFDA is to diversify target predictions~\citep{liang2021shot, yang2021exploiting, yang2022attracting}. Nevertheless, this mechanism is applied globally across all classes and samples, without identifying the individual samples responsible for the dominant classes, which limits its effectiveness.
In a WSOL setting for histopathology data, these limitations of SFDA affect both classification and localization, as localization is derived from class-discriminative evidence learned through image-level classification supervision~\citep{rony23, sfdawsol, pixelcam}.

To overcome these limitations, we propose a novel SFDA method, \sfdadep, that identifies regions of the target prediction space where erroneous assignments are more likely to occur.
Under severe domain shift, prediction errors can accumulate in classes that receive a disproportionately large share of target samples. \sfdadep first identifies these prediction-dominant classes and then uses predictive uncertainty to find the samples within them that are most likely to be unreliable. These samples form a \emph{forget set}, where the probability assigned to their current predicted class is reduced, allowing alternative class assignments to emerge. The remaining samples from all classes form a \emph{retain set}, which preserves stable target predictions and limits excessive model drift.

\noindent \textbf{The key contributions of our work are:}

\textbf{(1)} We identify and empirically characterize dominant-class prediction bias as a critical obstacle to SFDA of WSOL models for histopathology. In this case, strong shifts in staining, acquisition, tissue preparation, and organ-specific morphology can obscure subtle class-discriminative cues, causing the source model to assign most target samples to a small subset of dominant classes.
\textbf{(2)} \sfdadep, a novel SFDA strategy that acts directly on the uncertain target samples that predict dominant classes. A \emph{forget set} identifies uncertain samples currently assigned to dominant classes, and a new loss reduces the probability assigned to their current dominant-class prediction, allowing alternative class predictions to emerge. In parallel, a \emph{retain set} preserves the remaining target predictions to anchor adaptation and limit excessive model drift. Both sets are periodically reconstructed as the target decision boundary evolves. A jointly optimized pixel-level classifier using CAM-derived partial pseudo-labels further improves model localization.
\textbf{(3)} An extensive evaluation of \sfdadep under cross-organ and cross-center shifts involving \glas, \camsixteen, and the five \camseventeen centers, using CNN- and ViT-based WSOL models. For the six target domains, \sfdadep achieves the highest average localization performance for all three methods, outperforming the strongest competing SFDA baseline by over 10 \pxap points on average. 

\begin{figure*}[!t]
    \centering
    \includegraphics[width=\textwidth]{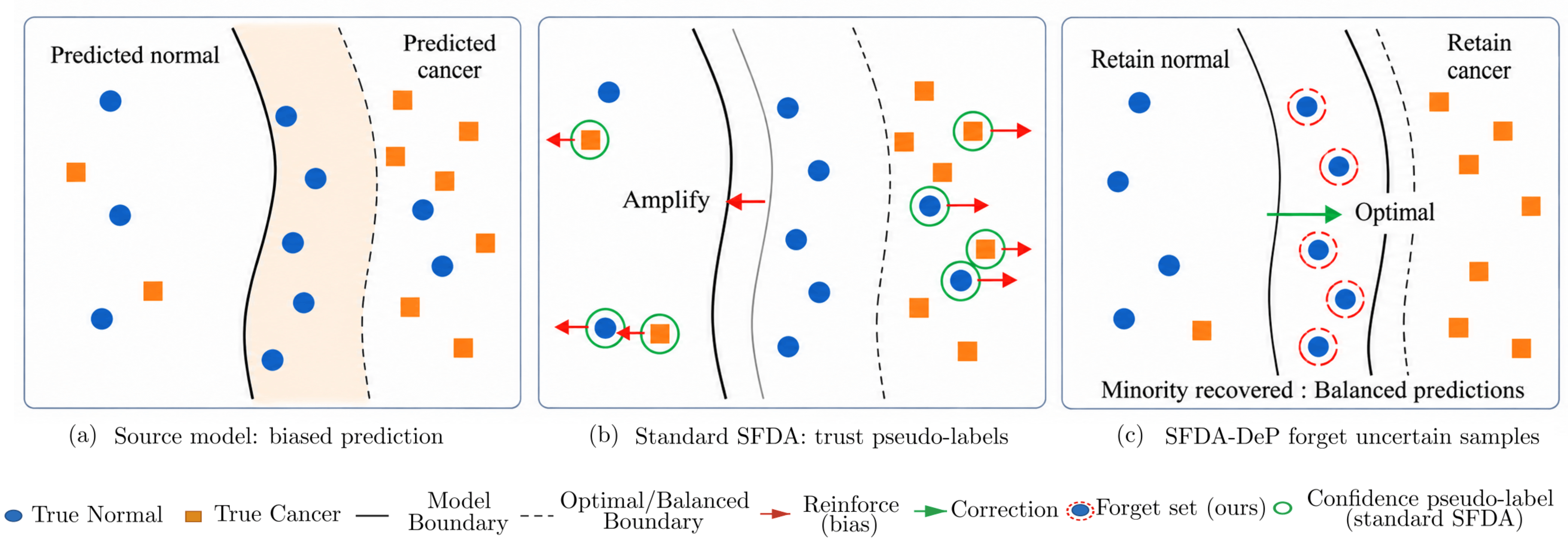}
      \caption{\textbf{Conceptual overview of \sfdadep.} In source-free adaptation for histological WSOL, the source model is biased toward the dominant class.
    \textbf{(a) Source model: biased prediction.} Many samples lie near the decision boundary, where high-entropy predictions are informative but error-prone.
    \textbf{(b) Standard SFDA: trust pseudo-labels.} Conventional methods reinforce confident pseudo-labels. However, in histology, confidence does not necessarily imply correctness, and erroneous pseudo-labels amplify the initial bias and attract uncertain samples toward the dominant class.
    \textbf{(c) \sfdadep forget uncertain samples (ours).} We instead identify uncertain dominant-class samples as a \emph{forget set}. The forgetting objective reduces the probability assigned to the current
    dominant-class prediction, allowing probability mass to be redistributed toward alternative classes.}
\label{fig:conceptual_overview}

\end{figure*}

\noindent\textbf{(a) Weakly Supervised Object Localization:} WSOL models predict spatial localization from image-level supervision through the class activation mapping (CAM) mechanism~\citep {cam, gradcam, gradcampp, layercam, fcam, tscam}, where localization maps are induced by classifier response. Representative WSOL approaches can be categorized into bottom-up (forward-pass information) and top-down strategies (exploit both forward and backward information)~\citep{zhangwsol, shaowsol, rony23}. Bottom-up approaches include \deepmil~\citep{deepmil}, which uses instance-level attention weights to identify diagnostically relevant patches. Transformer-based methods~\citep{chen2022lctr, gupta2022vitol} such as \sat~\citep{SAT} instead exploit interactions between image tokens to refine class-specific attention maps. In histopathology, methods such as \negev~\citep{negev} and \pixelcam~\citep{pixelcam} introduce additional spatial-refinement mechanisms to improve pixel-level attribution. 
Robustness of WSOL models under cross-domain conditions remains relatively underexplored in computational histopathology applications. Recent studies~\citep{sfdawsol, pixelcam} report significant degradation in both classification and localization when models are transferred across organs or centers. \\
\noindent\textbf{(b) Source-Free Domain Adaptation:} SFDA addresses domain shift~\citep{Li2020cvpr, kurmi2021domain, zhang2023source} by adapting a pretrained source model to a new target domain using only unlabeled target data, avoiding annotation costs and the need for source data. A major family of SFDA methods relies on using reliable target predictions and features~\citep{chen2021selfsupervised, ding2022proxymix, fang2024source, sfdawsol, li2024comprehensive}. For example, SHOT~\citep{liang2021shot} and CDCL~\citep{wang2022cross} refine target pseudo-labels using class centroids. SFDA-DE~\citep{ding2022sourcefree} instead estimates class-conditioned source feature distributions from high-confidence target pseudo-labels. Such strategies reduce the influence of noisy supervision, but their effectiveness strongly depends on the correctness of the selected pseudo-labels~\citep{sfdawsol}. Moreover, discarding less reliable samples limits the use of a substantial portion of samples from the target domain, including ambiguous samples located near the decision boundary.  To exploit a broader set of target samples, uncertainty-aware approaches assign sample-specific weights according to prediction confidence or entropy~\citep{lee2022confidence, roy2022uncertainty, litrico2023guiding, xu2025revisiting}. However, weighting modifies the impact of pseudo-labels without reconsidering the resulting class distribution. Under strongly concentrated target predictions, uncertain samples assigned to dominant classes may still influence adaptation, albeit with reduced weight. Consequently, these methods preserve or amplify bias rather than correct the underlying class assignment. Another common strategy is to encourage confident sample-level predictions and diverse class assignments~\citep{liang2021shot, yang2021exploiting, yang2022attracting}. However, those objectives act as global regularizers and do not identify which samples assigned to over-predicted classes should be reconsidered. \\ \\
\noindent\textbf{(c) Uncertainty in SFDA:} In the absence of source data and target labels, estimating the reliability
of target predictions is a central and key challenge in SFDA.
Several approaches exploit predictive or representation uncertainty to 
reduce the impact of unreliable images by reweighting target samples or prioritizing confident predictions during adaptation~\citep{roy2022uncertainty,lee2022confidence,litrico2023guiding}. These approaches use uncertainty to determine how strongly a target prediction should contribute to optimization. More recently,~\citep{xu2025revisiting} studied uncertainty in contrastive-learning-based SFDA from an instance-dependent perspective and proposed controlling prediction variance during adaptation. While this improves the
utilization of uncertain target data, the resulting supervision remains
derived from the original model's prediction history. Consequently, under severe domain shifts as in histopathology where predictions become skewed toward a 
dominant class, such supervision may still preserve candidate assignments induced by the biased source model.

Although these global regularizers can reduce excessively imbalanced predictions, they do not explicitly determine which classes are over-predicted or which samples assigned to these classes should be reconsidered.

As highlighted in~\citep{sfdawsol}, conventional SFDA methods face substantial difficulties when deployed with WSOL models for computational histopathology. Since localization maps are derived from class-specific activations, their quality is closely tied to the model's ability to correctly predict the image class. When SFDA reinforces erroneous pseudo-labels, the backbone is optimized using incorrect class-level signals, which degrades the spatial feature maps, leading to poor performance in both classification and localization tasks.

\section{SFDA with Selective Prediction Forgetting} \label{sec:unlearning_cls}
\begin{figure*}[t]
    \centering    \includegraphics[width=0.99\linewidth]{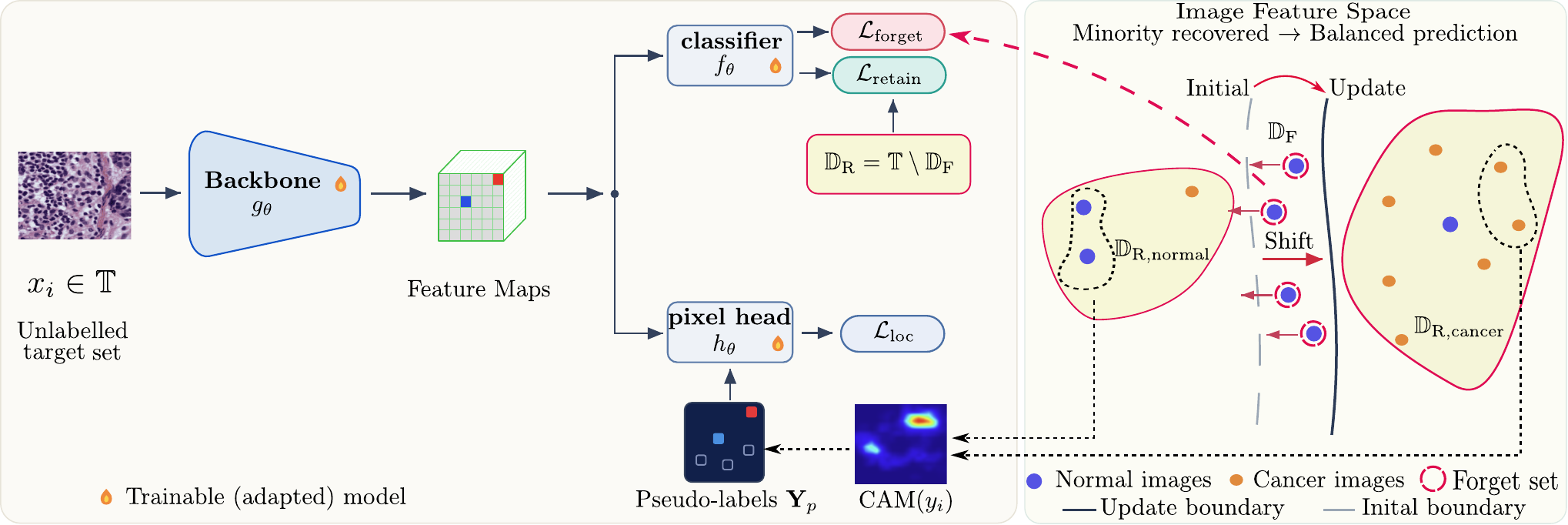}
\caption{\textbf{Overview of the proposed \sfdadep pipeline.}
Starting from a source-pretrained WSOL model and an unlabeled target set, \sfdadep first identifies dominant classes from the target prediction distribution. For each dominant class, the fraction $\rho$ of samples with the highest normalized entropy is assigned to the forget set, while the remaining target samples form the retain set. During adaptation, $\mathcal{L}_{\mathrm{F}}$ reduces the probability assigned to the current dominant-class predictions of forget samples, whereas $\mathcal{L}_{\mathrm{R}}$ preserves the current assignments of the remaining samples to limit excessive model drift. An auxiliary pixel-level loss, $\mathcal{L}_{\mathrm{L}}$, uses CAM-based partial pseudo-labels from the frozen source model to preserve spatial representations. Target predictions and the dominant, forget, and retain sets are periodically recomputed as the decision boundary evolves.}
\label{fig:unlearning_overview}
\end{figure*}
\noindent \textbf{Notation.}
Let us denote by ${\mathbb{T} = \{x_i\}_{i=1}^N}$ an unlabeled target training set composed of $N$ images ${x_i: \Omega \subset \mathbb{R}^2 }$, where ${\Omega}$ is a discrete image domain. We consider a classifier ${f_{\theta}(x) \in [0, 1]^K}$ that was pre-trained on source data ${\mathbb{S}=\{(x, y)_i\}_{i=1}^M}$ labeled only with image classes ${y \in \{1, \cdots, K\}}$, where $K$ is the total number of classes. The feature extractor of $f$ is denoted by ${g(x) = Z \in \mathbb{R}^{\abs{\Omega}\times d}}$, where $d$ is the feature dimension. We denote by ${z_s \in \mathbb{R}^d}$ the spatial embedding at location $s$ in ${\Omega}$. During adaptation, the source dataset $\mathbb{S}$ is no longer accessible. For each target image $x_i$, we denote its predicted class by
$\hat{y}_i=\arg\max_{k\in\{1,\ldots,K\}}p_{i,k}$. 

We first measure the empirical prediction distribution of the current
model. The fraction of target samples currently assigned to class $k$ is denoted by: 
\[
\widehat{\pi}(k)
=
\frac{1}{N}\sum_{i=1}^{N}\mathbf{1}[\hat{y}_i=k]
\]
Using $1/K$ as a class-symmetric reference, we define the set of
prediction-dominant classes as: $
\mathcal{C}_{\mathrm{dom}} = \left\{
k\in\{1,\ldots,K\}:
\widehat{\pi}(k)>\frac{1}{K}
\right\}$.
Since patch-level class proportions can naturally vary across histopathology cohorts, the uniform reference is not intended to model the underlying target class prevalence. Instead, $(1/K)$ provides a class-symmetric reference point for determining how the current model distributes its predictions across classes. A prediction frequency above this reference indicates that a class attracts a large fraction of the model predictions. Under domain shift, when the source model exhibits degraded target performance, such prediction concentration can cause prediction-dominant classes to accumulate an important fraction of the misclassified target samples. We therefore use prediction dominance as a dataset-level cue to identify where potentially erroneous assignments are likely to concentrate.
\sfdadep therefore combines this global signal with instance-level predictive uncertainty. Among samples assigned to prediction-dominant classes, only highly uncertain predictions are selected for forgetting, as these samples are more likely to lie near unreliable target decision boundaries, while more confident assignments are retained. In this way, the uniform reference identifies where prediction errors may accumulate, whereas uncertainty identifies which individual assignments should be explicitly reconsidered.

\paragraph{Forget and Retain Sets.}
\sfdadep identifies target samples whose current assignments may contribute to the observed prediction bias and progressively revises these assignments during adaptation. This process shifts the decision boundaries away from excessive dominant-class assignments, thereby reducing the skew in the target prediction distribution. 
To do this, we sample uncertain samples near the boundaries from dominant classes by using the entropy defined by:
\begin{equation}
H(x_i)
=
-\frac{1}{\log K}
\sum_{c=1}^{K}
p_{\theta}(y=c \mid x_i)
\log p_{\theta}(y=c \mid x_i).
\label{eq:normalized_entropy}
\end{equation}
For each dominant class $k \in \mathcal{C}_{\mathrm{dom}}$, we define its class-specific candidate set:
\begin{equation}
\mathbb{B}_{k}
=
\left\{
(x_i,\hat{y}_i)
\;\middle|\;
x_i \in \mathbb{T},
\ \hat{y}_i = k
\right\},
\qquad
\forall k \in \mathcal{C}_{\mathrm{dom}}.
\label{eq:class_candidate_set}
\end{equation}
Within each $\mathbb{B}_{k}$, we select the top $\rho$\% most uncertain samples, which are the closest to the boundary: 
\begin{equation} 
\mathbb{D}_{\mathrm{F}}^{k} = \operatorname{Top}_{\rho} \left( \mathbb{B}_{k}, H \right). 
\label{eq:class_forget_set} 
\end{equation} 
The complete forget set, $\mathbb{D}_{\mathrm{F}}$, is obtained by aggregating the selected samples from all dominant classes: \begin{equation} \mathbb{D}_{\mathrm{F}} = \bigcup_{k\in\mathcal{C}_{\mathrm{dom}}} \mathbb{D}_{\mathrm{F}}^{k}. \label{eq:forget_set} \end{equation}

To prevent excessive model drift and preserve the overall target prediction structure, we define the retain set, $\mathbb{D}_{\mathrm{R}}$, as the complement of the forget set:  
\begin{equation}
\mathbb{D}_{\mathrm{R}}
=
\left\{
(x_i,\hat{y}_i)
\;\middle|\;
x_i \in \mathbb{T},
\ (x_i,\hat{y}_i)
\notin
\mathbb{D}_{\mathrm{F}}
\right\}.
\label{eq:retain_set}
\end{equation}
Importantly, it contains samples from all classes, including confident samples assigned to dominant classes, which helps maintain the model’s ability to predict every class.

\paragraph{Forget and Retain Losses.}
For samples selected in the forget set, we explicitly reduce the probability assigned to their current predicted class. Since these samples are uncertain and currently assigned to prediction-dominant classes, preserving their pseudo-labels may reinforce potentially erroneous target assignments. We therefore define:
\begin{equation}
\mathcal{L}_{\mathrm{F}}
=
-\frac{1}{|\mathbb{D}_{\mathrm{F}}|}
\sum_{x_i\in\mathbb{D}_{\mathrm{F}}}
\log
\left(
1-p_{\theta}(y=\hat{y}_i\mid x_i)
\right),
\label{eq:forget_loss}
\end{equation}
where $\hat{y}_i$ denotes the current prediction of $x_i$.
By minimizing this objective over the forget set, we discourage the model
from preserving these uncertain dominant-class assignments, thereby allowing
alternative class predictions to emerge as the target decision boundary evolves. For samples in $\mathbb{D}_{\mathrm{R}}$, we seek to preserve their current pseudo-labels $\hat{y}_i$ assigned by the model using a standard cross-entropy objective:
\begin{equation}
\mathcal{L}_{\mathrm{R}}
=
-\frac{1}{|\mathbb{D}_{\mathrm{R}}|}
\sum_{x_i\in\mathbb{D}_{\mathrm{R}}}
\log
p_{\theta}(y=\hat{y}_i\mid x_i).
\label{eq:retain_loss}
\end{equation}
This objective maintains the prediction structure of samples that were not
selected for the forgetting process. This loss complements
$\mathcal{L}_{\mathrm{F}}$ by preserving the remaining class assignments
while the decision boundaries are adjusted around ambiguous samples from
dominant classes.

Our \emph{forget-retain} approach progressively reduces the skew in the target
class-prediction distribution. This gradually adjusts the decision boundaries away from dominant classes. As adaptation proceeds, some samples initially included in $\mathbb{D}_{\mathrm{F}}$ no longer provide relevant corrective signals once their assignments have changed or moved away from the updated decision boundary. Conversely, other samples become ambiguous with respect to the evolving boundary while remaining assigned to a dominant class. We periodically reassess the prediction bias and reconstruct both sets every
$m$ epochs. At each reconstruction step, we recompute the empirical target prediction distribution $\widehat{\pi}$, update the set of dominant classes
$\mathcal{C}_{\mathrm{dom}}$, and assign the currently most ambiguous samples
predicted as these classes to $\mathbb{D}_{\mathrm{F}}$. The remaining
target samples are assigned to $\mathbb{D}_{\mathrm{retain}}$.

\noindent\textbf{Localization Task:} In WSOL, localization is obtained by class activation maps conditioned on the predicted image-level class. Therefore, improving the accuracy of target pseudo-labels during adaptation provides more reliable class-specific training signals and encourages the model to focus on regions associated with the correct semantic class. As our forget/retain method improves image-level predictions, they can now be used to provide more reliable class-specific localization cues during adaptation. To further preserve discriminative local representations, we follow PixelCAM~\citep{pixelcam} and use an auxiliary pixel-level classifier $h:\mathbb{R}^{d}\rightarrow[0,1]^2$, which predicts whether a spatial embedding $z_s$ belongs to the foreground or background. The locations of the known pseudo-labels are encoded in ${\Omega^{\prime}}$. To this end, for each target sample, we define a mask of pixel-wise partial pseudo-labels denoted ${\bm{Y} \in \{0, 1\}^{\abs{\Omega}\times 2}}$, ${\bm{Y}_s \in \{0, 1\}^2}$ at location $s$ with value ${0}$ for background, ${1}$ for foreground. To reduce overfitting to potentially noisy pixel-level pseudo-labels, only a subset of pixels is randomly sampled from each image at every training iteration. This stochastic sampling encourages the model to explore different spatial regions. For each $x_i\in\mathbb{D}_{\mathrm{loc}}$, the current image-level pseudo-label $\hat{y}_i$ determines the class used for localization. We then extract the CAM associated with $\hat{y}_i$ from the frozen source-pretrained model and convert it into partial foreground and background pseudo-labels following PixelCAM~\citep{pixelcam}. The pixel-level classifier is trained using those pseudo-labels with standard binary cross-entropy loss.

\begin{equation}
\label{eq:pl}
\mathcal{L}_{\mathrm{loc}}
=
-\frac{1}{|\Omega^{\prime}|}
\sum_{s \in \Omega^{\prime}}
\left[
(1-\bm{Y}_s)\log(h(z_s)_0)
+
\bm{Y}_s\log(h(z_s)_1)
\right].
\end{equation}

The total adaptation loss consists of three jointly optimized terms. The first two terms reduce class prediction bias while preserving reliable predictions. The third improves localization by exploring the image and expanding to fully cover ROIs. The loss can be written as
\begin{equation}
\mathcal{L}
=
\lambda_{\text{R}} \mathcal{L}_{\text{R}}
+
\lambda_{\text{F}} \mathcal{L}_{\text{F}}
+
\lambda_{\text{L}} \mathcal{L}_{\text{loc}},
\end{equation}
\noindent where $\lambda_{\text{R}}$, $\lambda_{\text{F}}$ and $\lambda_{\text{L}}$ are weighting coefficients.

\section{Results and Discussion}
 
\subsection{Experimental Methodology}
\setlength{\tabcolsep}{3pt}
\renewcommand{\arraystretch}{1.0}

\begin{table*}[h]
\caption{\textbf{Localization (\pxap) and classification (\cl) accuracy.}
Results shown for \glas → \camsix, \camsevenzero, \camsevenone, \camseventwo, \camseventhree, \camsevenfour.}
\centering
\resizebox{0.9\textwidth}{!}{
\small
\begin{tabular}{|l c c 
| cc | cc | cc | cc | cc | cc || cc|}
\cline{1-17}
 & & &
\multicolumn{2}{c}{\camsix} &
\multicolumn{2}{c}{\camsevenzero} &
\multicolumn{2}{c}{\camsevenone} &
\multicolumn{2}{c}{\camseventwo} &
\multicolumn{2}{c}{\camseventhree} &
\multicolumn{2}{c}{\camsevenfour} &
\multicolumn{2}{c|}{\textbf{Avg}} \\

\cline{4-17}

 & \textbf{Method} &
 & \pxap & \cl
 & \pxap & \cl
 & \pxap & \cl
 & \pxap & \cl
 & \pxap & \cl
 & \pxap & \cl
 & \pxap & \cl \\

\hline \\[-1em]


\multirow{7}{*}{
\rotatebox{90}{
\begin{tabular}{c}
PixelCAM \\
\scriptsize (MIDL'25)
\end{tabular}
}}

& Source only 
& & 37.9 & 51.8 
& 37.2 & 41.2
& 27.2 & 58.7 
& 44.9 & 64.9
& 43.1 & 39.1
& 31.1 & 39.9 
& 36.9 & 49.3 \\

& SHOT {\scriptsize (ICML'20)}
& & 14.9 & \second{67.1} 
& 12.1 &  \second{71.7} 
& 19.2 & \textbf{70.9}  
&  24.4 & \second{71.6} 
& 15.1 & \second{78.9}  
& 13.9 & \textbf{56.4} 
& 16.6 & \textbf{69.4}  \\

& SFDA-DE {\scriptsize (CVPR'22)}
& & \second{37.9} & 51.8 
& 14.5 & 50.0 
& \textbf{36.7} & 65.7
& 28.9 & 70.5
& 18.9 & 49.9
& 31.1 & 39.9 
& 28.0 & 54.6 \\

& CDCL {\scriptsize (TMM'22)}
& & \second{37.9} & 51.8 
& 25.3 & 50.0 
& 21.2 & 59.3 
& 25.5 & 69.9 
& 19.3 & 49.9
& 22.6 & 40.9 
& 25.3 & 53.6 \\

& ERL {\scriptsize (ICLR'23)}
& & 29.9 & 59.7
& 13.2 & 59.5
& 14.6 & \second{67.4} 
& 43.0 & 69.4 
& 19.3 & 49.9
& \second{32.4} & \second{54.0} 
& 25.4 & \second{60.0} \\

& RGV {\scriptsize (CVPR'25)}
& & 30.7 & 52.9 
& \second{31.0} & 50.0 
& \second{29.4} & 53.5 
& \second{44.9} & 64.9 
& \second{40.2} & \second{49.9} 
& 32.2 & 41.6 
& \second{34.7} & 52.1 \\

& \cellcolor{lightgray}\textbf{Ours}
& \cellcolor{lightgray}
& \cellcolor{lightgray}\textbf{49.5} & \cellcolor{lightgray}\textbf{67.2}
& \cellcolor{lightgray}\textbf{49.9} & \cellcolor{lightgray}\textbf{86.2}
& \cellcolor{lightgray}27.2 & \cellcolor{lightgray}41.3
& \cellcolor{lightgray}\textbf{50.2} & \cellcolor{lightgray}\textbf{76.1}
& \cellcolor{lightgray}\textbf{53.8} & \cellcolor{lightgray}\textbf{80.6}
& \cellcolor{lightgray}\textbf{33.9} & \cellcolor{lightgray}51.3
& \cellcolor{lightgray}\textbf{44.1} & \cellcolor{lightgray}\second{67.1} \\

\hline \\[-1em]

\multirow{7}{*}{
\rotatebox{90}{
\begin{tabular}{c}
SAT \\
\scriptsize (ICCV'23)
\end{tabular}
}}

& Source only 
& & 24.9 & 54.0 
& 16.8 & 54.9
& 17.4 & 50.6 
& 28.8 & 52.5
& 21.0 & 53.2
& 18.9 & 47.3 
& 21.3 & 52.1 \\

& SHOT {\scriptsize (ICML'20)}
& & 27.2 & 65.1 
& \second{16.9} & 68.7 
& \second{18.9} & 50.0  
& \second{33.7} & \textbf{73.6} 
& 19.7 & 76.3  
& 17.9 & 55.0 
& \second{22.4} & 64.8  \\

& SFDA-DE {\scriptsize (CVPR'22)}
& & 26.8 & 62.9 
& 14.9 & \textbf{76.3} 
& 17.8 & \textbf{61.0} 
& 32.8 & 69.9 
& 14.1 & 76.3 
& \textbf{22.9} & \second{65.8} 
& 21.6 & 68.7 \\

& CDCL {\scriptsize (TMM'22)}
& & 32.3 & \second{66.6} 
& 14.4 & 72.5 
& 18.3 & 56.4 
& 33.5 & 72.8
& 16.1 & \second{77.4} 
& 17.4 & 55.4 
& 22.0 & 66.9 \\

& ERL {\scriptsize (ICLR'23)}
& & \second{32.6} & \second{66.6} 
& 14.4 & 69.8 
& 15.1 & 55.2 
& 32.2 & \second{73.2} 
& \second{22.4} & \textbf{79.3} 
& 16.3 & \textbf{69.1} 
& 22.2 & \second{68.9} \\

& RGV {\scriptsize (CVPR'25)}
& & 24.9 & 54.0 
& 16.8 & 54.9 
& 17.4 & 50.6 
& 28.0 & 52.4 
& 20.9 & 49.9 
& 18.9 & 48.3 
& 21.2 & 51.7 \\

& \cellcolor{lightgray}\textbf{Ours} 
& \cellcolor{lightgray} 
& \cellcolor{lightgray}\textbf{36.4} & \cellcolor{lightgray}\textbf{66.8} 
& \cellcolor{lightgray}\textbf{24.9} & \cellcolor{lightgray}\second{74.8} 
& \cellcolor{lightgray}\textbf{26.9} & \cellcolor{lightgray}\second{57.6} 
& \cellcolor{lightgray}\textbf{43.7} & \cellcolor{lightgray}\second{73.2} 
& \cellcolor{lightgray}\textbf{27.8} & \cellcolor{lightgray}\second{77.4} 
& \cellcolor{lightgray}\second{22.3} & \cellcolor{lightgray}65.4 
& \cellcolor{lightgray}\textbf{30.3} & \cellcolor{lightgray}\textbf{69.2} \\

\hline \\[-1em]

\multirow{7}{*}{
\rotatebox{90}{
\begin{tabular}{c}
DeepMIL \\
\scriptsize (ICML'18)
\end{tabular}
}}

& Source only 
& & 24.1 & 54.0 
& 14.1 & 45.0
& 18.4 & 56.4 
& 30.5 & 61.8
& 18.1 & 38.8
& 19.1 & 42.6 
& 20.7 & 49.8 \\

& SHOT {\scriptsize (ICML'20)}
& & 25.2 & \second{61.4} 
& 12.8 & \second{70.6}
& 17.3 & \textbf{68.0}  
& 27.5 &  \second{74.1} 
& 19.2 & \second{75.0}  
& 14.8 & \second{67.8} 
& 19.5 & \second{69.5}  \\

& SFDA-DE {\scriptsize (CVPR'22)}
& & 21.6 & 51.9 
& 14.5 & 50.0
& \second{18.4} & 56.4 
& \second{32.5} &72.5
& 16.6 & 49.9 
& 19.1 & 42.6 
& 20.5 & 53.9 \\

& CDCL {\scriptsize (TMM'22)}
& & 24.0 & 54.1 
& \second{27.2} & 50.0 
& \second{18.4} & 56.4 
& 27.8 & 72.3 
& \second{37.6} & 49.9 
& \second{28.6} & 50.0
& \second{27.3} & 55.5 \\

& ERL {\scriptsize (ICLR'23)}
& & \second{25.7} & 53.8 
& 12.1 & 50.0 
& 10.9 & 65.1 
& 19.4 & 71.8 
& 16.1 & 59.3 
& 12.9 & 46.6 
& 16.2 & 57.8 \\

& RGV {\scriptsize (CVPR'25)}
& & 24.7 & 52.9 
& 27.1 & 50.0
& \second{18.4} & \second{56.9} 
& 31.1 & 62.1 
& 21.0 & 49.9 
& 21.7 & 43.9 
& 24.0 & 52.6 \\

& \cellcolor{lightgray}\textbf{Ours}
& \cellcolor{lightgray}
& \cellcolor{lightgray}\textbf{53.5} & \cellcolor{lightgray}\textbf{72.1}
& \cellcolor{lightgray}\textbf{34.0} & \cellcolor{lightgray}\textbf{82.8}
& \cellcolor{lightgray}\textbf{38.8} & \cellcolor{lightgray}54.1
& \cellcolor{lightgray}\textbf{45.4} & \cellcolor{lightgray}\textbf{79.2}
& \cellcolor{lightgray}\textbf{37.7} & \cellcolor{lightgray}\textbf{77.9}
& \cellcolor{lightgray}\textbf{34.5} & \cellcolor{lightgray}\textbf{74.5}
& \cellcolor{lightgray}\textbf{40.7} & \cellcolor{lightgray}\textbf{73.4} \\

\hline
\end{tabular}
}
\label{tab:all-second-values}
\end{table*}

\setlength{\tabcolsep}{3pt}
\renewcommand{\arraystretch}{1.0}

\begin{table*}[h]
\setlength{\tabcolsep}{2.5pt}
\renewcommand{\arraystretch}{0.9}

\caption{\textbf{Localization (\pxap) and classification (\cl) accuracy.}
Inter-center SFDA on \camseventeen. The source model is trained on center 0 and independently adapted to centers 1--4.}
\centering

\resizebox{0.8\textwidth}{!}{
\scriptsize
\begin{tabular}{|l c c
| cc | cc | cc | cc |}

\cline{1-11}

& & &
\multicolumn{2}{c}{\camsevenone} &
\multicolumn{2}{c}{\camseventwo} &
\multicolumn{2}{c}{\camseventhree} &
\multicolumn{2}{c|}{\camsevenfour}  \\

\cline{4-11}

& \textbf{Methods} &
& \pxap & \cl
& \pxap & \cl
& \pxap & \cl
& \pxap & \cl \\

\hline \\[-1em]


\multirow{6}{*}{
\rotatebox{90}{
\begin{tabular}{c}
PixelCAM \\
\scriptsize (MIDL'25)
\end{tabular}
}}

& SHOT {\scriptsize (ICML'20)}
& & 22.9 & 52.9
& 31.6 & \second{77.0}
& 46.8 & 88.5
& \second{51.4} & \second{74.1}
\\

& SFDA-DE {\scriptsize (CVPR'22)}
& & 23.9 & 49.4
& 18.7 & 50.0
& 50.7 & \second{92.0}
& \second{51.4} & \second{74.1}
\\

& CDCL {\scriptsize (TMM'22)}
& & \second{40.1} & \second{73.8}
& 30.6 & 74.7
& \second{59.4} & 88.3
& 50.2 & \textbf{83.2}
\\

& ERL {\scriptsize (ICLR'23)}
& & 17.4 & 62.2
& \second{45.6} & 58.4
& 50.7 & \second{92.0}
& \second{51.4} & \second{74.1}
\\

& RGV {\scriptsize (CVPR'25)}
& & 36.1 & 56.9
& 41.2 & 50.00
& 50.7 & \second{92.0}
& \textbf{54.3} & 70.4
\\

& \cellcolor{lightgray}\textbf{Ours}
& \cellcolor{lightgray}
& \cellcolor{lightgray} \textbf{48.5} & \cellcolor{lightgray} \textbf{74.4}
& \cellcolor{lightgray} \textbf{58.1} & \cellcolor{lightgray} \textbf{81.6}
& \cellcolor{lightgray} \textbf{60.4} & \cellcolor{lightgray} \textbf{93.8}
& \cellcolor{lightgray} 46.1 & \cellcolor{lightgray} \textbf{83.2}
\\

\hline
\end{tabular}
}

\label{tab:c17_intercenter_src0}
\end{table*}

\noindent\textbf{Datasets:}
Our approach was evaluated on three publicly available histopathology benchmarks that enable both cross-organ and cross-center domain shifts. 
\glas provides gland-level annotations for colon cancer~\citep{sirinukunwattana2015stochastic, rony23}, while \camsixteen and \camseventeen focus on breast lymph-node metastasis detection~\citep{camelyon2016paper, c17, rony23}. We use \camseventeen~\citep{c17} and construct a WSOL-compliant patch-level benchmark that supports classification and localization evaluation with the provided tumor segmentation masks.
Following~\citep{rony23}, we perform a \emph{center-wise} WSI split that simulates cross-center shifts.
Each patch is labeled with an image-level class and a localization mask. Additional details are provided in the supplementary material.  
\newline
\noindent\textbf{Backbone and WSOL Models:} CNN-based models use ResNet-50~\citep{pixelcam} and transformer-based models use DeiT-Tiny~\citep{vit}. We evaluate three WSOL models: \deepmil~\citep{deepmil}, \sat~\citep{SAT}, \pixelcam~\citep{pixelcam}. For fair comparison, we follow~\citep{sfdawsol,pixelcam} for source pretraining. 
\noindent\textbf{\sfdadep Hyperparameters:}
The retain and forget loss weights are set as 
$\lambda_{\mathrm{R}}, \lambda_{\mathrm{F}} \in \{0.5, 1.0\}$. The localization loss weight is chosen from 
$\lambda_{\mathrm{L}} \in \{0.5, 1.0, 5.0\}$. The forgetting ratio is explored within  $\rho \in \{5\%, 15\%, 25\%\}$. Following prior SFDA protocols~\citep{sfdawsol,xu2025revisiting}, hyperparameters are selected on a separate validation set, Additional details are provided in Appendix~\ref{app:implementation}. \\
\noindent\textbf{Implementation Details:}

We compare our method against recent source-free domain adaptation (SFDA) approaches. 
We evaluate \shot~\citep{liang2021shot}, \sfdade~\citep{ding2022sourcefree}, \cdcl ~\citep{wang2022cross}, \erl~\citep{erl}, and \rgv~\citep{zhu2025revisiting}. Additional information regarding the hyper-parameters for each method is provided in Appendix~\ref{app:implementation}.

\begin{table*}[h]
\setlength{\tabcolsep}{2.5pt}
\renewcommand{\arraystretch}{0.9}

\caption{
\textbf{Robustness to class imbalance in the target adaptation set on \camseventeen.}
Starting from the balanced setting ($50{:}50$), we progressively increase the
proportion of Normal samples up to $90{:}10$. The test set remains balanced
across all settings; only the target training set is imbalanced.
}
\centering

\resizebox{0.8\textwidth}{!}{
\scriptsize
\begin{tabular}{|l c
| cc | cc | cc | cc | cc |}

\cline{1-12}

& &
\multicolumn{2}{c|}{$50{:}50$}
& \multicolumn{2}{c|}{$60{:}40$}
& \multicolumn{2}{c|}{$70{:}30$}
& \multicolumn{2}{c|}{$80{:}20$}
& \multicolumn{2}{c|}{$90{:}10$}
\\

\cline{3-12}

& \textbf{Methods}
& \pxap & \cl
& \pxap & \cl
& \pxap & \cl
& \pxap & \cl
& \pxap & \cl
\\

\hline


\multirow{5}{*}{
\rotatebox{90}{
\begin{tabular}{c}
PixelCAM \\
\scriptsize (MIDL'25)
\end{tabular}
}}

& \rule{0pt}{2.4ex}SHOT {\scriptsize (ICML'20)}
& 19.2 & 70.0
& 22.9 & 41.6
& 37.2 & 41.2
& 24.2 & 41.6
& 30.5 & 73.6
\\

& CDCL {\scriptsize (TMM'22)}
& 25.3 & 50.0
& 25.5 & 45.4
& 22.6 & 44.6
& 22.4 & 44.6
& 22.8 & 44.6
\\

& ERL {\scriptsize (ICLR'23)}
& 13.2 & 59.5
& 21.8 & 44.2
& 20.4 & 43.1
& 21.5 & 43.5
& 21.0 & 43.5
\\

& RGV {\scriptsize (CVPR'25)}
& 31.0 & 50.0
& 33.7 & 50.0
& 33.3 & 50.0
& 33.1 & 50.0
& 32.8 & 50.0
\\

& \cellcolor{lightgray}\rule[-1.0ex]{0pt}{2.8ex}\textbf{Ours}
& \cellcolor{lightgray}\textbf{49.9}
& \cellcolor{lightgray}\textbf{86.2}
& \cellcolor{lightgray}\textbf{43.7}
& \cellcolor{lightgray}\textbf{79.0}
& \cellcolor{lightgray}\textbf{55.1}
& \cellcolor{lightgray}\textbf{82.8}
& \cellcolor{lightgray}\textbf{46.0}
& \cellcolor{lightgray}\textbf{79.4}
& \cellcolor{lightgray}\textbf{58.8}
& \cellcolor{lightgray}\textbf{82.8}
\\

\hline
\end{tabular}
}

\label{tab:imbalance_results}
\end{table*}

\subsection{Impact on classification and localization}
Table~\ref{tab:all-second-values} reports localization (\pxap) and classification (\cl)
accuracy for the three WSOL methods across the cross-organ
(\glas$\rightarrow$ \camsix, \camseven ) and Table~\ref{tab:c17_intercenter_src0}--\ref{tab:imbalance_results} cross-center \camseven settings.
It shows that \sfdadep improves WSOL adaptation on most target domains and across
architectures. The largest gains appear on challenging domains where several SFDA baselines remain near $50\%$ \cl and show strong dominant-class prediction bias. On \pixelcam, \sfdadep reaches 44.1 \pxap and 67.1 \cl on average ($+16.1$~\pxap / $+12.5$~\cl over SFDA-DE), with the largest gains on the centers where the source model is most biased and the baselines remain around $50\%$~\cl. On \camsevenzero and \camseventhree, \cl rises from $\sim\!50\%$ to $86.2\%$ and $80.6\%$, while localization also improves, confirming that correcting the classification bias directly benefits the spatial cues that drive the CAMs. 
\begin{figure}[t]
    \centering
    \includegraphics[width=0.24\textwidth,
        trim=1cm 0 0 0,
        clip]
        {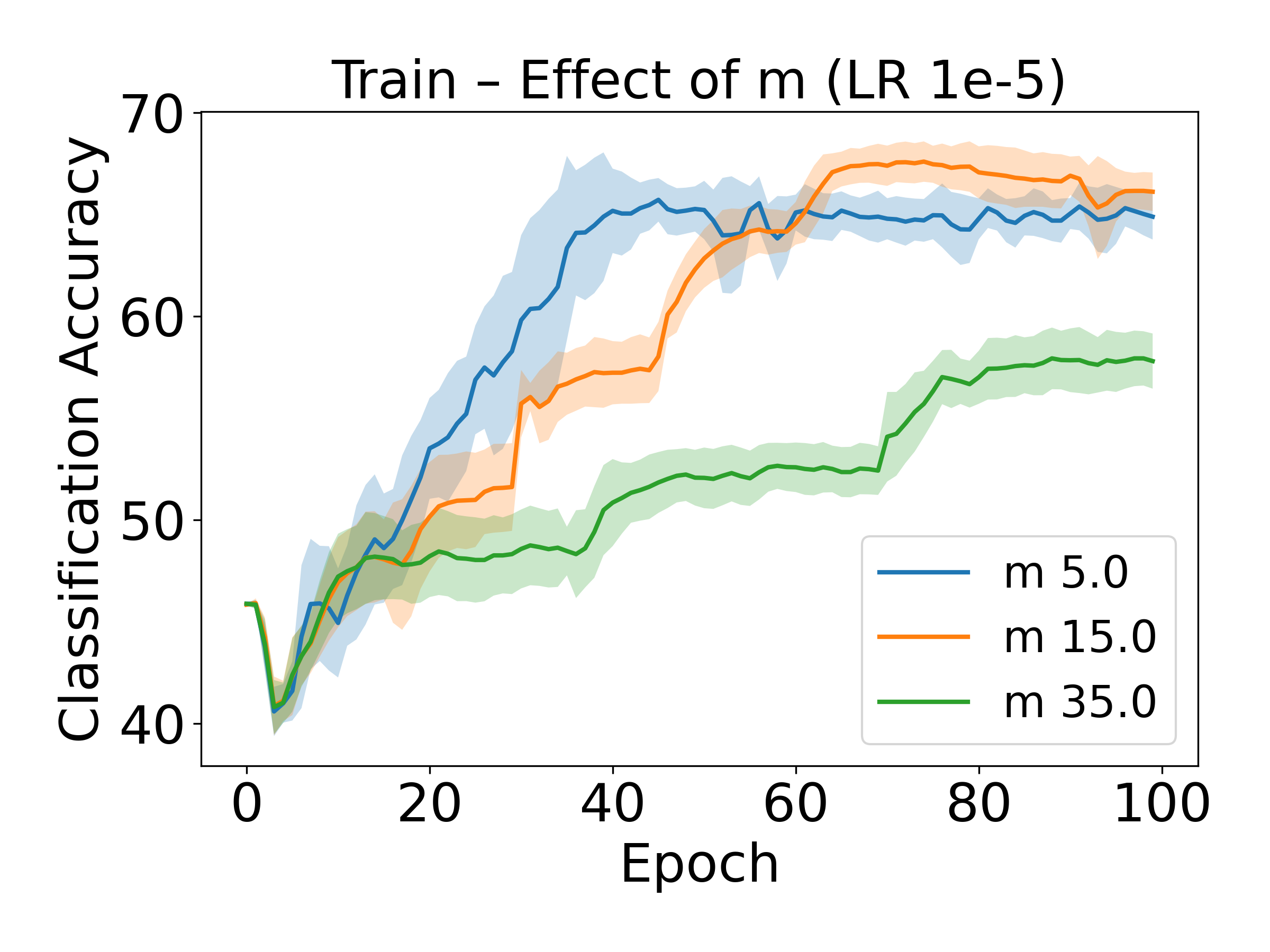}
    \hfill
    \includegraphics[width=0.22\textwidth,
        trim=3cm 0 0 0,
        clip]
        {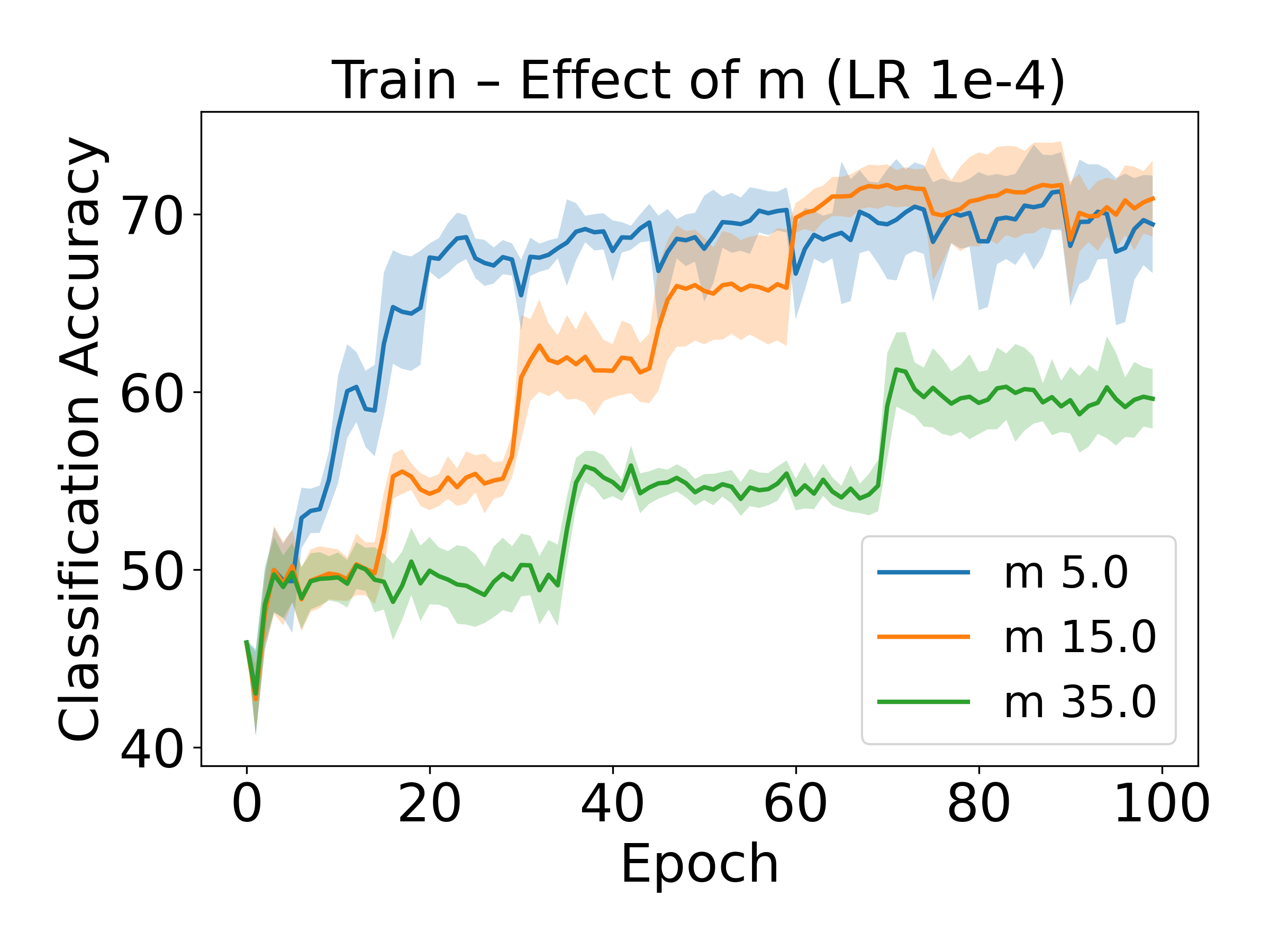}
    \hfill
    \includegraphics[width=0.22\textwidth,
        trim=3cm 0 0 0,
        clip]
        {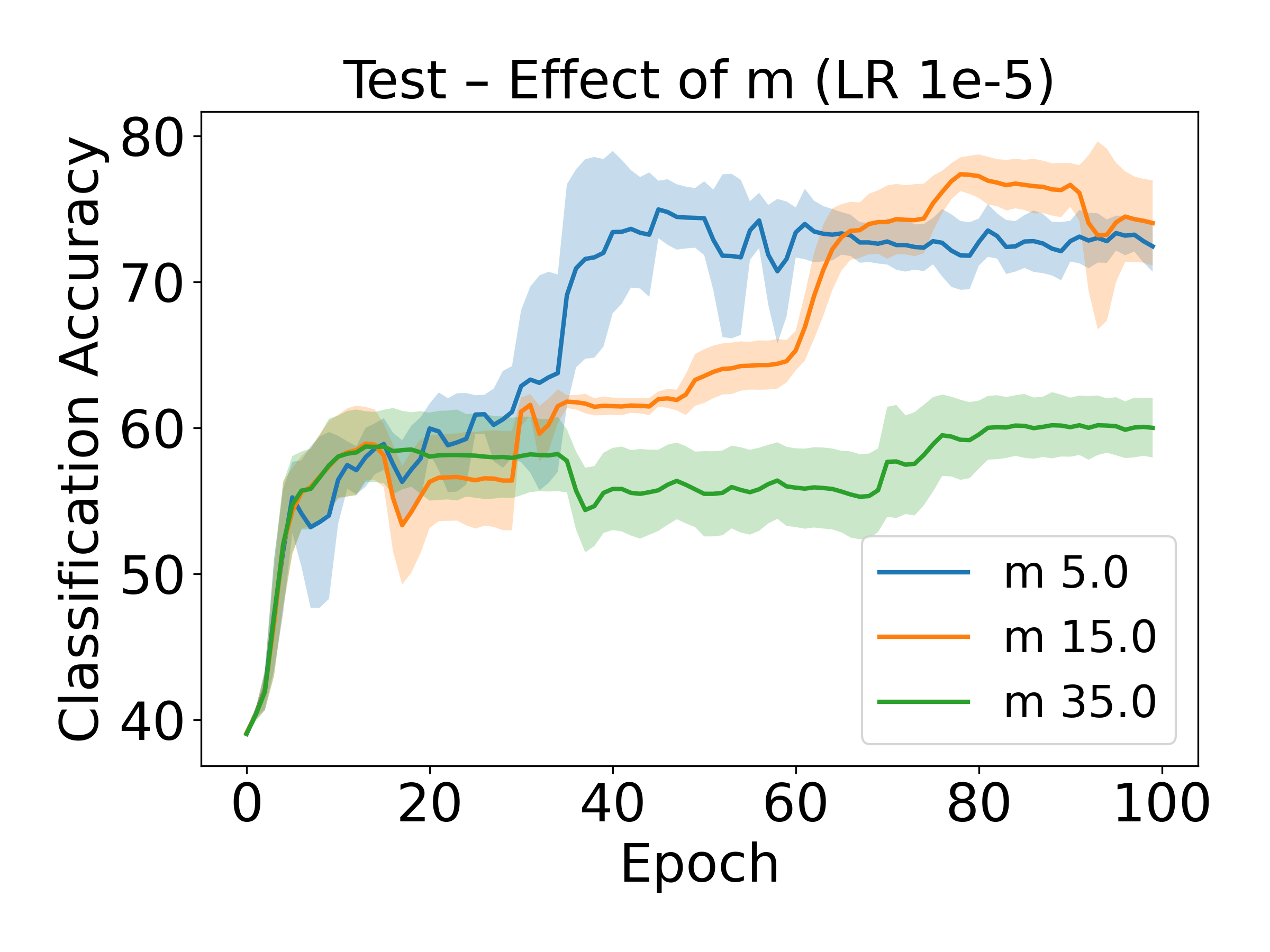}
    \hfill
    \includegraphics[width=0.22\textwidth,
        trim=3cm 0 0 0,
        clip]
        {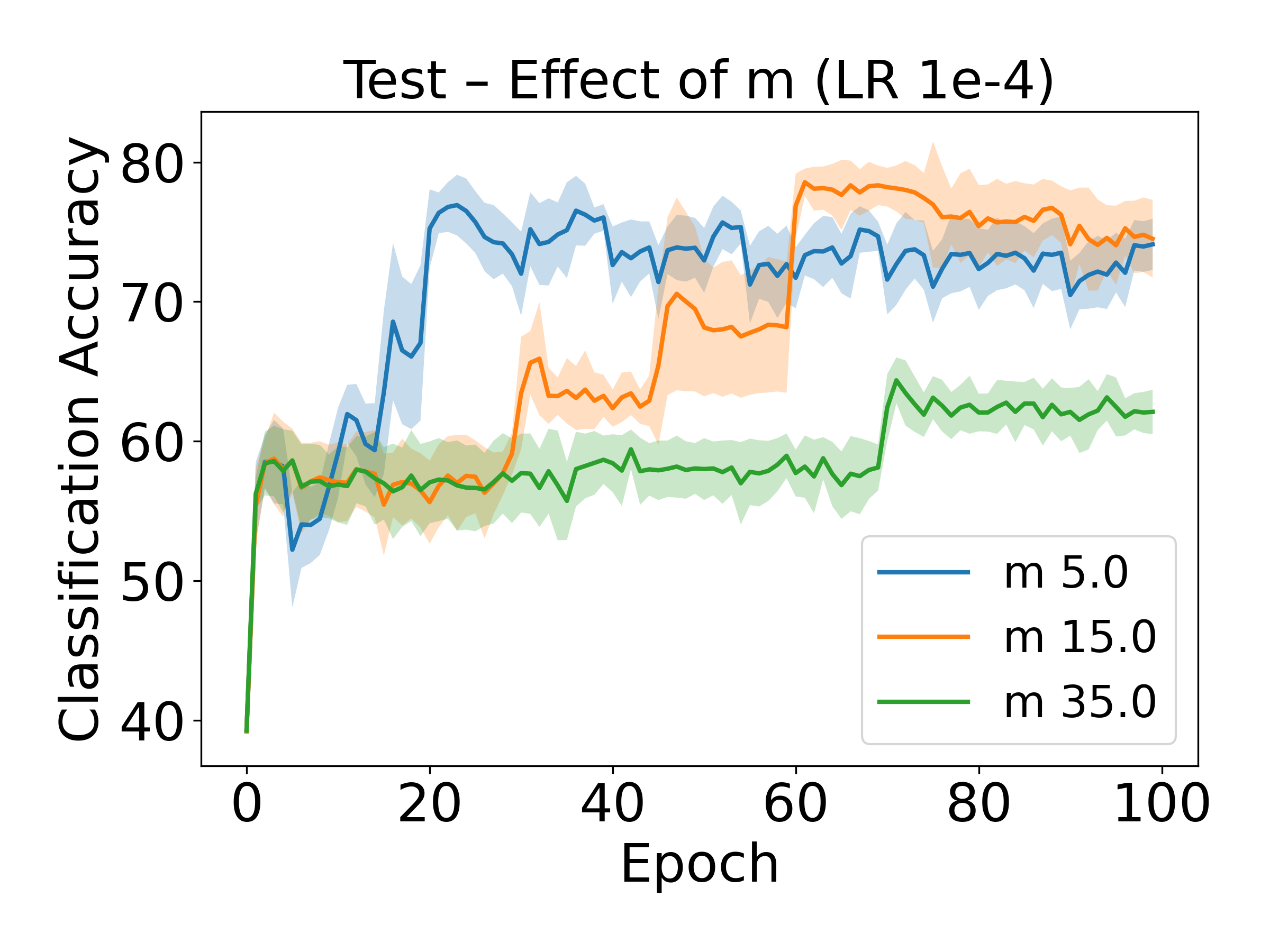}

    \caption{
        Effect of dynamic resampling frequency during adaptation
        with \pixelcam.
    }
    \label{fig:ablations_KP}
\end{figure}

\subsection{Ablation studies}

\noindent\textbf{Contribution of Loss Components.}
Table~\ref{tab:loss_ablation} highlights the role of our three objectives.
\begin{wraptable}{r}{0.33\columnwidth}
\centering
\caption{Impact of loss components during
\sfdadep on GlaS $\rightarrow$ C17-0 with PixelCAM.}
\label{tab:loss_ablation}
\setlength{\tabcolsep}{2pt}
\renewcommand{\arraystretch}{0.7}
\resizebox{\linewidth}{!}{
\begin{tabular}{ccc|cc}
    \toprule
    $\mathcal{L}_{\mathrm{R}}$ &
    $\mathcal{L}_{\mathrm{F}}$ &
    $\mathcal{L}_{\mathrm{L}}$ &
    \pxap $\uparrow$ &
    CL $\uparrow$ \\
    \midrule
    \checkmark &            &            & 27.6 & 41.2 \\
               & \checkmark &            & 10.4 & 85.1 \\
               &            & \checkmark & 9.5 & 48.1 \\
    \checkmark & \checkmark &            & 13.1 & 84.7 \\
    \checkmark & \checkmark & \checkmark & \textbf{49.9} & \textbf{86.1} \\
    \bottomrule
\end{tabular}
}
\vspace{-6pt}
\end{wraptable}
Using $\mathcal{L}_{\mathrm{R}}$ alone better preserves localization
(27.6 \pxap) but fails to correct the classification bias (41.2 \cl),
whereas $\mathcal{L}_{\mathrm{F}}$ improves classification
(85.1 \cl) but provides poor localization (10.4 \pxap). Combining both
image-level objectives maintains high classification accuracy (84.7 \cl),
but remains insufficient for spatial adaptation (13.1 \pxap). Including
$\mathcal{L}_{\mathrm{L}}$ raises \pxap to 49.9 while preserving 86.1 \cl,
showing that prediction debiasing and explicit localization guidance are
required to jointly improve both tasks.

\begin{figure}[!ht]
    \centering

    \begin{subfigure}[t]{0.49\textwidth}
        \centering
        \includegraphics[width=0.9\linewidth]
            {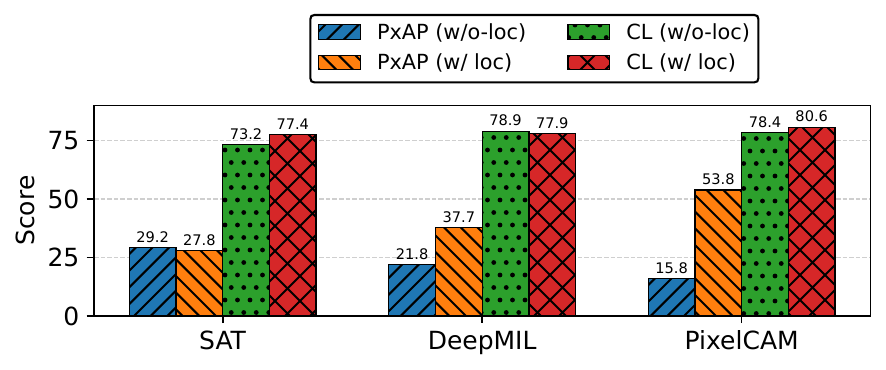}
        \caption{}
        \label{fig:ablation_loc_loss}
    \end{subfigure}
    \hfill
    \begin{subfigure}[t]{0.49\textwidth}
        \centering
        \includegraphics[width=0.65\linewidth]
            {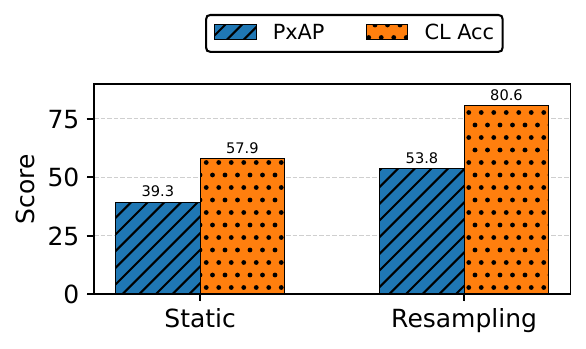}
        \caption{}
        \label{fig:ablation_resampling}
    \end{subfigure}
    \caption{
        Ablation analysis of \sfdadep on \glas $\rightarrow$ \camseventhree with \pixelcam.
        \textbf{(a)} Impact of the localization loss during adaptation.
        \textbf{(b)} Comparison between \textit{static sets}, where forget and retain sets are constructed once before adaptation, and \textit{dynamic resampling} are periodically reconstructed.
    }
    \label{fig:ablation_training_components}
\end{figure}

\noindent\textbf{Localization Task.}
As shown in Fig.~\ref{fig:ablation_loc_loss}, adding $\mathcal{L}_{\mathrm{L}}$ improves \pxap for
\pixelcam from 15.8 to 53.8 and for \deepmil from 21.8 to 37.7,
while producing smaller changes in \cl. This confirms that the
localization objective primarily preserves spatial discrimination.
For \sat, \pxap instead slightly decreases from 29.2 to 27.8,
possibly because the lower-resolution representations provide
less reliable pixel-level pseudo-labels. Nevertheless, \sfdadep reaches an average \pxap of 30.3 with \sat, compared with 21.3 for source-only inference and 22.4 for the strongest SFDA baseline, highlighting the effectiveness of the forget-and-retain mechanism for improving transformer-based WSOL
adaptation.  We analyze the quality of these source-derived localization cues in Appendix~\ref{app:source_cams}. 
\begin{figure}[h]
    \centering
    \includegraphics[width=0.7\linewidth]
    {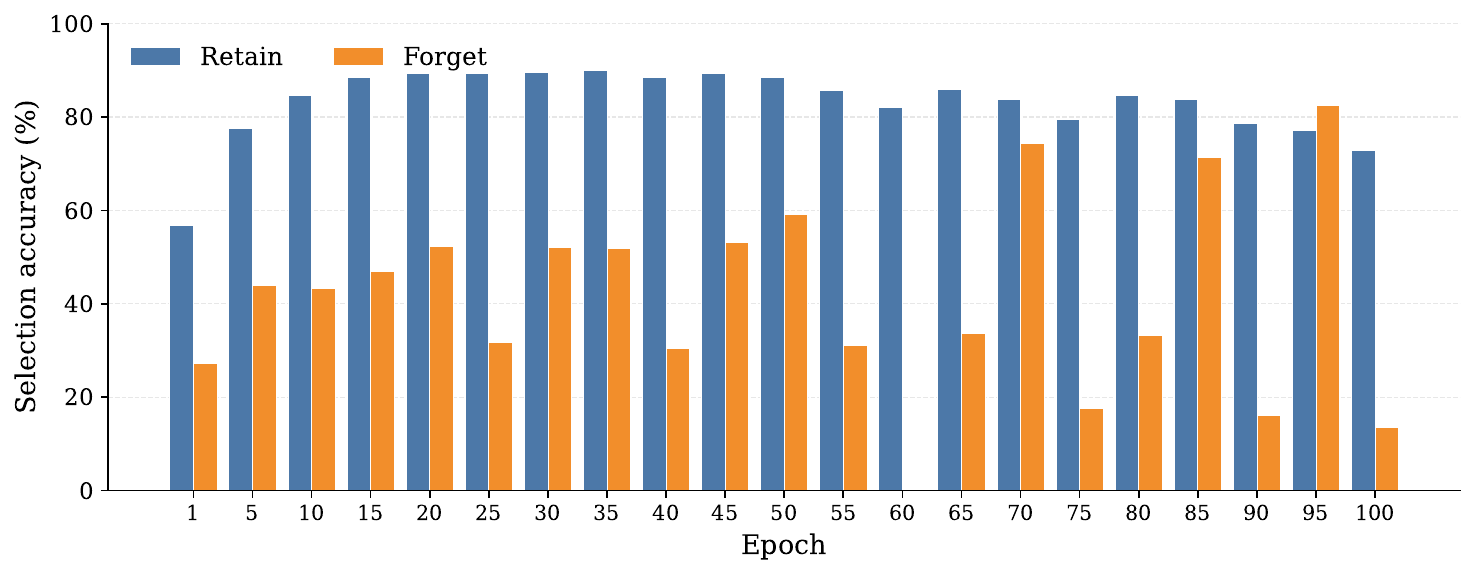}
    \captionsetup{font=small}
    \caption{Evolution of retain- and forget-set selection accuracy during \glas \(\rightarrow\) \camsevenfour adaptation with \deepmil. The forget and retain sets are dynamically reconstructed every five epochs.}
    \label{fig:forget_accuracy}
\end{figure}
\noindent\textbf{Impact of dynamic resampling.}
As illustrated in Fig.~\ref{fig:ablations_KP}, more frequent resampling enables \sfdadep to improve classification more rapidly. This observation supports our hypothesis that uncertain samples located near the current decision boundary provide useful signals to correct dominant-class prediction bias. As shown in Fig.~\ref{fig:ablation_resampling}, the static strategy limits both classification and localization performance. Indeed, as adaptation progressively modifies the target decision boundary, the initial set assignments become increasingly irrelevant. 
\noindent\textbf{Quality of forget-set selection.}
Because \sfdadep relies on identifying samples whose current dominant-class assignments should be reconsidered, Fig.~\ref{fig:forget_accuracy} shows the evolution of retain and forget selection for \deepmil on \camseventeenfour with  \glas as the source.

At each reconstruction step, selection accuracy measures the agreement between
the dynamically selected retain/forget samples and their ground-truth status. The retain set remains consistently more reliable, while the forget set exhibits variations throughout adaptation, with low accuracy during training.

\begin{figure}[h]
    \centering
    \includegraphics[width=0.85\linewidth,trim={0 40.5cm 0 0},clip]
    {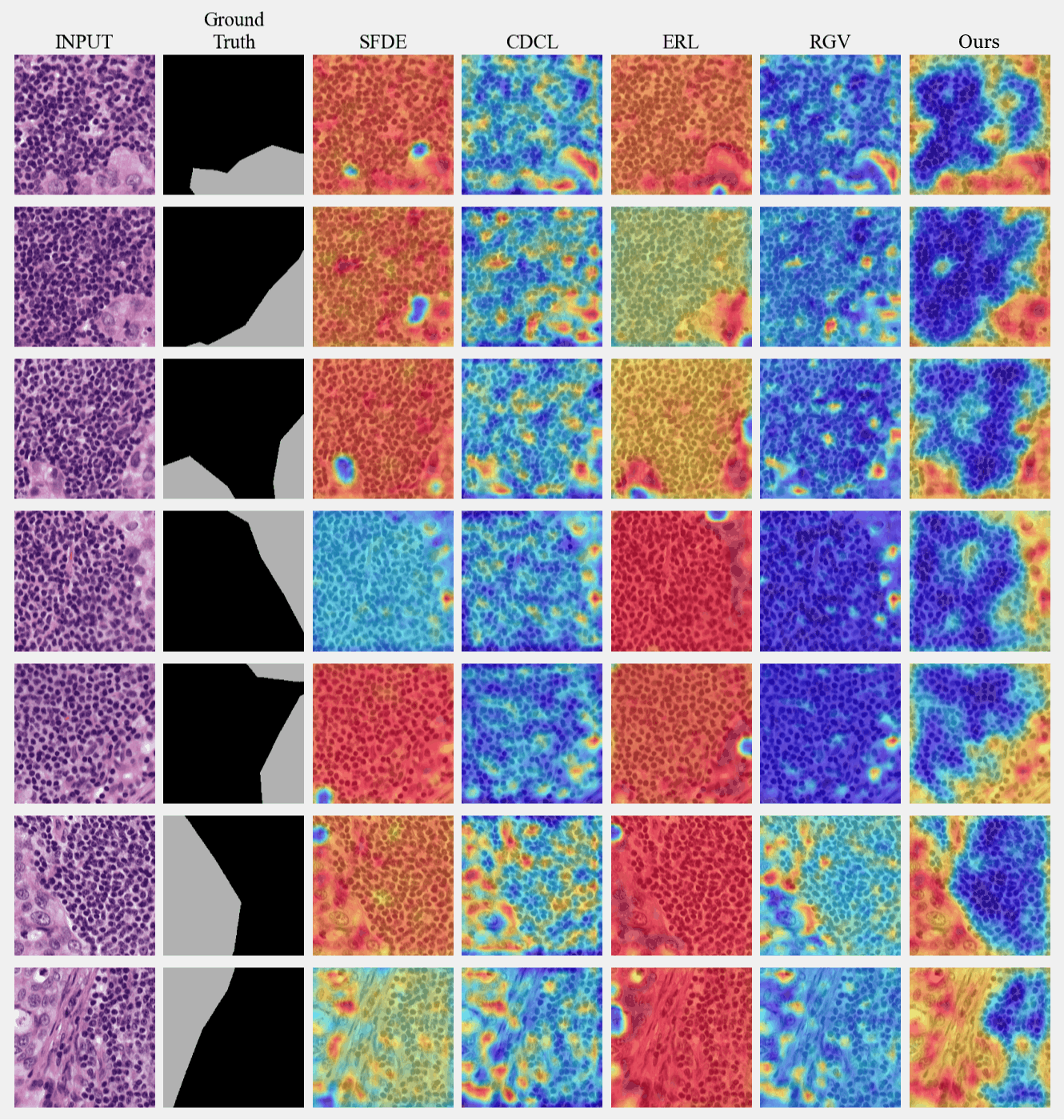}
    \caption{Localization produced by the \pixelcam WSOL model after \sfdadep and the state-of-the-art SFDA methods (\glas $\rightarrow$ \camseventeenthree).}
    \label{fig:cam-example}
\end{figure}

\section{Conclusion}
This paper addresses dominant-class prediction bias, a critical challenge in source-free adaptation for WSOL models in histopathology. We introduced \sfdadep, a new SFDA method that corrects this bias through complementary forget and retain objectives. The forget set contains uncertain samples currently assigned to dominant classes and likely to lie near the target decision boundary. The proposed forgetting objective reduces the probability assigned to their current dominant class, allowing under-predicted classes to emerge and progressively reducing the skew in the target prediction distribution. In parallel, the retain objective preserves the predictions of remaining target samples through cross-entropy supervision, thereby maintaining useful target structure and limiting excessive model drift. 
Experiments show that \sfdadep achieves the best average localization performance across all evaluated WSOL architectures while improving average classification accuracy.

\noindent\textbf{Supplementary Material:} It provides additional information on datasets, implementation, and quantitative results. It also includes additional analyses of target-class imbalance, hyperparameter sensitivity, source-CAM reliability, the multi-class setting, and entropy-based prediction behavior.

\balance
\bibliographystyle{abbrv}
\bibliography{main}

\newpage
\appendix
\setcounter{topnumber}{3}
\setcounter{bottomnumber}{2}
\setcounter{totalnumber}{5}

\renewcommand{\topfraction}{0.95}
\renewcommand{\bottomfraction}{0.90}
\renewcommand{\textfraction}{0.05}
\renewcommand{\floatpagefraction}{0.80}

\section*{Supplementary Material}

\begin{center}
{\large\textbf{Table of Contents}}
\end{center}

\vspace{0.4em}

{
\noindent\textbf{A\quad Dataset Construction and Experimental Protocol}
\dotfill\pageref{app:dataset_construction}\par


\noindent\hspace*{1.5em}
A.1\quad CAMELYON17 Dataset Construction
\dotfill\pageref{app:c17_construction}\par


\noindent\hspace*{1.5em}
A.2\quad Visualization of CAMELYON17
\dotfill\pageref{app:c17_visualization}\par


\noindent\hspace*{1.5em}
A.3\quad EBHI Corruption Benchmark
\dotfill\pageref{app:ebhi_corruptions}\par


\noindent\hspace*{1.5em}
A.4\quad Visualization of EBHI
\dotfill\pageref{app:ebhi_visualization}\par

\vspace{0.35cm}

\noindent\textbf{B\quad Implementation and Reproducibility Details}
\dotfill\pageref{app:implementation}\par


\noindent\hspace*{1.5em}
B.1\quad Training and Adaptation Hyperparameters
\dotfill\pageref{app:hyperparameters}\par


\noindent\hspace*{1.5em}
B.2\quad Experimental Settings
\dotfill\pageref{app:experimental_settings}\par


\noindent\hspace*{1.5em}
B.3\quad \sfdadep Training Algorithm
\dotfill\pageref{app:algorithm}\par

\vspace{0.35cm}

\noindent\textbf{C\quad Additional Quantitative Results}
\dotfill\pageref{app:additional_results}\par


\noindent\hspace*{1.5em}
C.1\quad Cross-Organ Results
\dotfill\pageref{app:cross_organ}\par


\noindent\hspace*{1.5em}
C.2\quad Robustness to Target-Class Imbalance
\dotfill\pageref{app:imbalance_cancer}\par


\noindent\hspace*{1.5em}
C.3\quad Sensitivity to the Forgetting Proportion
\dotfill\pageref{tab:forget_ratio_ablation}\par


\noindent\hspace*{1.5em}
C.4\quad Sensitivity to the Localization Keep Ratio
\dotfill\pageref{tab:keep_ratio}\par


\noindent\hspace*{1.5em}
C.5\quad Backbone and Classifier Adaptation
\dotfill\pageref{tab:freezing_ablation}\par


\noindent\hspace*{1.5em}
C.6\quad Extension to the Multi-Class Setting
\dotfill\pageref{app:cnn_transformer}\par


\noindent\hspace*{1.5em}
C.7\quad CNNs versus Transformer-Based Models
\dotfill\pageref{app:cnn_transformer}\par

\vspace{0.35cm}

\noindent\textbf{D\quad Reliability of Source CAMs for Pixel Pseudo-Labeling}
\dotfill\pageref{app:source_cams}\par


\noindent\hspace*{1.5em}
D.1\quad Reliability of High-Confidence Foreground Regions
\dotfill\pageref{app:foreground_reliability}\par


\noindent\hspace*{1.5em}
D.2\quad Reliability of Low-Activation Background Regions
\dotfill\pageref{app:background_reliability}\par

\vspace{0.35cm}

\noindent\textbf{E\quad Additional Visualizations}
\dotfill\pageref{app:visualizations}\par


\noindent\hspace*{1.5em}
E.1\quad Entropy Distributions
\dotfill\pageref{app:entropy_distributions}\par


\noindent\hspace*{1.5em}
E.2\quad Entropy--Accuracy Relationships
\dotfill\pageref{app:entropy_accuracy}\par

\noindent\hspace*{1.5em}
E.3\quad Additional Visualizations
\dotfill\pageref{app:Visualizations}\par
}

\section{Dataset Construction and Experimental Protocol}
\label{app:dataset_construction}
\subsection{CAMELYON17 Dataset Construction}
\label{app:c17_construction}
In this work, we do not use the standard \camseventeen~\citep{c17} classification benchmark. Instead, we construct a patch-level version of \camseventeen specifically designed for Weakly Supervised Object Localization (WSOL). Each extracted patch is associated with an image-level normal/tumor label used for classification and WSOL training, while the corresponding pixel-level tumor annotation is retained exclusively for localization evaluation. Our construction protocol closely follows the patch-sampling strategy introduced in \citep{rony23} for \camsixteen~\citep{rony23,pixelcam}. Importantly, patches originating from the five acquisition centers of \camseventeen are kept as separate domains, resulting in five domain-specific datasets, denoted \camsevenzero to \camsevenfour. This construction enables controlled source-to-target experiments across acquisition centers while preserving the same tissue type and prediction task.
\subsubsection{Patch Extraction Procedure}
\paragraph{Sampling of Metastatic Patches}
Metastatic patches are sampled exclusively from metastatic whole-slide images (WSIs) and only in regions surrounding cancerous tissue.
Each sampled patch is associated with both an image-level label and a pixel-level annotation mask derived from the WSI annotations.

\begin{algorithm}[H]
\caption{Sampling of Metastatic Patches for WSOL}
\label{alg:metastatic_sampling}

{\footnotesize
\begin{algorithmic}[1]

\Require
\parbox[t]{0.8\columnwidth}{
Metastatic WSI with pixel-level annotation, patch size $(H,W)$,\\
and thresholds $p_t$ and $p_0$
}

\Ensure
\parbox[t]{0.78\columnwidth}{
Patch $x$ with image-level label and pixel-level mask $Y$
}

\State Select a metastatic WSI
\State Sample a patch $x$ of size $(H,W)$
\State Binarize $x$ into a tissue mask $x_b$ using Otsu thresholding
\State Compute tissue ratio $p_x^t$ in $x_b$

\If{$p_x^t < p_t$}
\State \textbf{discard} $x$
\EndIf

\State Extract metastatic binary mask $Y$ from the WSI annotation
\State Compute metastatic ratio $p_x^m$ in $Y$

\If{$p_x^m < p_0$}
\State \textbf{discard} $x$
\Else
\State Assign image-level label \emph{metastatic}
\State \textbf{return} $(x,Y)$
\EndIf

\end{algorithmic}
}
\end{algorithm}

\paragraph{Calibration of Metastatic Patches}

Naively sampling all eligible metastatic patches results in a strong bias toward patches fully occupied by cancerous tissue, typically located at the center of metastatic regions.
Such patches produce binary masks almost filled with positive labels, which artificially inflates localization performance and undermines a fair evaluation of WSOL methods.
To mitigate this issue, we adopt the calibration strategy proposed in \citep{rony23}, summarized in Algorithm~\ref{alg:metastatic_calibration}.

\begin{algorithm}[H]
\caption{Calibration of Metastatic Patches}
\label{alg:metastatic_calibration}

{\footnotesize
\begin{algorithmic}[1]

\Require
\parbox[t]{0.8\columnwidth}{
Set of metastatic patches $\mathcal{P}$ with metastatic ratios $p_x^m$,
thresholds $p_0$ and $p_1$, sampling ratio $p_n$, and number of bins $b$
}

\Ensure
\parbox[t]{0.82\columnwidth}{
Calibrated set of metastatic patches $\mathcal{P}_{\mathrm{cal}}$
}

\State Partition $\mathcal{P}$ into:
\State \hspace{0.5em}
$\mathcal{P}_1 = \{x \mid p_0 \leq p_x^m \leq p_1\}$
\State \hspace{0.5em}
$\mathcal{P}_2 = \{x \mid p_x^m > p_1\}$

\State Initialize $\mathcal{P}_{\mathrm{cal}} \leftarrow \mathcal{P}_1$
\State $n \leftarrow |\mathcal{P}_1|$
\State $k \leftarrow n p_n$

\State Compute histogram $\mathcal{H}$ of $p_x^m$ for patches in
$\mathcal{P}_2$ using $b$ bins

\While{$|\mathcal{P}_{\mathrm{cal}}| < n + k$}
    \State Uniformly sample a bin from $\mathcal{H}$ with $p_x^m > p_1$
    \State Uniformly sample a patch $x$ from the selected bin
    \State Add $x$ to $\mathcal{P}_{\mathrm{cal}}$
\EndWhile

\State \textbf{return} $\mathcal{P}_{\mathrm{cal}}$

\end{algorithmic}
}
\end{algorithm}

\paragraph{Sampling of Normal Patches}

Normal patches are sampled exclusively from normal WSIs.
Patches are extracted randomly and uniformly without overlap.
A patch is accepted if its tissue ratio $p^t_x$ satisfies $p^t_x \geq p_t$.

To ensure class balance at the patch level, we sample the same number of normal patches as metastatic patches within each data split (training, validation, or test).
Given $h$ normal WSIs and $n_m$ metastatic patches, we sample $n_m / h$ normal patches per WSI.
Tissue mass estimation is performed at resolution level~6 using Otsu thresholding and verified at level~0.
\paragraph{Final Configuration}
All patches are extracted without overlap and resized to $512 \times 512$ pixels.
We adopt the same hyperparameter configuration as in \citep{rony23}. This construction yields a \camseventeen WSOL benchmark with realistic spatial ambiguity, making it well suited for evaluating weakly supervised localization and source-free domain adaptation methods.
\paragraph{Dataset Statistics}
\begin{table} [h]
\caption{Dataset splits used in our experiments, showing the number of patches per split for each dataset and center.}
\label{tab:datasplit_appendix}
\centering
\small
\setlength{\tabcolsep}{2.8pt}
\renewcommand{\arraystretch}{0.9}
\begin{tabular}{lccccccc}
\toprule
 & GlaS & C16 & C17-0 & C17-1 & C17-2 & C17-3 & C17-4 \\
\midrule
Train      & 67 & 24348 & 634 & 1066 & 498 & 816 & 940 \\
Val (CL)   & 18 & 8850 & 144 & 38 & 110 & 102 & 146 \\
Val (\pxap) & 6 & 10 & 10 & 10 & 10 & 10 & 10 \\
Test       & 80 & 15664 & 262 & 172 & 448 & 376 & 298 \\
\bottomrule
\end{tabular}
\end{table}
Table~\ref{tab:datasplit_appendix} reports the resulting number of patches for each
center and experimental split.
Following the same protocol across centers, we construct separate training,
classification-validation, localization-validation, and test sets.
The localization-validation images and their pixel-level annotations are
never used to optimize the adaptation objective and are only employed for model selection/evaluation of localization performance, whereas the
classification-validation split is used for classification model selection. We provide additional visualization in the following section. For cancer images, ground-truth tumor regions are overlaid in red, whereas normal images contain no annotated tumor regions.

\subsection{Visualization of \camseventeen}
\label{app:c17_visualization}

\begin{figure*}[t]
\centering
\includegraphics[
    width=0.9\linewidth
]{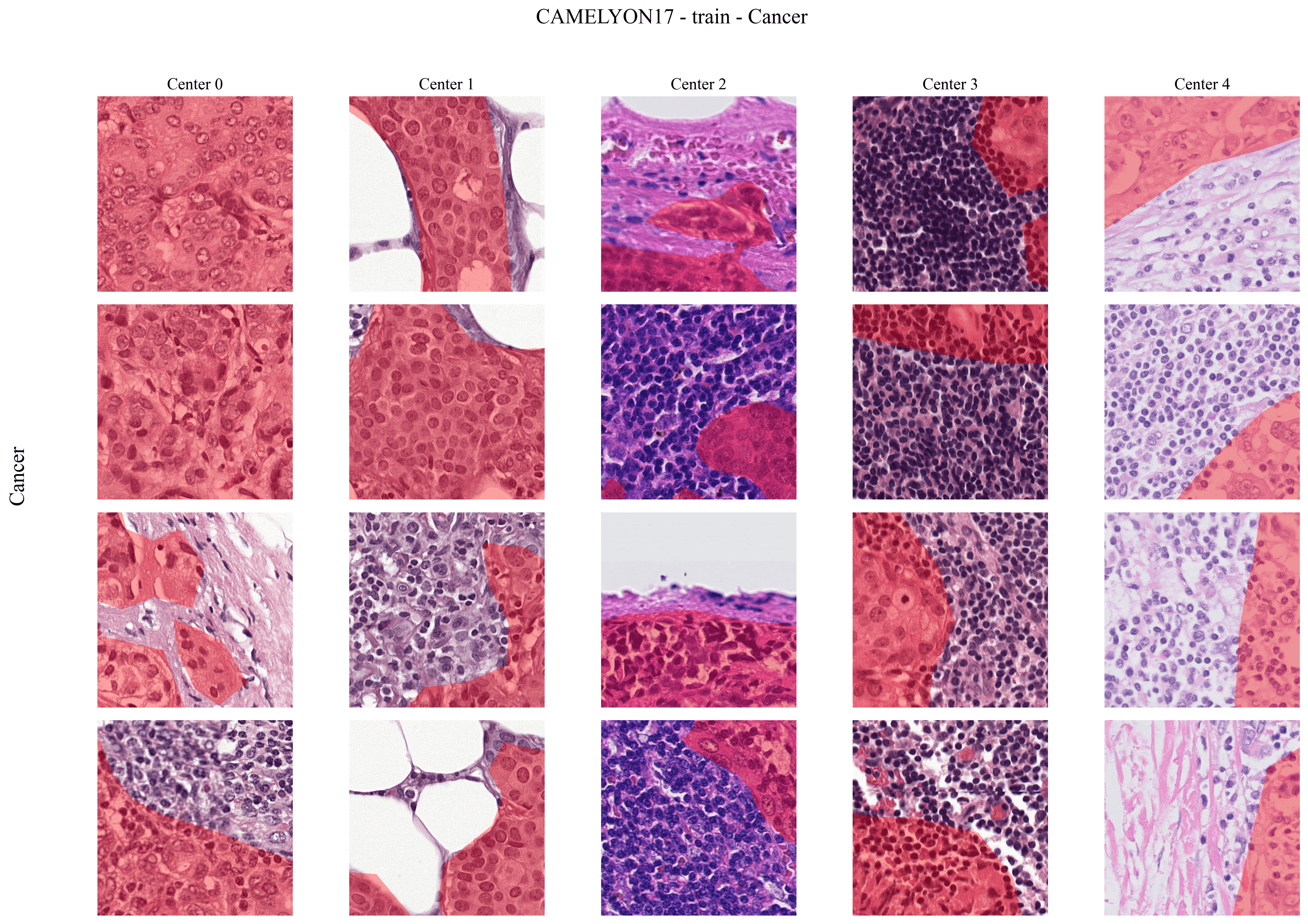}
\caption{
Representative cancer examples from the \textbf{training split} of the constructed
\camseventeen WSOL benchmark across the five centers.
Ground-truth tumor regions are overlaid in red.
}
\label{fig:camelyon17_wsol_train_c}
\end{figure*}

\begin{figure*}[t]
\centering
\includegraphics[
    width=0.9\linewidth
]{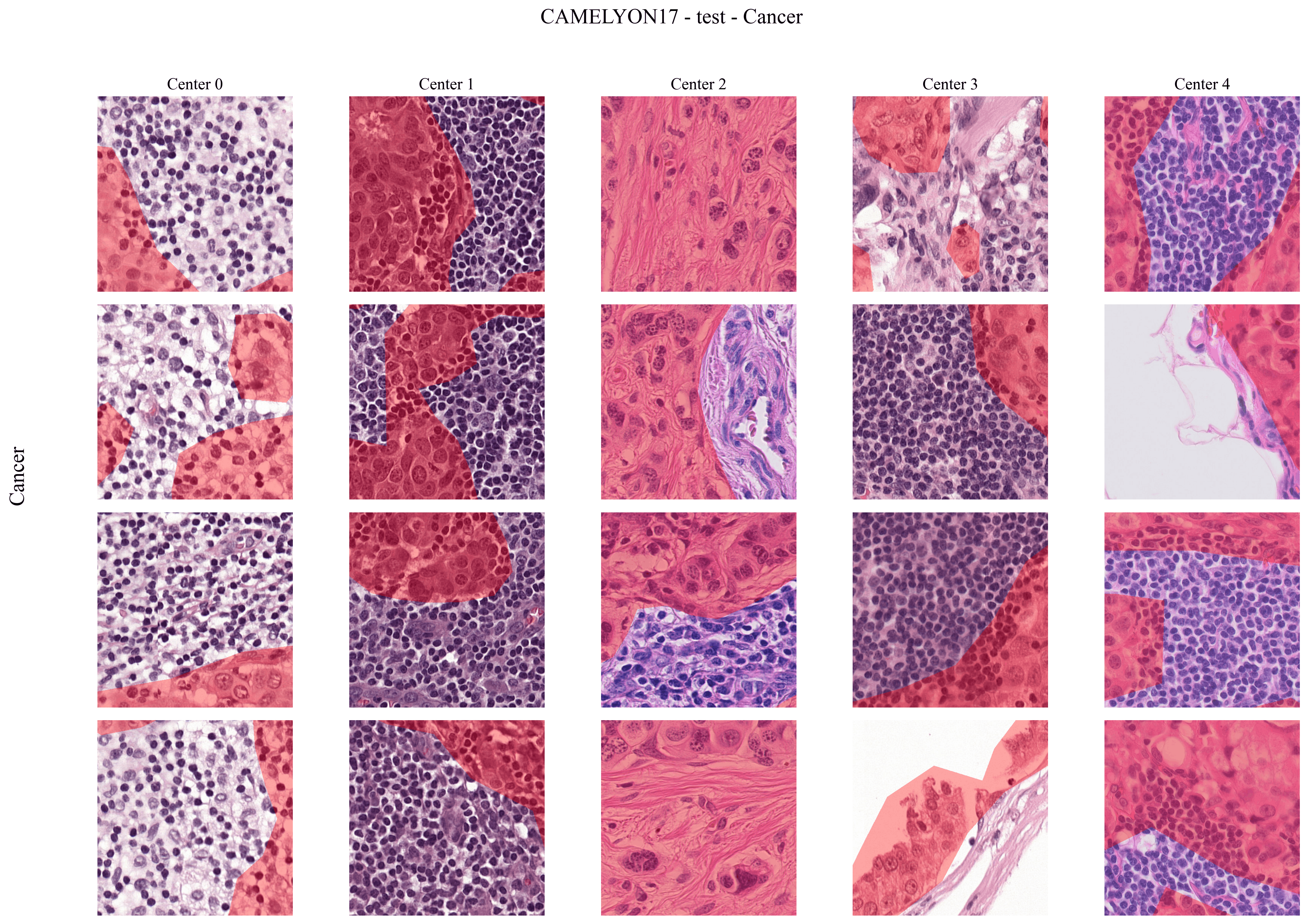}
\caption{
Representative cancer examples from the \textbf{testing split} of the constructed
\camseventeen WSOL benchmark across the five centers.
Ground-truth tumor regions are overlaid in red.
}
\label{fig:camelyon17_wsol_test_c}
\end{figure*}
\subsection{EBHI Corruption Benchmark}
\label{app:ebhi_corruptions}
To extend our evaluation to a challenging multi-class setting, we construct a new benchmark for SFDA and WSOL in colorectal histopathology based on \ebhi~\cite{ebhi}. \ebhi comprises six classes and has been previously used for colorectal cancer classification and segmentation. We split the dataset into disjoint source and target subsets. The source subset contains clean histopathology images and is used exclusively for source-model training, while the target subset is held out from source training and used to construct different target-domain shifts. We further divide each subset into training, validation, and test sets while preserving the original class distribution. Table~\ref{tab:ebhi_split} summarizes the resulting class-wise partition.
\paragraph{Corruption Types :} To construct controlled target-domain shifts, we follow the Histopath-C benchmark~\citep{noori2026histopathc}, which introduces
histopathology-specific synthetic corruptions designed to mimic distribution shifts encountered in digital pathology. Histopath-C groups ten corruption types into five categories:
\textit{staining}, \textit{contamination}, \textit{blurring}, \textit{noise}, and \textit{illumination}.

Specifically, we consider two staining perturbations (\textit{Stain-Light} and \textit{Stain-Heavy}), two contamination artifacts (\textit{Dust} and \textit{Air Bubble}), two blur
perturbations (\textit{Defocus Blur} and \textit{Motion Blur}), two noise perturbations (\textit{Gaussian Noise} and \textit{Shot Noise}), and two illumination perturbations
(\textit{Brightness} and \textit{Contrast}).

We apply each corruption independently to the held-out \ebhitarget subset, resulting in 11 distinct target domains. The source domain remains composed essentially of clean images,
ensuring that the observed distribution shift arises only from the applied corruption.

\begin{table}[H]
\centering
\caption{Class distribution of the EBHI dataset and the constructed source/target splits.}
\label{tab:ebhi_split}
\resizebox{0.99\columnwidth}{!}{
\begin{tabular}{lrrrrr}
\toprule
\textbf{Class} & \textbf{Dataset} & \textbf{Source/Target} & \textbf{Train} & \textbf{Val} & \textbf{Test} \\
\midrule
Normal              & 152   & 76  & \textbf{46}  & \textbf{15}  & \textbf{15}  \\
Polyp               & 948   & 474 & \textbf{284} & \textbf{95}  & \textbf{95}  \\
Low-grade IN        & 1,276 & 638 & \textbf{382} & \textbf{128} & \textbf{128} \\
High-grade IN       & 372   & 186 & \textbf{112} & \textbf{37}  & \textbf{37}  \\
Serrated adenoma    & 116   & 58  & \textbf{34}  & \textbf{12}  & \textbf{12}  \\
Adenocarcinoma      & 1,590 & 795 & \textbf{477} & \textbf{159} & \textbf{159} \\
\midrule
\textbf{Total}      & \textbf{4,454} & \textbf{2,227} & \textbf{1,335} & \textbf{446} & \textbf{446} \\
\bottomrule
\end{tabular}
}
\end{table}
\begin{table}[h]
\centering
\caption{Parameters used to construct the corrupted EBHI-Seg target
domains. We follow the official Histopath-C implementation~\citep{noori2026histopathc}.}
\label{tab:ebhi_corruption_params}
\resizebox{0.95\columnwidth}{!}{
\begin{tabular}{lll}
\toprule
\textbf{Category} & \textbf{Corruption} & \textbf{Parameters} \\
\midrule
Staining
    & Stain-Light
    & $\theta=0.05$ \\
    & Stain-Heavy
    & $\theta=0.10$ \\
\midrule
Contamination
    & Dust
    & intensity $=0.7$, $N_{\mathrm{smudges}}\in[3,10]$ \\
    & Air Bubble
    & $N_{\mathrm{bubbles}}\in[10,20]$, $\alpha=0.3$, blur severity $=2$ \\
\midrule
Blurring
    & Defocus Blur
    & severity $=5$, radius $=10$, alias blur $=0.5$ \\
    & Motion Blur
    & severity $=5$, radius $=20$, $\sigma=15$ \\
\midrule
Noise
    & Gaussian Noise
    & severity $=3$, $\sigma=0.18$ \\
    & Shot Noise
    & severity $=3$, $c=12$ \\
\midrule
Illumination
    & Brightness
    & severity $=5$, offset $=0.5$ \\
    & Contrast
    & severity $=5$, scale $=0.05$ \\
\bottomrule
\end{tabular}
}
\end{table}
\subsection{Visualization of EBHI}
\label{app:ebhi_visualization}
\begin{figure*}[!t]
\centering
\includegraphics[width=0.99\linewidth,trim={0 22.8cm 0 0},clip]{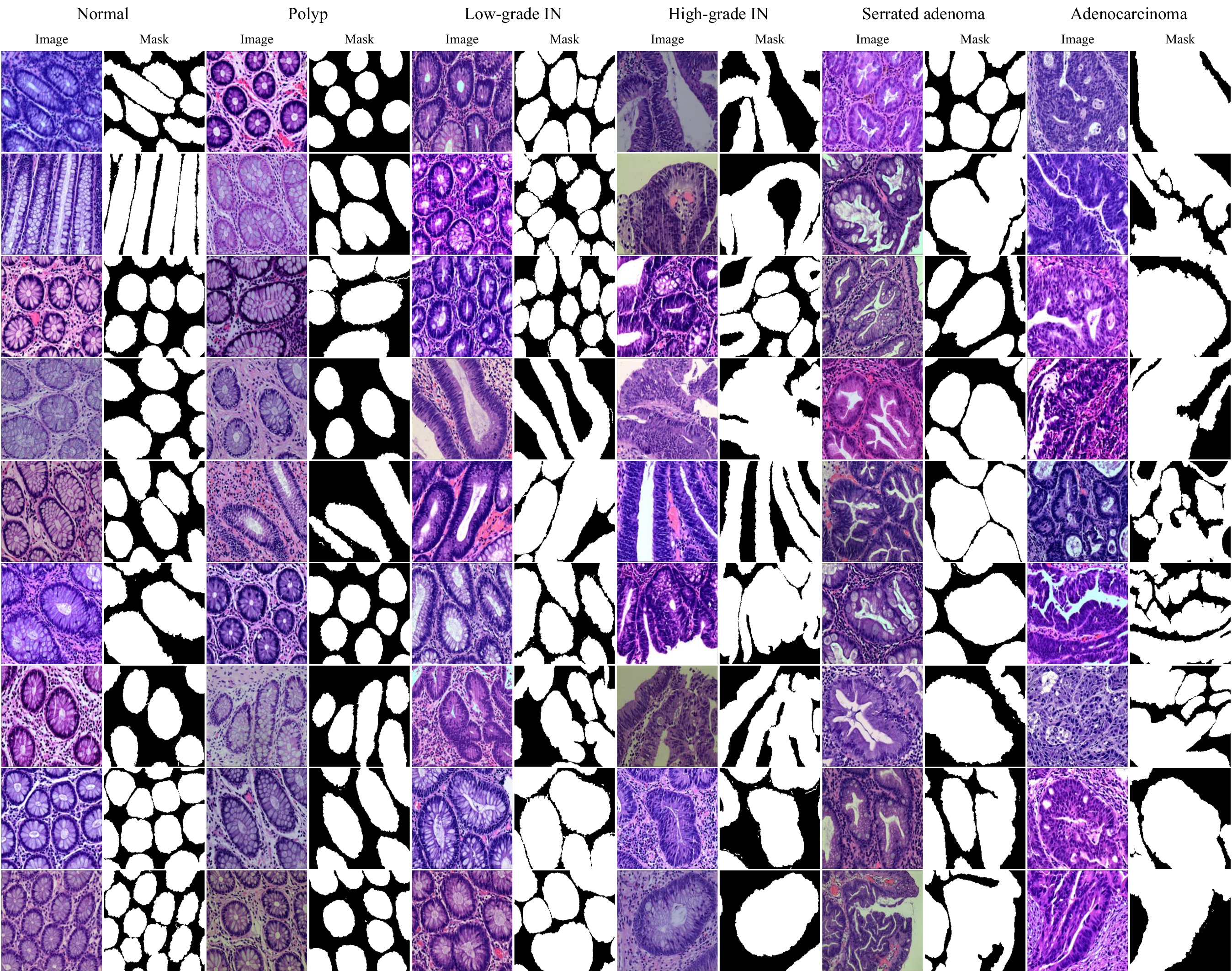}
\caption{
Examples from the \textbf{training split} of the constructed \ebhisource WSOL benchmark.
}
\label{fig:camelyon17_wsol_train_c}
\end{figure*}

\begin{figure*}[!t]
\centering
\includegraphics[width=0.99\linewidth]{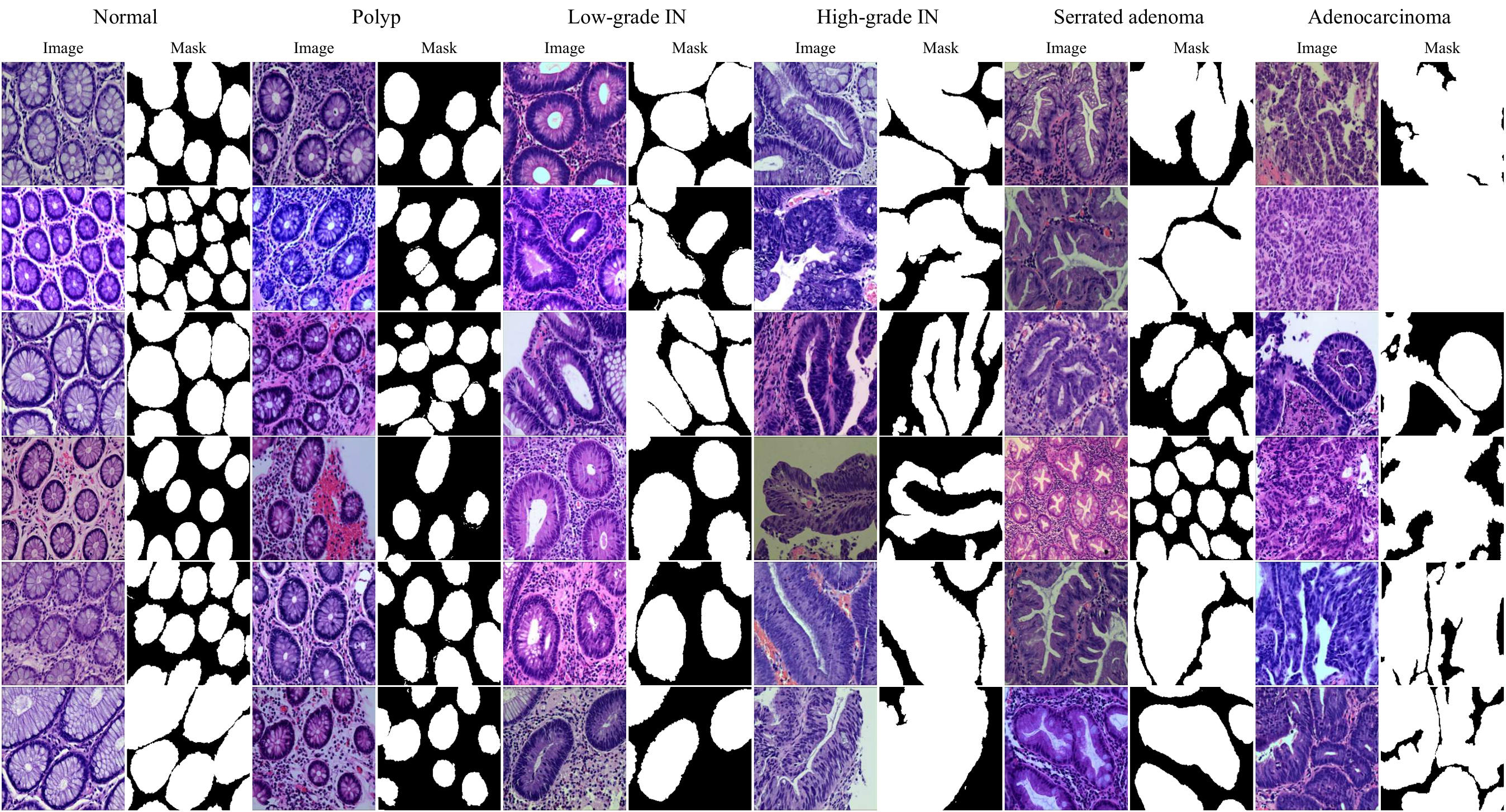}
\caption{
Examples from the \textbf{testing split} of the constructed \ebhisource WSOL benchmark.
}
\label{fig:camelyon17_wsol_train_c}
\end{figure*}

\begin{figure*}[!t]
\centering
\includegraphics[width=0.99\linewidth,trim={0 0 0 0},clip]{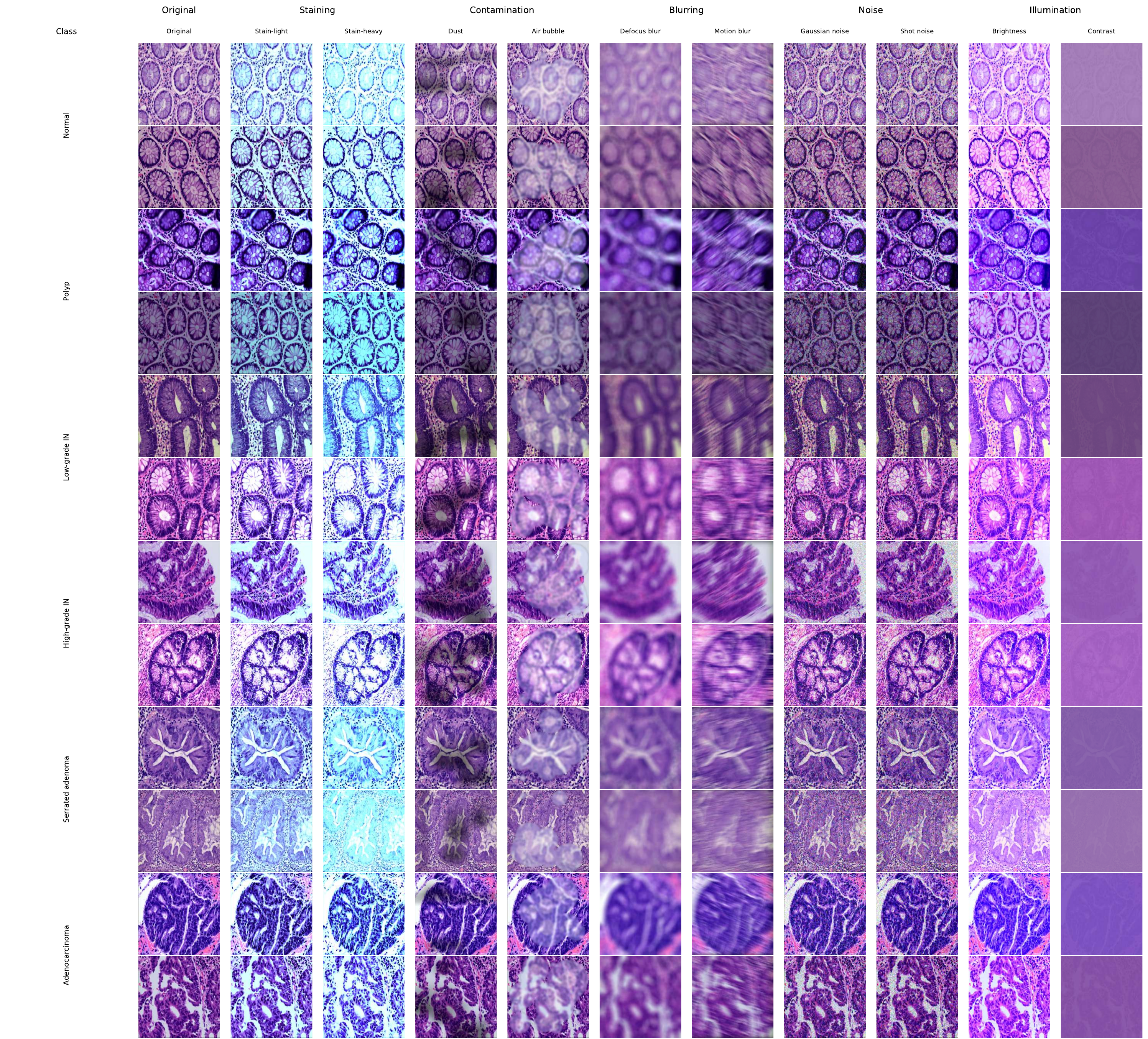}
\caption{
Examples from the \textbf{testing split} of the constructed \camseventeen WSOL benchmark.
}
\label{fig:camelyon17_wsol_train_c}
\end{figure*}
    


\section{Implementation and Reproducibility Details}
\label{app:implementation}

\subsection{Training and Adaptation Hyperparameters}
\label{app:hyperparameters}
This section provides the complete list of hyperparameters used throughout our experiments.
We distinguish between general training parameters, WSOL-specific parameters, and method-specific hyperparameters for SFDA baselines.
\begin{table}[h]
\renewcommand{\arraystretch}{1.3}
\caption{General hyperparameters used during source model training.}
\label{tab:tabx-general-hyper-params-source}
\centering
\resizebox{.8\linewidth}{!}{
\begin{tabular}{lccc}
    Hyperparameter  &  Value  \\
    \toprule
    Backbones & ResNet50 / DeiT\\
    \hline
    Optimizer &  SGD\\
    \hline
    Nesterov acceleration & True\\
    \hline
    Momentum & ${\in \{0.1, 0.4, 0.9\}}$ for source models\\
    \hline
    Weight decay & $0.0001$\\
    \hline
    Learning rate & \makecell{${\in \{0.0001, 0.001, 0.002, 0.01, 0.02, 0.1\}}$} for source\\
    \hline
    Learning rate decay
     & \glas: 0.1 ${\in  \{150,250,350e  \}}$ epochs for source\\
     \hline
    Mini-batch size & \makecell{$32$} \\
    \hline
    Random flip & Horizontal/vertical random flip \\
    \hline
    \multirow{2}{*}{Random color jittering} 
    & $\mathrm{Brightness}$, $\mathrm{contrast}$, \\
    & and $\mathrm{saturation}$ at $0.5$ and $\mathrm{hue}$ at $0.05$ \\
    \hline
    \multirow{2}{*}{Image size} 
    & Resize image to $225\times 225$. \\
    & Then, crop random patches of $224\times 224$\\
    \hline
    Learning epochs & \makecell{\glas: $1000$} \\
    \bottomrule
\end{tabular}
}
\end{table}

During the SFDA, all architectural choices and data augmentations are kept identical to source model training.
Only optimization-related parameters and specific hyperparameters are varied. All SFDA methods share the same backbone, optimizer, batch size, and training schedule. Only method-specific loss weights and thresholds are varied.

\begin{table}[h]
\renewcommand{\arraystretch}{1.3}
\caption{General hyperparameters used during SFDA training.}
\label{tab:tabx-general-hyper-params}
\centering
\resizebox{.8\linewidth}{!}{
\begin{tabular}{lccc}
    Hyperparameter  &  Value  \\
    \toprule
    Momentum & 0.9\\
    \hline
    Weight decay & $0.0001$\\
    \hline
    Learning rate & \makecell{${\in \{0.000001, 0.00001, 0.0001,0.001\}}$}\\
    \hline
    \multirow{2}{*}{Learning rate decay} 
     & \camseventeen: 0.1 ${\in  \{20\}}$ epochs\\
     & \camsixteen: 0.1 ${\in  5}$  epochs\\
     \hline
    Flip ratio (forgotten samples) & ${\in  \{0.05,0.15\}}$\\
    \hline
    \multirow{2}{*}{\sfdadep period (epochs)} 
     & \camseventeen: 5,15,35 epochs\\
     & \camsixteen: 3,5 epochs\\
    \hline
    Learning epochs & \makecell{\camseventeen: $100$, \camsixteen: $15$} \\
     \hline
     \multirow{3}{*}{\sfdadep} 
            & ${\lambda_{F} \in \{0.5, 1.0\}}$ \\
            & ${\lambda_{R} \in \{0.5, 1.0\}}$\\
            & ${\lambda_{L} \in \{0.5,1.0,5.0\}}$\\
    \bottomrule
\end{tabular}
}
\end{table}

\begin{table}[h]
\renewcommand{\arraystretch}{1.}
\caption{Per-method hyperparameters for SFDA methods.}
\label{tab:tabx-per-method-hyper-params-target}
\centering
\resizebox{.4\linewidth}{!}{
\begin{tabular}{lccc}
    Hyperparameter  &  Value  \\
    \toprule
    \multirow{3}{*}{SHOT} 
            & ${\lambda_{ent}  \in \{0.5, 1.0\}}$ \\
            & ${\lambda_{div} \in \{0.5, 1.0\}}$\\
            & ${\lambda_{peu} \in \{0.5, 1.0\}}$\\
    \hline
    \multirow{2}{*}{CDCL} 
            & ${\tau \in \{0.04, 0.08, 0.5\}}$ \\
            & ${\lambda_{cdcl} \in \{0.5, 1.0\}}$\\
    \hline
     \multirow{2}{*}{SFDE} 
            & ${\tau \in \{0.8, 1.0\}}$ \\
            & ${\lambda_{sfde} \in \{0.5, 1.0\}}$\\
    \hline
    \multirow{3}{*}{ERL} 
            & ${\tau \in \{0.04, 0.08, 0.5\}}$ \\
            & ${\lambda_{cdcl} \in \{0.5, 1.0\}}$\\
            & ${\lambda_{erl} \in \{0.5, 1.0\}}$\\
    \bottomrule
\end{tabular}
}
\end{table}

The hyperparameter search spaces reported in
Tables~\ref{tab:tabx-general-hyper-params-source},
\ref{tab:tabx-general-hyper-params}, and
\ref{tab:tabx-per-method-hyper-params-target}
correspond to the primary experimental setting reported in the main paper,
i.e., adaptation from \glas to \camsixteen and the five \camseventeen centers.
For the additional experimental scenarios reported in the supplementary material and cross-center adaptation, we use a fixed learning rate of $10^{-4}$, a learning-rate decay factor
$\gamma=0.9$, and train the adaptation model for $100$ epochs.
To limit the computational cost of the extended experiments while maintaining a
consistent model-selection protocol across methods, all method-specific loss
weights are selected from the search space $\{0.5,1.0\}$, independently of the considered SFDA method and loss term. Unless otherwise stated, all remaining architectural choices, optimization settings, and data augmentations are kept unchanged across scenarios. All experiments were conducted using NVIDIA H100 GPUs.

\subsection{Experimental Settings}
\subsubsection{SFDA Methods}
We compare our proposed method against five representative SFDA baselines covering different adaptation strategies. 

\begin{itemize}
    \item \textbf{SHOT}~\citep{liang2021shot} exploits the structure of the target domain through pseudo-label refinement, with a diversity objective (information maximization loss) that discourages predictions from collapsing toward a subset of classes.
    \item \textbf{SFDA-DE}~\citep{ding2022sourcefree} estimates source feature distributions from the source model classifier and reliable target features, and uses the surrogate source representations to reduce the source--target distribution discrepancy.
    \item \textbf{CDCL}~\citep{wang2022cross} relies essentially on confident pseudo-labels and performs cross-domain contrastive learning using class prototypes (from the source classifier), encouraging features from the same pseudo-labels to remain close while separating features associated with different classes.
    \item \textbf{ERL}~\citep{erl} exploits prediction uncertainty through entropy-based regularization to progressively refine the target model without accessing source samples.
    \item \textbf{RGV}~\citep{zhu2025revisiting} leverages a memory-based mechanism to exploit relationships among target samples during adaptation.
\end{itemize}

\subsubsection{Hyperparameter Selection and Validation Protocol}
\label{app:experimental_settings}
Hyperparameter selection constitutes a key challenge in SFDA. While traditionally SFDA is performed without any source or target annotations, benchmarking additionally requires a model-selection protocol for choosing hyperparameters such as the learning rate, number of epochs, and method-specific parameters.

Previous work has highlighted that model selection remains an unresolved
issue in domain adaptation. Different protocols are used across the literature~\citep{ericsson2023better}. Historically, oracle selection based on target performance has been used in several adaptation methods, while other works rely on source
validation data or dedicated validation sets~\citep{sfdawsol, fg_masoumeh, xu2025revisiting}. In this work, we adopt an explicit validation-assisted protocol. The target training set used by all the SFDA methods remains entirely unlabeled, and no target annotation is involved in the adaptation objective. Hyperparameters and checkpoints are instead selected using
a separate labeled validation split, while the test set remains strictly
unseen and is accessed only for the final evaluation.

This protocol is consistent with validation strategies considered in
recent SFDA methods. For example,~\citep{boudiaf2023search} explicitly use a validation set for hyperparameter and model selection when evaluating SFDA approaches. More recently,~\citep{xu2025revisiting} investigate a similar validation-assisted setting by manually annotating 300 samples from the target domain to form a small validation set for hyperparameter selection. \newline
Our protocol follows such previous work and our adaptation remains source-free and is performed only on unlabeled target training data, whereas a
separate validation set is used only to determine the experimental configuration. 

\subsection{\sfdadep Training Algorithm}
\label{app:algorithm}
\begin{algorithm}[h]
\caption{SFDA-SPF Adaptation}
\label{alg:sfdadep}

{\footnotesize
\begin{algorithmic}[1]

\Require
\parbox[t]{0.78\columnwidth}{
Unlabeled target set $\mathbb{T}$, source-pretrained WSOL model
$f^{s}$, forgetting ratio $\rho$, reconstruction period $m$,
number of epochs $E$, and loss weights
$\lambda_{\mathrm{F}},\lambda_{\mathrm{R}},\lambda_{\mathrm{L}}$
}

\Ensure Target-adapted WSOL model $f$

\State Initialize $f$ from the source-pretrained model $f^{s}$
\State Keep a frozen copy of $f^{s}$ for CAM pseudo-label generation

\For{$e=1$ to $E$}
\If{$e=1$ \textbf{or} $(e-1)\bmod m=0$}

    \ForAll{$x_i\in\mathbb{T}$}
        \State Compute prediction $p_i$, pseudo-label $\hat{y}_i$,
        and normalized entropy $H(x_i)$
    \EndFor

    \State Estimate the target prediction distribution $\hat{\pi}$
    \State Identify the dominant classes $\mathcal{C}_{\mathrm{dom}}$

    \ForAll{$k\in\mathcal{C}_{\mathrm{dom}}$}
        \State Construct the class-specific candidate set $\mathbb{B}_k$
        \State Select its top-$\rho$ highest-entropy samples
        as $\mathbb{D}_{\mathrm{F}}^{k}$
    \EndFor

    \State Construct the forget set
    $\mathbb{D}_{\mathrm{F}}
    =\bigcup_{k\in\mathcal{C}_{\mathrm{dom}}}
    \mathbb{D}_{\mathrm{F}}^{k}$

    \State Construct the retain set
    $\mathbb{D}_{\mathrm{R}}
    =\mathbb{T}\setminus\mathbb{D}_{\mathrm{F}}$

\EndIf

\ForAll{target mini-batches $\mathcal{B}\subset\mathbb{T}$}

    \State Obtain forget and retain subsets
    $\mathcal{B}_{\mathrm{F}}$ and $\mathcal{B}_{\mathrm{R}}$

    \State Compute the forgetting loss $\mathcal{L}_{\mathrm{F}}$
    on $\mathcal{B}_{\mathrm{F}}$

    \State Compute the retaining loss $\mathcal{L}_{\mathrm{R}}$
    on $\mathcal{B}_{\mathrm{R}}$

    \ForAll{$x_i\in\mathcal{B}$}
        \State Generate CAM-based partial pseudo-labels using
        the frozen source model $f^{s}$
        \State Randomly sample supervised pixel locations
        $\Omega_i^{\prime}$
    \EndFor

    \State Compute the localization loss $\mathcal{L}_{\mathrm{L}}$
    \State Compute the total adaptation loss $\mathcal{L}$
    \State Update the backbone, image-level classifier,
    and pixel-level classifier

\EndFor
\EndFor

\State \Return $f$

\end{algorithmic}
}
\end{algorithm}

\section{Additional Quantitative Results}
\label{app:additional_results}

\subsection{Cross-Organ Results}
\label{app:cross_organ}
In Tables~\ref{tab:all-second-values}, \ref{tab:c17_intercenter_src0}, and~\ref{tab:imbalance_results}, we evaluate \sfdadep under both intra-organ and inter-organ domain shifts. The intra-organ setting corresponds to breast-to-breast adaptation across different \camseventeen centers, whereas the inter-organ setting considers adaptation from colon to breast tissue. In this section, we further evaluate the reverse inter-organ scenario (breast-to-colon adaptation). Inter-organ adaptation represents a particularly challenging setting because the target domain differs from the source not only in acquisition conditions and staining characteristics, but also in tissue morphology and the histological manifestations of malignancy. For instance, the cellular and tissue structures associated with normal and cancerous regions in colorectal tissue strongly differ from those observed in breast tissue.

\begin{table*}[h]
\centering
\caption{
Reverse cross-dataset adaptation from each \camseventeen center to \glas.
Each entry reports classification accuracy/localization \pxap.
The last column reports the average across the five \camseventeen source centers.
}
\label{tab:c17_to_glas}
\resizebox{0.99\textwidth}{!}{
\begin{tabular}{lcccccc}
\toprule
Method
& \camsevenzero$\rightarrow$\glas
& \camsevenone$\rightarrow$\glas
& \camseventwo$\rightarrow$\glas
&\camseventhree$\rightarrow$\glas
& \camsevenfour$\rightarrow$\glas
& Average \\
\midrule

SHOT
& 52.50 / 62.14
& 55.00 / 59.33
& 48.75 / 56.84
& \textbf{53.75} / 66.37
& 53.75 / 56.83
& 52.75 / 60.30 \\

SFDA-DE
& \textbf{53.75} / 63.54
& 50.00 / 56.22
& 55.00 / 55.90
& \textbf{53.75} / \textbf{66.37}
& 38.75 / 59.71
& 50.25 / \textbf{60.35} \\

CDCL
& \textbf{53.75} / 63.64
& 46.25 / 52.55
& 47.50 / 55.60
& \textbf{53.75} / 64.25
& 46.25 / 52.70
& 49.50 / 57.75 \\

ERL
& \textbf{53.75} / 56.92
& 70.00 / \textbf{59.40}
& 56.25 / 56.21
& \textbf{53.75} / 43.52
& 51.25 / 60.24
& 57.00 / 55.26 \\

RGV
& \textbf{53.75} / \textbf{64.84}
& 53.75 / 50.40
& 55.00 / 55.90
& \textbf{53.75} / 63.72
& 53.75 / \textbf{60.62}
& 54.00 / 59.10 \\

\cellcolor{lightgray}\textbf{\sfdadep}
& \cellcolor{lightgray}\textbf{53.75} / 53.11
& \cellcolor{lightgray}\textbf{77.50} / 56.62
& \cellcolor{lightgray}\cellcolor{lightgray}\cellcolor{lightgray}\cellcolor{lightgray}\textbf{61.25} / \textbf{59.83}
& \cellcolor{lightgray}\cellcolor{lightgray}\cellcolor{lightgray}\textbf{53.75} / 62.04
& \cellcolor{lightgray}\cellcolor{lightgray}\textbf{55.00} / 59.87
& \cellcolor{lightgray}\textbf{60.25} / 58.29 \\

\bottomrule
\end{tabular}
}
\end{table*}

As observed in the breast-to-colon adaptation scenario, most competing methods experience a substantial performance degradation. In contrast, \sfdadep consistently achieves better classification performance when adapting from each \camseventeen center to \glas.

\subsection{Robustness to Target-Class Imbalance (Cancer scenario)}
\label{app:imbalance_cancer}

In Table~\ref{tab:imbalance_results}, we analyze the robustness of \sfdadep under different levels of target-domain class imbalance. We first consider a Normal-majority setting, which reflects a realistic histopathology scenario in which tumor regions occupy only a limited fraction of a whole-slide image (WSI), while large portions of the surrounding tissue remain non-tumorous. Consequently, patch-based sampling from such slides may yield substantially more normal than cancer samples. We also investigate the complementary Cancer-majority setting in Table~\ref{tab:imbalance_results_cancer}, which can arise when tissue sampling is concentrated within large tumor regions, resulting in only a small proportion of normal tissue patches. 


\begin{table*}[h]
\centering
\caption{
Robustness to cancer-majority class imbalance in the target adaptation (train) set on
\camseventeen.
Starting from the balanced setting ($50{:}50$), we progressively increase
the Cancer/Normal ratio up to $90{:}10$.
Results are reported as classification accuracy (\cl) / localization \pxap (\%).
Best results are shown in \textbf{bold}.
}
\label{tab:imbalance_results_cancer}

\small
\setlength{\tabcolsep}{6pt}
\renewcommand{\arraystretch}{0.5}

\begin{tabular}{lcccccc}
\toprule
\textbf{Method}
& $50{:}50$
& $60{:}40$
& $70{:}30$
& $80{:}20$
& $90{:}10$
& \textbf{Average}
\\

& \textbf{CL / PxAP}
& \textbf{CL / PxAP}
& \textbf{CL / PxAP}
& \textbf{CL / PxAP}
& \textbf{CL / PxAP}
& \textbf{CL / PxAP}
\\
\midrule

CDCL
& 50.0 / 25.3
& 44.6 / 23.4
& 48.1 / 25.3
& 46.6 / 25.2
& 46.6 / 25.5
& 47.2 / 24.9
\\

SHOT
& 70.0 / 19.2
& 41.2 / 37.2
& 41.2 / 37.2
& 41.2 / 37.2
& 41.2 / 37.2
& 47.0 / 33.6
\\

ERL
& 59.5 / 13.2
& 43.1 / 28.2
& 44.3 / 21.9
& 43.9 / 22.1
& 43.9 / 21.7
& 46.9 / 21.4
\\

RGV
& 50.0 / 31.0
& 50.0 / 34.2
& 50.0 / 34.3
& 50.0 / 34.9
& 50.0 / 35.3
& 50.0 / 33.9
\\

\cellcolor{lightgray}\textbf{\sfdadep}
& \cellcolor{lightgray}\textbf{86.2} / \textbf{49.9}
& \cellcolor{lightgray}\textbf{82.1} / \textbf{47.6}
& \cellcolor{lightgray}\textbf{83.2} / \textbf{50.9}
& \cellcolor{lightgray}\textbf{83.9} / \textbf{49.7}
& \cellcolor{lightgray}\textbf{84.7} / \textbf{50.2}
& \cellcolor{lightgray}\textbf{84.0} / \textbf{49.7}
\\

\bottomrule
\end{tabular}
\end{table*}

\begin{table}[h]
\centering
\caption{
Sensitivity to the forgetting proportion $p$ under inter-center adaptation on
\camseventeen.
The source model is trained on center~0 and independently adapted to centers~1--4.
Results are reported as classification accuracy (CL) / localization PxAP (\%).
Best results for each target are shown in \textbf{bold}.
}
\label{tab:forget_ratio_ablation}

\small
\setlength{\tabcolsep}{6pt}
\renewcommand{\arraystretch}{1.05}

\begin{tabular}{lccc}
\toprule
\textbf{Target}
& $p=0.01$
& $p=0.05$
& $p=0.10$
\\

& \textbf{CL / PxAP}
& \textbf{CL / PxAP}
& \textbf{CL / PxAP}
\\
\midrule

C17-0$\rightarrow$1
& 67.44 / 25.37
& 74.42 / 34.69
& \textbf{75.58} / \textbf{40.59}
\\

C17-0$\rightarrow$2
& 78.35 / 48.17
& \textbf{79.24} / 54.34
& 79.02 / \textbf{59.12}
\\

C17-0$\rightarrow$3
& 92.29 / \textbf{63.45}
& \textbf{93.88} / 60.44
& 93.09 / 62.39
\\

C17-0$\rightarrow$4
& \textbf{83.22} / 46.13
& 79.87 / \textbf{48.88}
& 77.85 / 47.43
\\



\bottomrule
\end{tabular}
\end{table}

\begin{table}[t]
\centering
\caption{
Sensitivity to the localization keep ratio on \camseventeen inter-center adaptation
using center~0 as source.
Results are reported as classification accuracy (\cl) / localization \pxap (\%).
}
\label{tab:keep_ratio}

\small
\setlength{\tabcolsep}{5pt}
\renewcommand{\arraystretch}{1.05}

\begin{tabular}{lccc}
\toprule
\textbf{Target}
& \multicolumn{3}{c}{\textbf{Ratio of Images for Localization}} \\
\cmidrule(lr){2-4}

& $0.2$
& $0.5$
& $1.0$
\\

& \textbf{CL / PxAP}
& \textbf{CL / PxAP}
& \textbf{CL / PxAP}
\\
\midrule

C17-1
& 68.02 / 37.95
& \textbf{75.58 / 40.59}
& 74.42 / 34.69
\\

C17-2
& 79.24 / \textbf{59.38}
& 79.02 / 53.47
& \textbf{79.24} / 54.34
\\

C17-3
& 92.29 / \textbf{63.45}
& 93.35 / 60.61
& \textbf{93.88} / 60.44
\\

C17-4
& \textbf{83.22} / 46.13
& 78.52 / 38.03
& 79.87 / \textbf{48.88}
\\



\bottomrule
\end{tabular}
\end{table}

\begin{table}[h]
\centering
\caption{Impact of jointly adapting the backbone and image-level classifier.
Results are reported on \glas $\rightarrow$ \camsevenzero with \pixelcam. The source model used for CAM-based pixel pseudo-label generation remains frozen in all configurations.}

\label{tab:freezing_ablation}
\setlength{\tabcolsep}{2pt}
\renewcommand{\arraystretch}{1.3}

\begin{tabular}{cc|cc}
\toprule
Backbone &
Classifier &
PxAP $\uparrow$ &
CL $\uparrow$ \\
\midrule
Frozen    & Trainable & \underline{31.4} & 47.7 \\
Trainable & Frozen    & \underline{31.4} & \underline{85.8} \\
Trainable & Trainable & \textbf{49.9} & \textbf{86.1} \\
\bottomrule
\end{tabular}
\end{table}

Many SFDA approaches freeze the source classifier and adapt only the feature extractor, based on the assumption that the classifier preserves a class-discriminative structure learned from the source domain that remains transferable to the target domain. We examine whether this assumption remains valid under a strong cross-organ shift. Table~\ref{tab:freezing_ablation} reports the effect of adapting the backbone and image-level classifier independently or jointly. Updating only the classifier while keeping the backbone frozen yields 47.7 \cl and 31.4 \pxap, indicating that adjusting the decision layer alone cannot compensate for the mismatch between source and target image representations. In contrast, adapting the backbone while freezing the classifier improves classification accuracy to 85.8 \cl, showing that feature adaptation plays an important role in recovering target-domain class discrimination. However, localization performance remains limited to 31.4 \pxap. Jointly adapting the backbone and classifier increases \pxap to 49.9, while improving classification accuracy of 86.1 \cl.

\subsection{CNNs versus transformer-based models.} 
\label{app:cnn_transformer}
\begin{wrapfigure}{r}{0.53\columnwidth}
    \centering
    \vspace{-30pt}
    \includegraphics[width=0.9\linewidth]
    {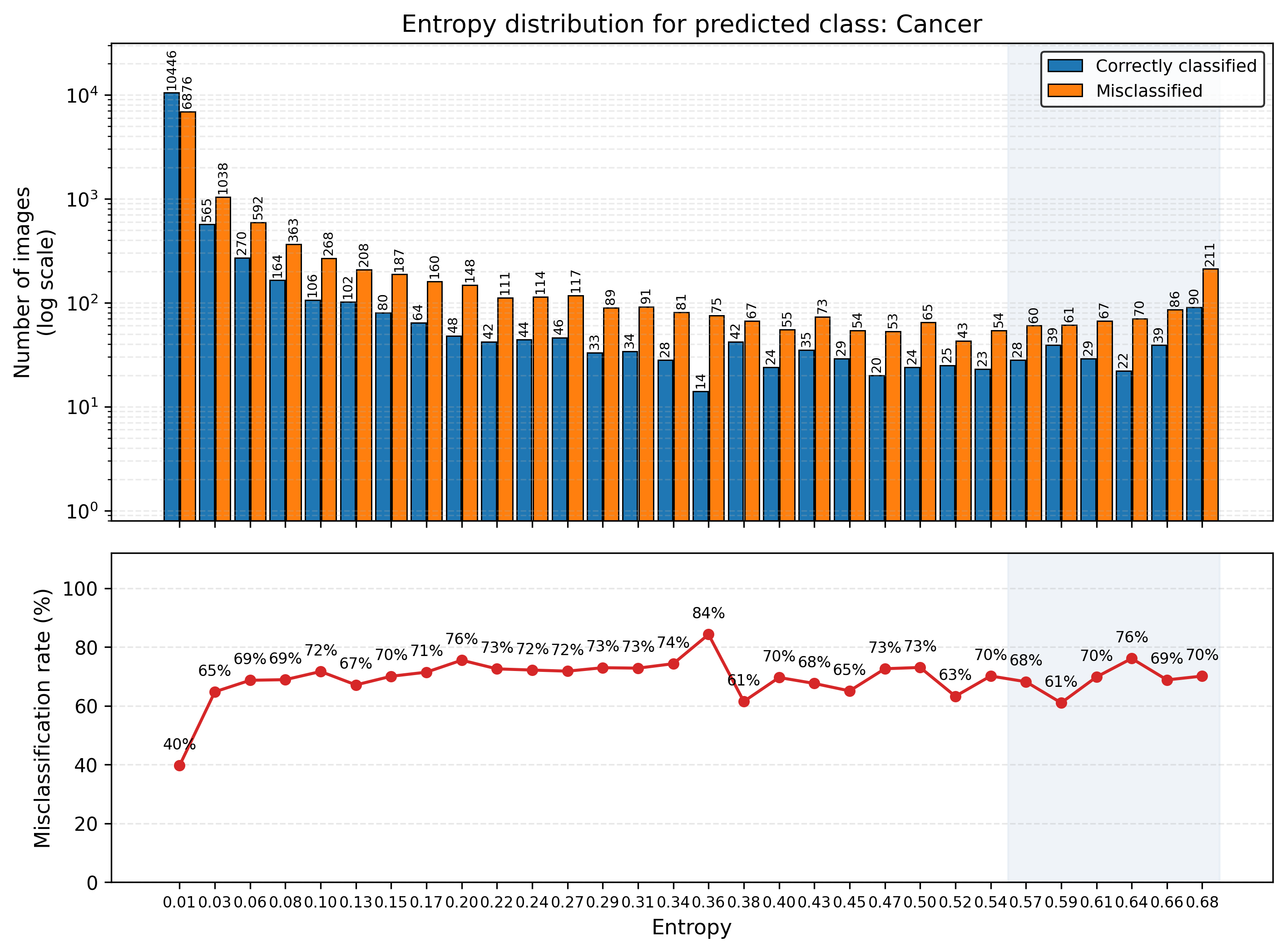}
    \captionsetup{font=small}
    \caption{Entropy distributions of correctly and incorrectly classified
    samples predicted as cancer by SAT.}
    \label{fig:entropy_sat}
\end{wrapfigure}

Figure~\ref{fig:entropy_sat} provides an explanation for the more modest gains observed with transformer-based WSOL models compared with CNN-based models. Unlike \pixelcam, for which uncertain dominant-class predictions are enriched in misclassification errors, \sat produces a highly concentrated low-entropy distribution. 

Consequently, many erroneous predictions remain highly confident and are assigned to the retain set, making entropy less effective at separating reliable from incorrect predictions. Despite this less favorable confidence structure, \sfdadep improves \sat from 21.3 to 30.3 \pxap and from 52.1 to 69.2 \cl on average. Its benefit is particularly pronounced for localization, exceeding the strongest competing SFDA method by 7.9 \pxap points.

\FloatBarrier

\subsection{Extension to the Multi-Class Setting}
\label{app:multi_class}
\begin{table*}[h] 
\caption{ 
Localization (\pxap) and classification (F1) performance on \ebhitarget under controlled domain shifts.
The source model is trained on the clean source domain and independently adapted to each target condition. 
For all methods, results are reported at the checkpoint selected according to the best F1 score. 
Best results are shown in \textbf{bold} and second-best results are \underline{underlined}. 
} 
\label{tab:ebhi_acc}
\centering 
 
\resizebox{\textwidth}{!}{ 
\scriptsize 
\begin{tabular}{|l 
| cc | cc | cc | cc 
|| cc|} 

\cline{1-11} 

& \multicolumn{2}{c|}{SHOT {\scriptsize (ICML'20)}} 
& \multicolumn{2}{c|}{CDCL {\scriptsize (TMM'22)}} 
& \multicolumn{2}{c|}{ERL {\scriptsize (ICLR'23)}} 
& \multicolumn{2}{c||}{RGV {\scriptsize (CVPR'25)}} 
& \multicolumn{2}{c|}{\textbf{Ours}} 
\\ 
 
\cline{2-11} 
 
\textbf{Target condition} 
& \pxap & F1 
& \pxap & F1 
& \pxap & F1 
& \pxap & F1 
& \cellcolor{lightgray}\pxap & \cellcolor{lightgray}F1 
\\ 
 
\hline 
 
Target clean 
& \underline{83.7} & \underline{77.2} 
& \underline{83.7} & \underline{77.2} 
& \underline{83.7} & \underline{77.2} 
& \underline{83.7} & \underline{77.2} 
& \cellcolor{lightgray}\textbf{89.3} 
& \cellcolor{lightgray}\textbf{82.3} 
\\ 
 
Stain-light 
& 83.5 & \textbf{75.1} 
& \underline{83.8} & 67.1 
& 83.7 & 69.9 
& \underline{83.8} & 67.1 
& \cellcolor{lightgray}\textbf{90.9} 
& \cellcolor{lightgray}\textbf{75.1} 
\\ 
 
Stain-heavy 
& \textbf{86.7} & \textbf{70.3} 
& 81.3 & 55.9 
& 70.5 & 56.7 
& 80.8 & 43.9 
& \cellcolor{lightgray}\underline{82.7} 
& \cellcolor{lightgray}\underline{57.9} 
\\ 
 
Dust 
& 82.9 & \textbf{79.4} 
& \underline{83.1} & 71.0 
& \underline{83.1} & 71.0 
& \underline{83.1} & 71.0 
& \cellcolor{lightgray}\textbf{86.7} 
& \cellcolor{lightgray}\underline{72.4} 
\\ 
 
Air bubble 
& \underline{77.8} & \textbf{78.1} 
& 76.8 & 45.3 
& 76.8 & 45.3 
& 76.8 & 45.3 
& \cellcolor{lightgray}\textbf{78.4} 
& \cellcolor{lightgray}\underline{66.5} 
\\ 
 
Defocus blur 
& \textbf{73.3} & \textbf{42.2} 
& \underline{71.9} & \underline{17.4} 
& \underline{71.9} & \underline{17.4} 
& \underline{71.9} & \underline{17.4} 
& \cellcolor{lightgray} 76.5 
& \cellcolor{lightgray} 21.6
\\ 
 
Motion blur 
& \underline{70.7} & \underline{21.1} 
& 67.9 & 19.4 
& 67.9 & 19.4 
& 67.9 & 19.4 
& \cellcolor{lightgray}\textbf{72.9} 
& \cellcolor{lightgray}\textbf{29.3} 
\\ 
 
Gaussian noise 
& 82.9 & \textbf{60.9} 
& \underline{84.1} & 45.5 
& 77.1 & 53.1 
& \underline{84.1} & 45.5 
& \cellcolor{lightgray}\textbf{88.2} 
& \cellcolor{lightgray}\underline{57.2} 
\\ 
 
Shot noise 
& \underline{84.3} & 46.5 
& 79.1 & 50.5 
& 75.3 & \textbf{54.5} 
& \underline{84.3} & 46.5 
& \cellcolor{lightgray}\textbf{86.8} 
& \cellcolor{lightgray}\underline{52.6} 
\\ 
 
Brightness 
& \underline{81.1} & 70.9 
& \underline{81.1} & 70.9 
& 78.0 & \underline{72.5} 
& \underline{81.1} & 70.9 
& \cellcolor{lightgray}\textbf{84.6} 
& \cellcolor{lightgray}\textbf{79.4} 
\\ 
 
Contrast 
& \textbf{71.1} & \textbf{30.8} 
& \underline{63.4} & 9.1 
& \underline{63.4} & 9.1 
& \underline{63.4} & 9.1 
& \cellcolor{lightgray}62.6 
& \cellcolor{lightgray}\underline{12.1} 
\\ 
 
\hline 
\hline 
 
\textbf{Average} 
& 79.8 & \textbf{59.3} 
& 77.8 & 48.1 
& 75.6 & 49.7 
& 78.3 & 46.7 
& \cellcolor{lightgray}\textbf{82.3} 
& \cellcolor{lightgray}\underline{58.5} 
\\ 
 
\hline 
 
\end{tabular} 
} 
 
\label{tab:ebhi_corruptions_f1} 
\end{table*}

\begin{table}[h]
\centering
\caption{
Per-class source-model predictions under representative target shifts on the EBHI training set.
For each class, we report the ground-truth distribution (GT),
per-class recall (Rec.), precision (Prec.), F1-score, and predicted
class distribution (Pred.). Target Clean and Brightness represent
conditions where the source model preserves predictions across classes,
whereas Defocus Blur and Contrast illustrate severe prediction collapse.
}
\label{tab:ebhi_source_distribution}

\resizebox{\textwidth}{!}{
\begin{tabular}{|lcc
| cccc
| cccc
| cccc
| cccc|}
\hline

&
&
&
\multicolumn{4}{c|}{\textbf{Target Clean (Shift 1)}} &
\multicolumn{4}{c|}{\textbf{Brightness (Shift 10)}} &
\multicolumn{4}{c|}{\textbf{Defocus Blur (Shift 6)}} &
\multicolumn{4}{c|}{\textbf{Contrast (Shift 11)}} \\

\cline{4-19}

\textbf{Class} &
\textbf{GT} &
\textbf{GT (\%)} &
\textbf{Rec.} &
\textbf{Prec.} &
\textbf{F1} &
\textbf{Pred.} &
\textbf{Rec.} &
\textbf{Prec.} &
\textbf{F1} &
\textbf{Pred.} &
\textbf{Rec.} &
\textbf{Prec.} &
\textbf{F1} &
\textbf{Pred.} &
\textbf{Rec.} &
\textbf{Prec.} &
\textbf{F1} &
\textbf{Pred.} \\

\hline

Normal &
22 &
3.3 &
13.6 &
100.0 &
24.0 &
3 (0.5\%) &
18.2 &
80.0 &
29.6 &
5 (0.8\%) &
0.0 &
0.0 &
0.0 &
0 (0.0\%) &
0.0 &
0.0 &
0.0 &
0 (0.0\%) \\

Polyp &
143 &
21.4 &
94.4 &
83.9 &
88.8 &
161 (24.1\%) &
95.1 &
81.0 &
87.5 &
168 (25.2\%) &
10.5 &
65.2 &
18.1 &
23 (3.4\%) &
6.3 &
81.8 &
11.7 &
11 (1.6\%) \\

Low-grade IN &
191 &
28.6 &
91.1 &
91.1 &
91.1 &
191 (28.6\%) &
81.2 &
92.8 &
86.6 &
167 (25.0\%) &
0.0 &
0.0 &
0.0 &
0 (0.0\%) &
0.0 &
0.0 &
0.0 &
0 (0.0\%) \\

High-grade IN &
55 &
8.2 &
69.1 &
73.1 &
71.0 &
52 (7.8\%) &
69.1 &
64.4 &
66.7 &
59 (8.8\%) &
90.9 &
9.4 &
17.0 &
535 (\textbf{80.2\%}) &
98.2 &
8.8 &
16.1 &
614 (\textbf{92.1\%}) \\

Serrated adenoma &
17 &
2.5 &
94.1 &
94.1 &
94.1 &
17 (2.5\%) &
94.1 &
94.1 &
94.1 &
17 (2.5\%) &
5.9 &
100.0 &
11.1 &
1 (0.1\%) &
0.0 &
0.0 &
0.0 &
0 (0.0\%) \\

Adenocarcinoma &
239 &
35.8 &
96.2 &
94.7 &
95.4 &
243 (36.4\%) &
96.7 &
92.0 &
94.3 &
251 (37.6\%) &
25.1 &
55.6 &
34.6 &
108 (16.2\%) &
14.2 &
81.0 &
24.2 &
42 (6.3\%) \\

\hline
\end{tabular}
}
\end{table}

In the 2-class setting, \sfdadep achieves the best localization and classification performance in the majority of domain shifts, supporting its effectiveness. In a more challenging multi-class setting with class imbalance, \sfdadep remains competitive in classification, outperforming CDCL, ERL, and RGV on most shifts. It achieves particularly large gains on Motion Blur and Brightness cases. Nevertheless, all SFDA methods experience degradation under some important shifts, such as Dust, Air Bubble, Defocus Blur, Noise, and Contrast, where SHOT can outperform our approach. This limitation is explained by the combination of severe class imbalance and multiclass similarity. 

Table~\ref{tab:ebhi_source_distribution} provides insight into the shift-dependent behavior observed in Table~\ref{tab:ebhi_acc}. For Target Clean and Brightness, where \sfdadep performs particularly well, the source model preserves meaningful predictions across all six classes. In contrast, Defocus Blur and Contrast induce severe prediction collapse: although High-grade IN represents only 8.2\% of the target samples, it accounts for 80.2\% and 92.1\% of the source-model predictions, respectively, while some classes receive no predictions at all. Under such extreme collapse, identifying samples to forget is only part of the problem. Unlike in the 2-class setting, reconsidering a dominant prediction leaves several possible alternative classes, some of which are absent from the initial predictions. This extreme prediction collapse leaves very limited information for recovering classes that are entirely absent from the initial predictions, making adaptation particularly difficult for all evaluated SFDA methods. In this specific scenario, SHOT can occasionally benefit from its diversity objective, which encourages predictions across multiple classes and may help recover initially unrepresented classes. This advantage is, however, specific to some severe multi-class collapse cases and is not consistently
observed across the evaluated shifts.

\FloatBarrier


\section{Reliability of Source CAMs for Pixel Pseudo-Labeling}
\label{app:source_cams}





\subsection{Reliability of High-Confidence Foreground Regions}
\label{app:foreground_reliability}
\begin{figure*}[!htbp]
\centering
\includegraphics[
    width=0.99\linewidth
]{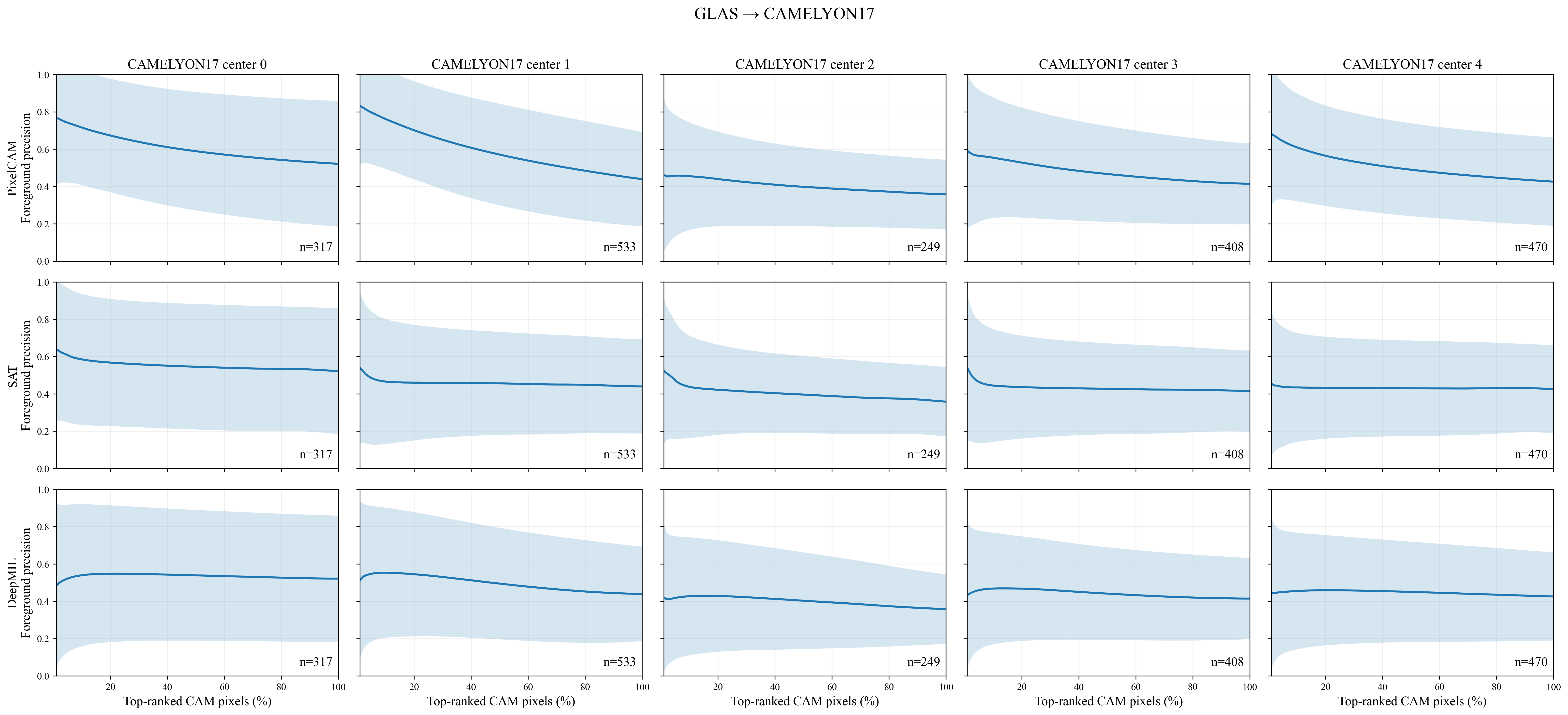}
\caption{
Reliability of high-confidence foreground regions for
\glas $\rightarrow$ \camseventeen.
For each positive target image, CAM pixels are ranked according to their
activation score, and foreground precision is computed as the fraction
of top-ranked pixels overlapping the ground-truth foreground mask.
Results are reported for PixelCAM, SAT, and DeepMIL across the five
\camseventeen centers.
}
\label{fig:fg_precision_glas_c17}
\end{figure*}

\begin{figure*}[!htbp]
\centering
\includegraphics[
    width=0.99\linewidth
]{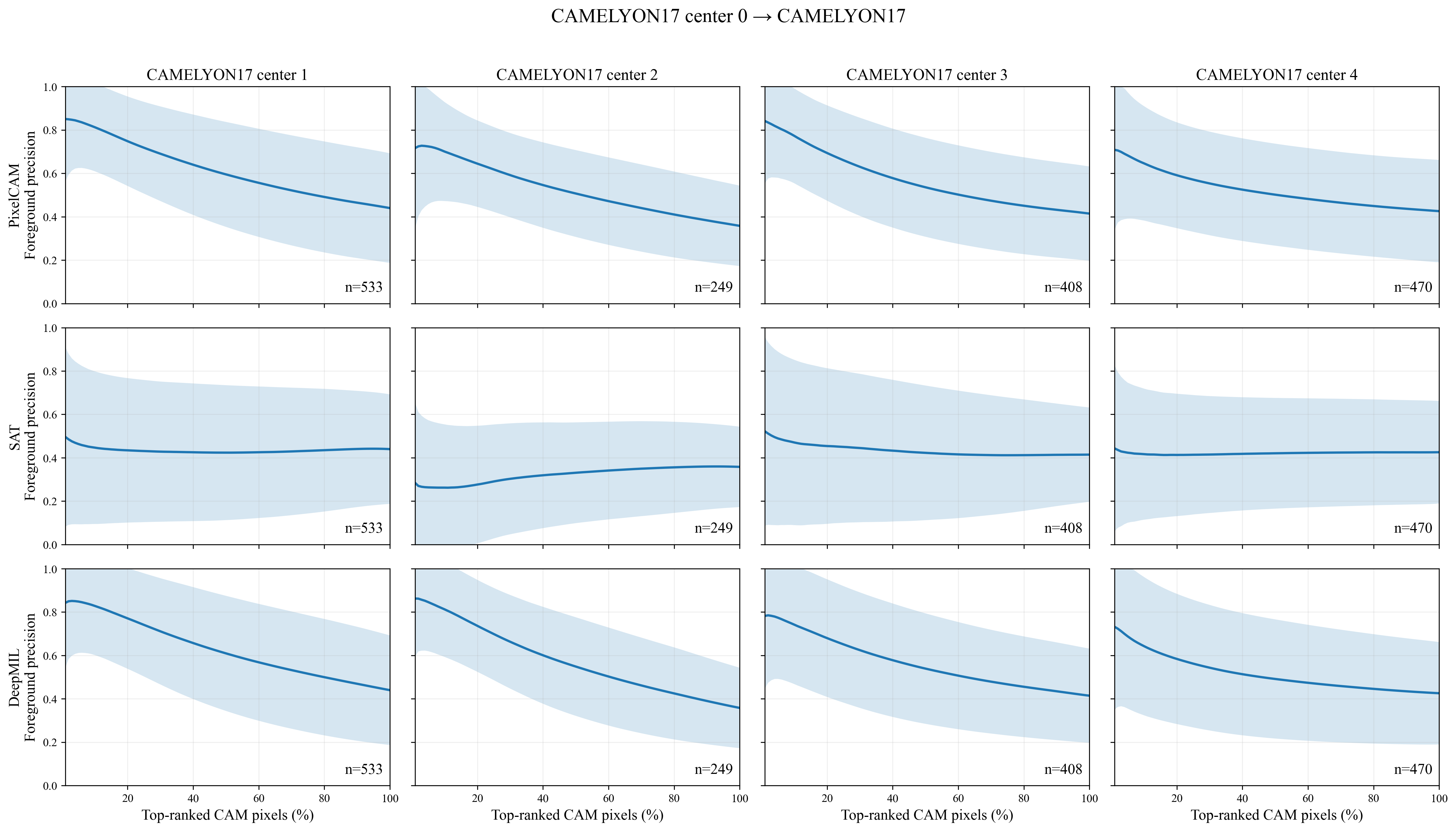}
\caption{
Reliability of high-confidence foreground regions under
\camseventeen inter-center domain shifts using center~0 as the source.
The source models are evaluated on centers~1--4.
Results are reported for PixelCAM, SAT, and DeepMIL.
}
\label{fig:fg_precision_c17_0}
\end{figure*}

\begin{figure*}[!htbp]
\centering
\includegraphics[
    width=0.99\linewidth
]{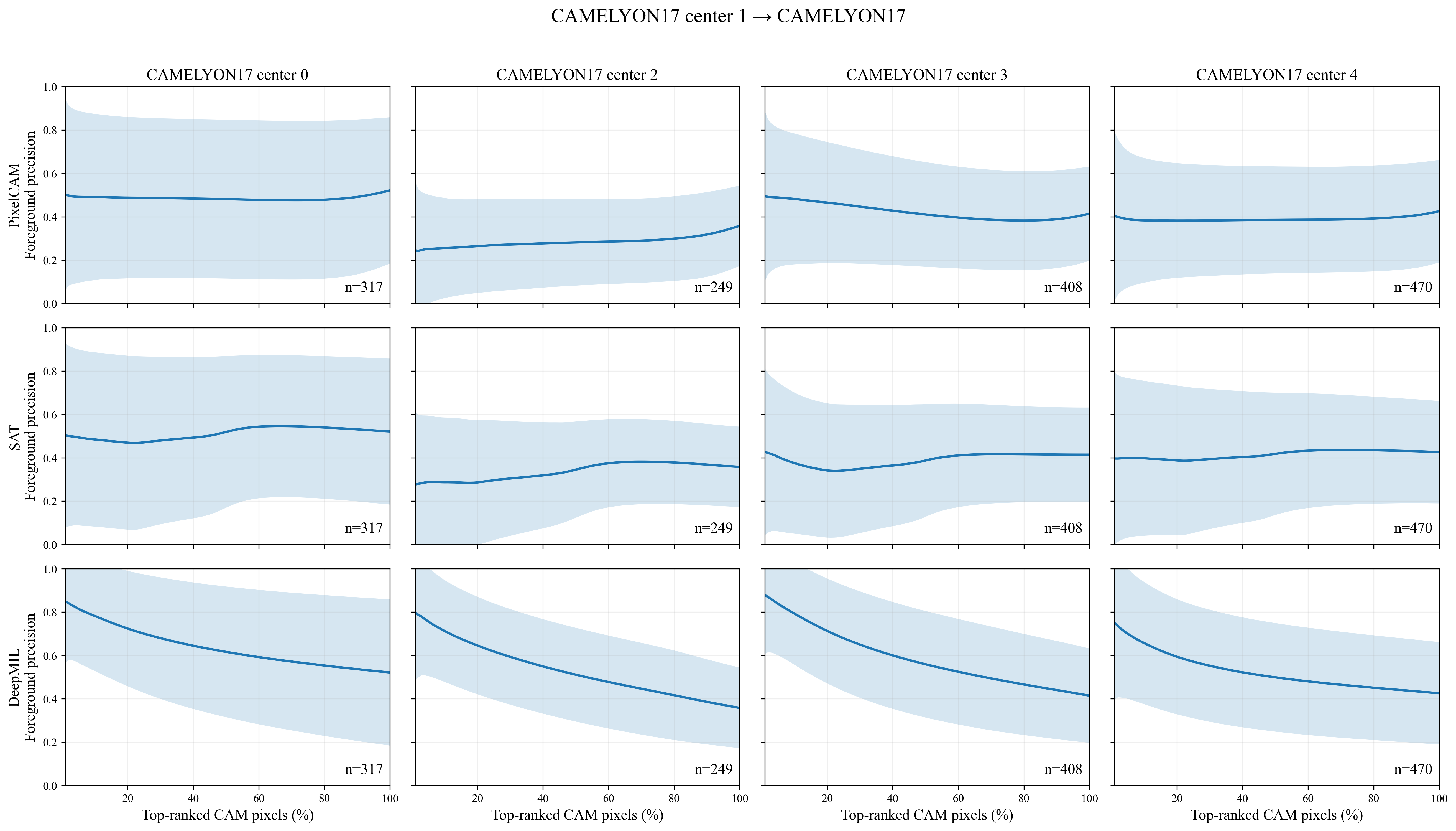}
\caption{
Reliability of high-confidence foreground regions under
\camseventeen inter-center domain shifts using center~1 as the source.
The source models are evaluated on centers~0, 2, 3, and~4.
Results are reported for PixelCAM, SAT, and DeepMIL.
}
\label{fig:fg_precision_c17_1}
\end{figure*}

\begin{figure*}[!htbp]
\centering
\includegraphics[
    width=0.99\linewidth
]{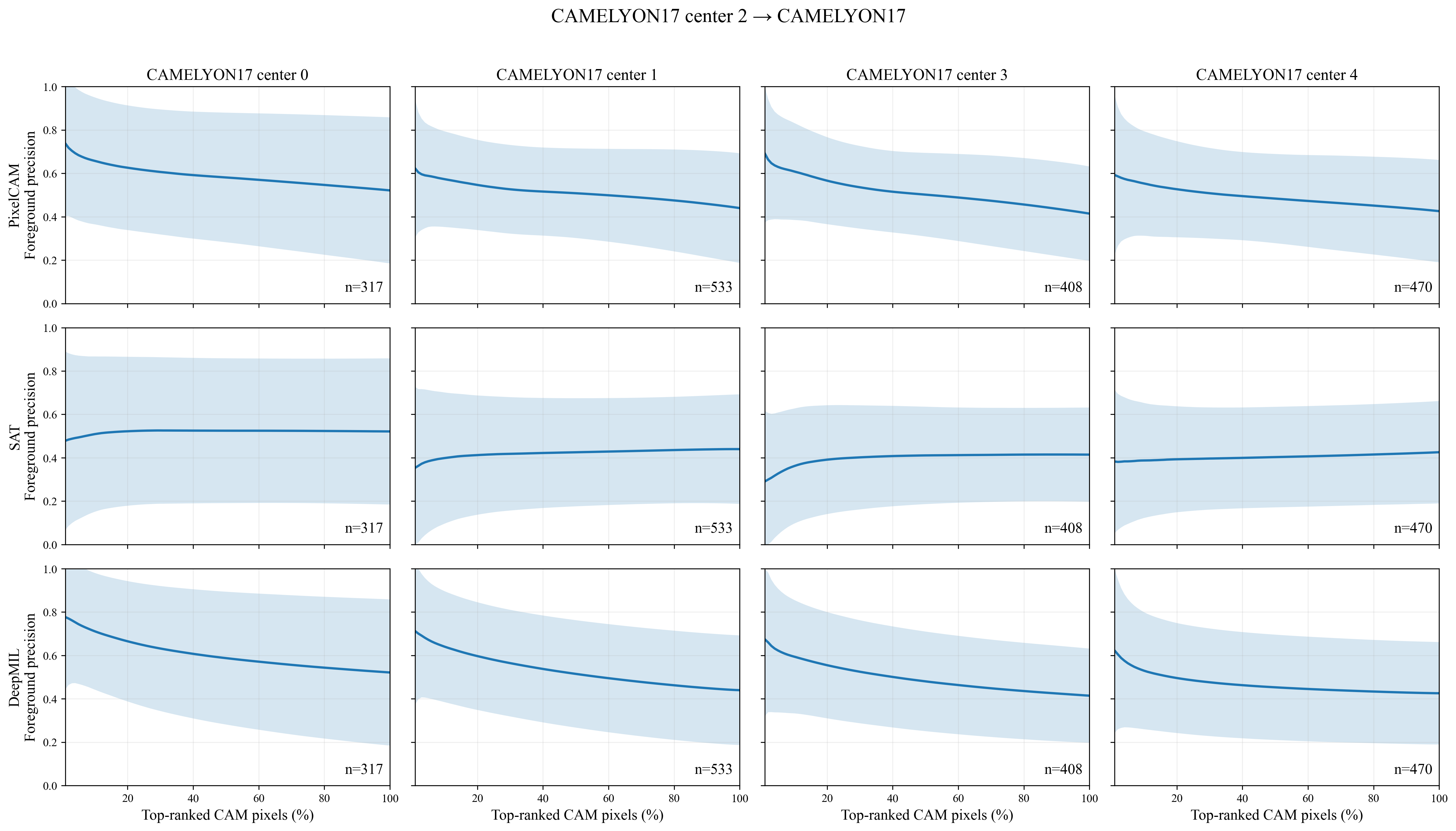}
\caption{
Reliability of high-confidence foreground regions under
\camseventeen inter-center domain shifts using center~2 as the source.
The source models are evaluated on centers~0, 1, 3, and~4.
Results are reported for PixelCAM, SAT, and DeepMIL.
}
\label{fig:fg_precision_c17_2}
\end{figure*}

\begin{figure*}[!htbp]
\centering
\includegraphics[
    width=0.99\linewidth
]{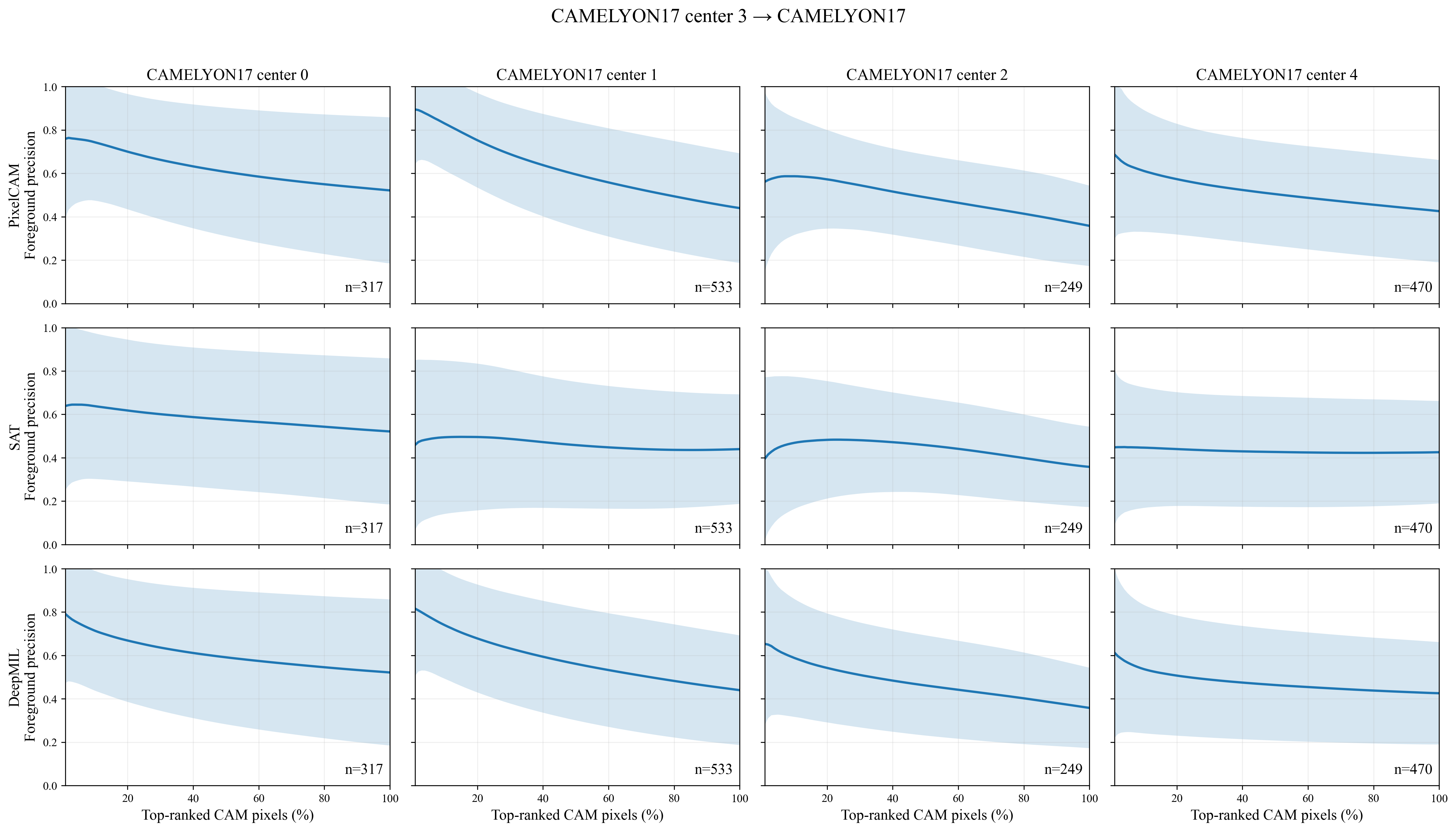}
\caption{
Reliability of high-confidence foreground regions under
\camseventeen inter-center domain shifts using center~3 as the source.
The source models are evaluated on centers~0, 1, 2, and~4.
Results are reported for PixelCAM, SAT, and DeepMIL.
}
\label{fig:fg_precision_c17_3}
\end{figure*}

\begin{figure*}[!htbp]
\centering
\includegraphics[
    width=0.99\linewidth
]{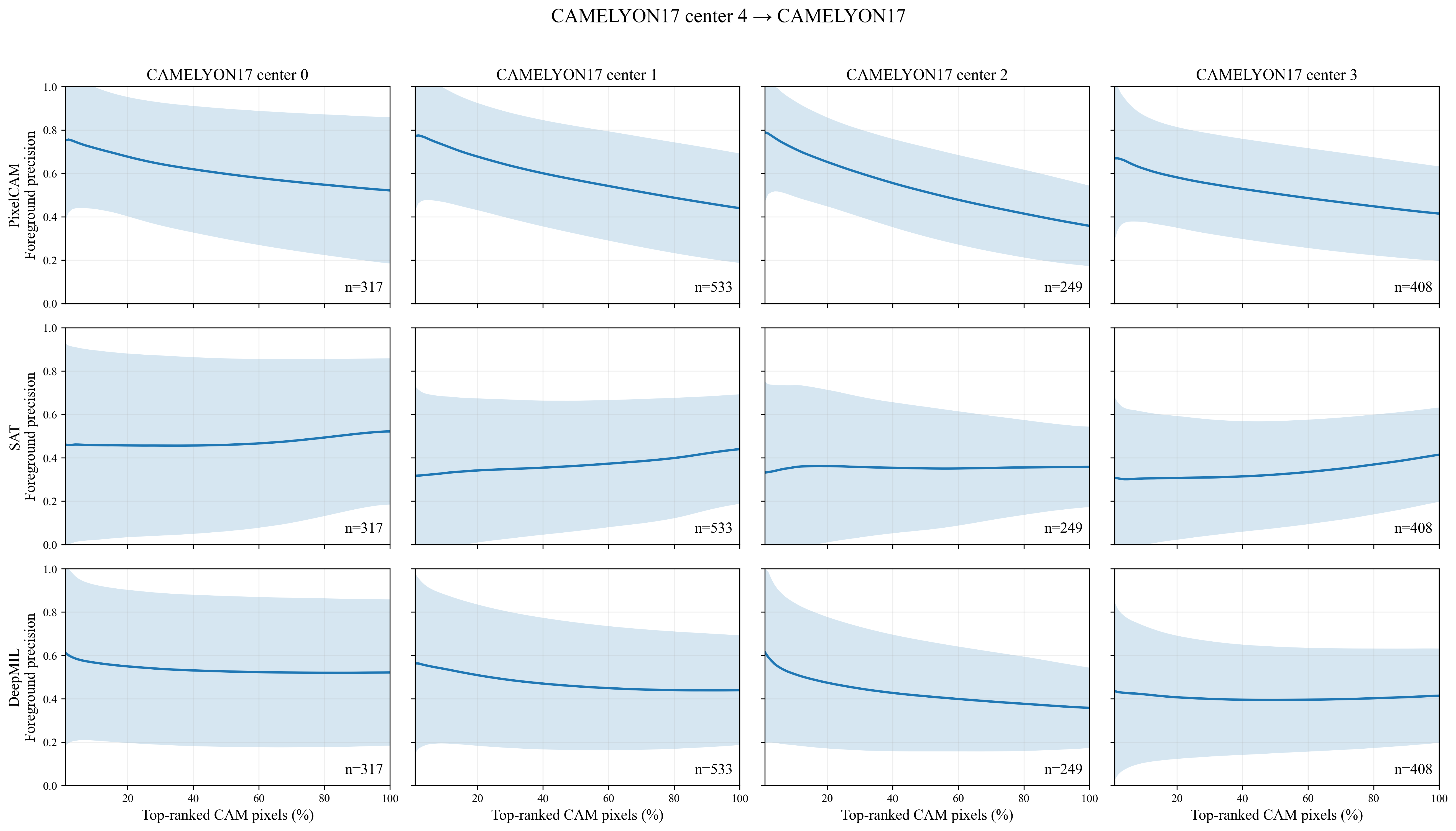}
\caption{
Reliability of high-confidence foreground regions under
\camseventeen inter-center domain shifts using center~4 as the source.
The source models are evaluated on centers~0--3.
Results are reported for PixelCAM, SAT, and DeepMIL.
}
\label{fig:fg_precision_c17_4}
\end{figure*}

\FloatBarrier
\subsection{Reliability of Low-Activation Background Regions}
\label{app:background_reliability}
\begin{figure*}[!htbp]
\centering
\includegraphics[
    width=0.99\linewidth
]{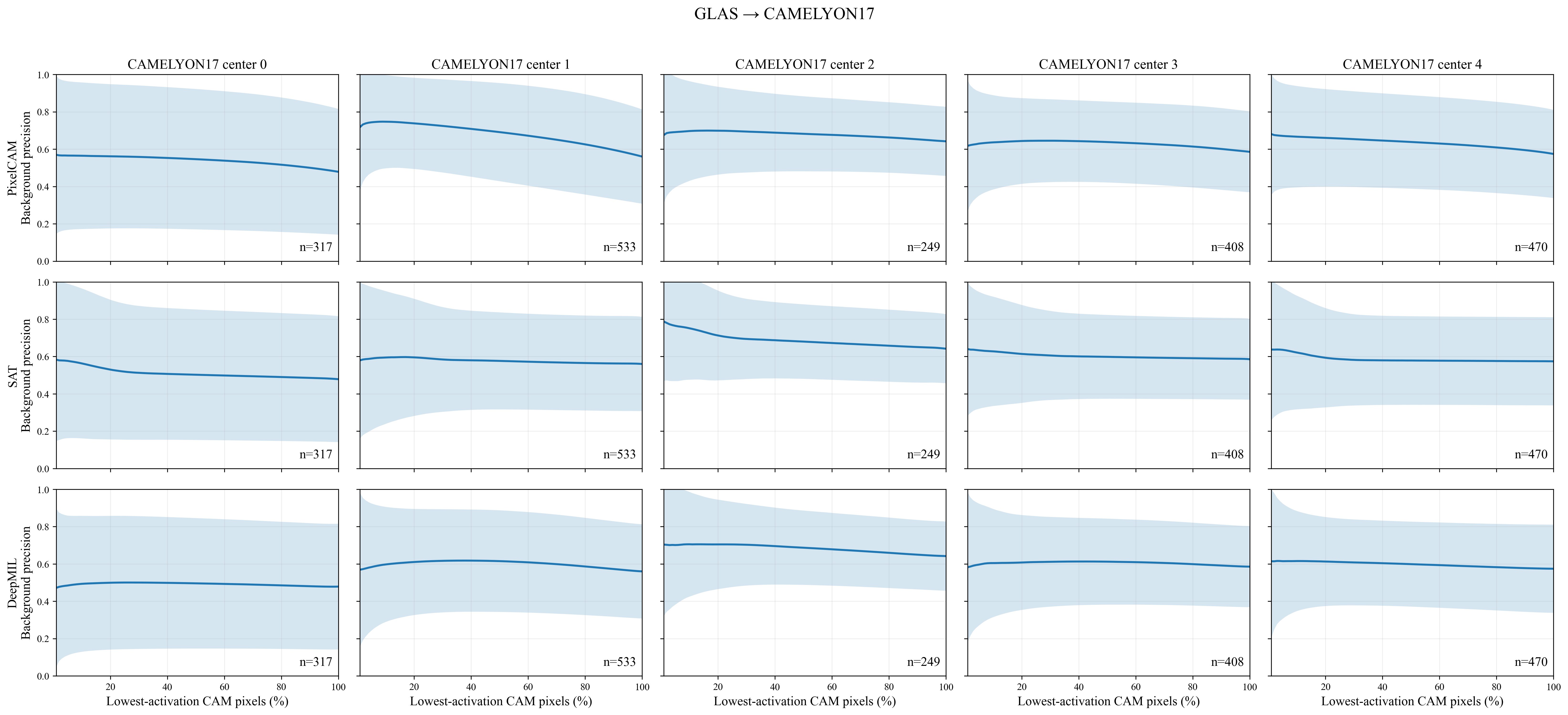}
\caption{
Reliability of low-activation background regions for
\glas $\rightarrow$ \camseventeen.
For each positive target image, CAM pixels are ranked according to their
activation score, and background precision is computed as the fraction
of the lowest-activation pixels overlapping the ground-truth background region.
Results are reported for PixelCAM, SAT, and DeepMIL across the five
\camseventeen centers.
}
\label{fig:bg_precision_glas_c17}
\end{figure*}

\begin{figure*}[!htbp]
\centering
\includegraphics[
    width=0.99\linewidth
]{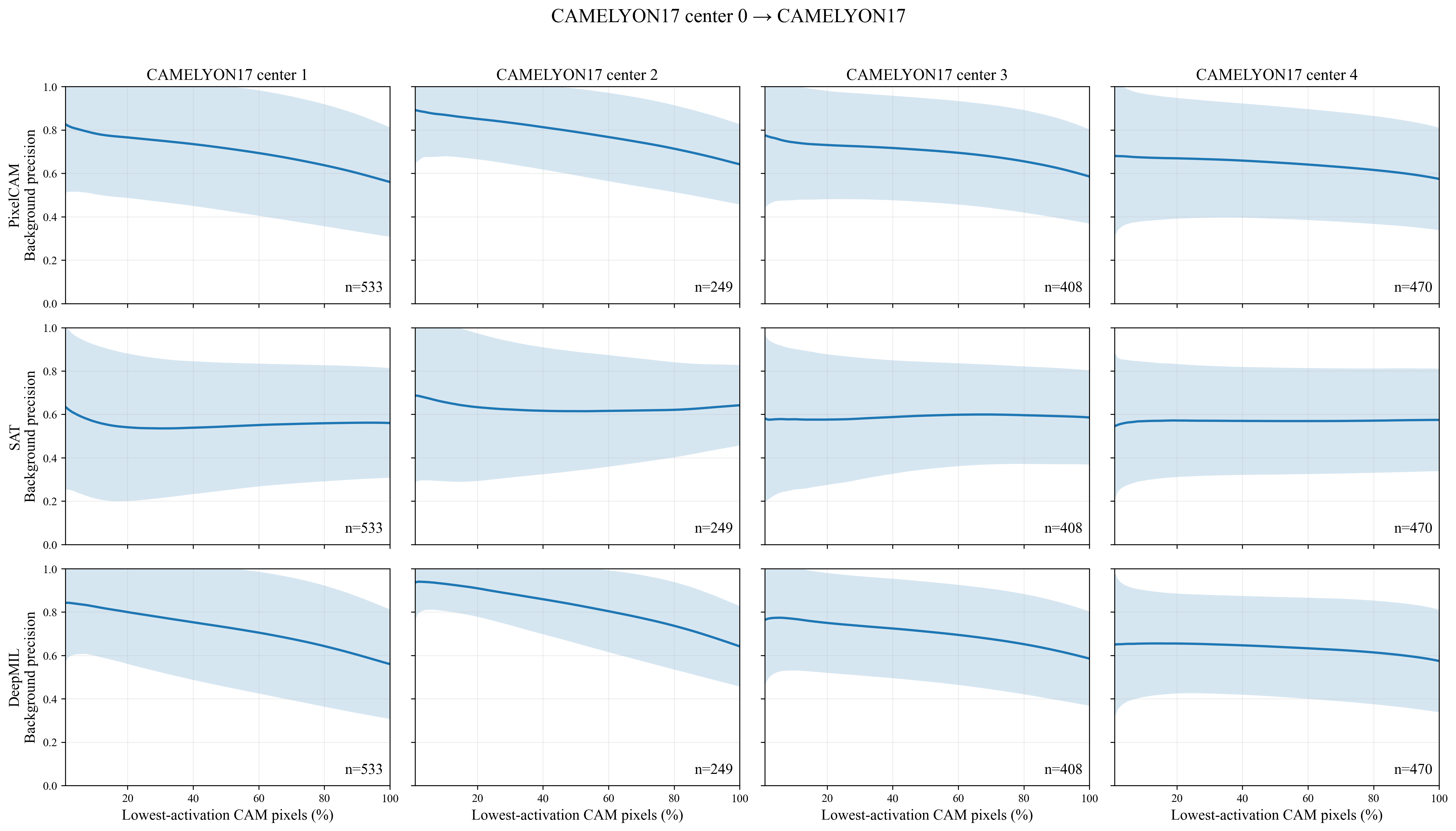}
\caption{
Reliability of low-activation background regions under
\camseventeen inter-center domain shifts using center~0 as the source.
The source models are evaluated on centers~1--4.
Results are reported for PixelCAM, SAT, and DeepMIL.
}
\label{fig:bg_precision_c17_0}
\end{figure*}

\begin{figure*}[!htbp]
\centering
\includegraphics[
    width=0.99\linewidth
]{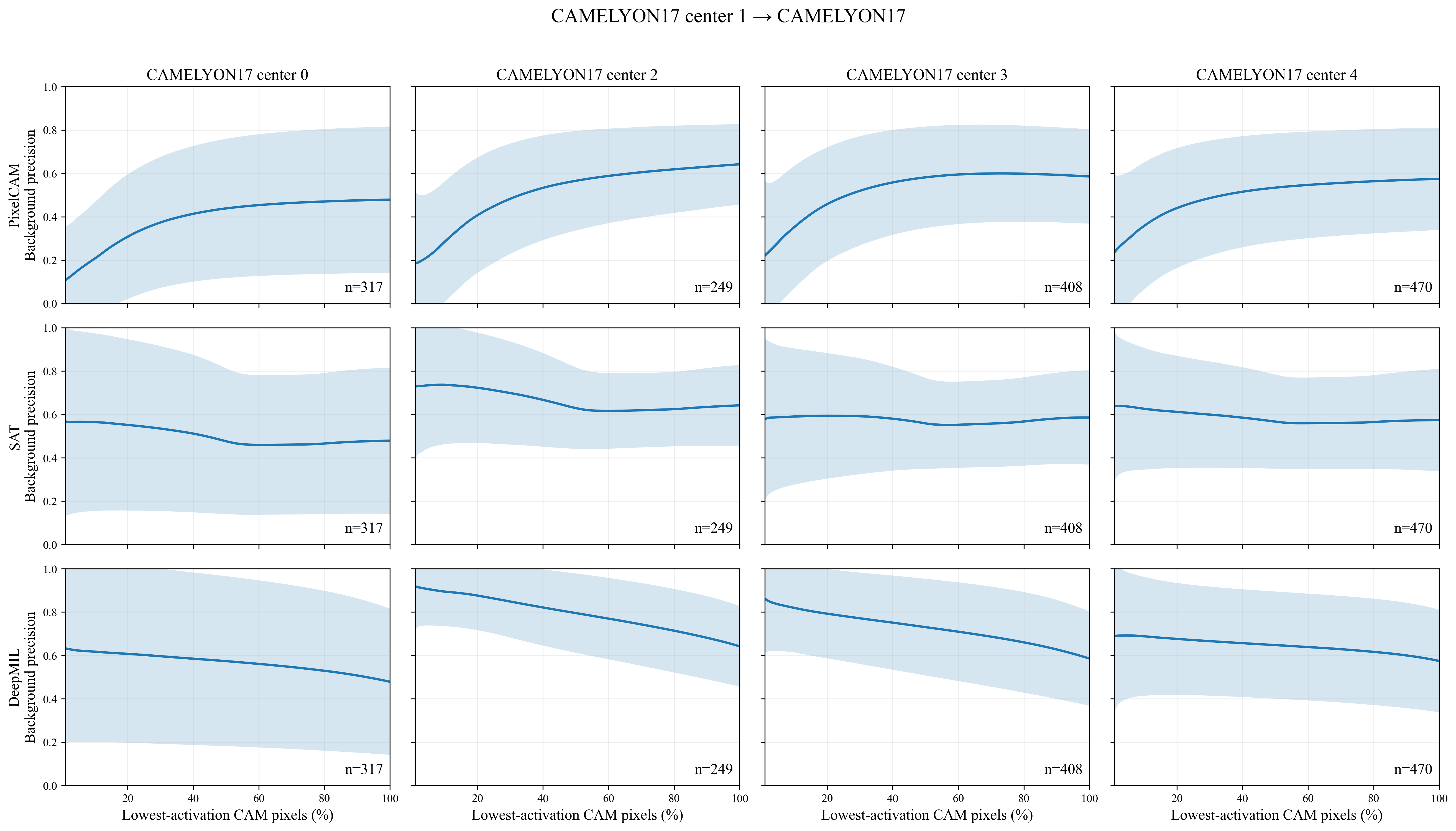}
\caption{
Reliability of low-activation background regions under
\camseventeen inter-center domain shifts using center~1 as the source.
The source models are evaluated on centers~0, 2, 3, and~4.
Results are reported for PixelCAM, SAT, and DeepMIL.
}
\label{fig:bg_precision_c17_1}
\end{figure*}

\begin{figure*}[!htbp]
\centering
\includegraphics[
    width=0.99\linewidth
]{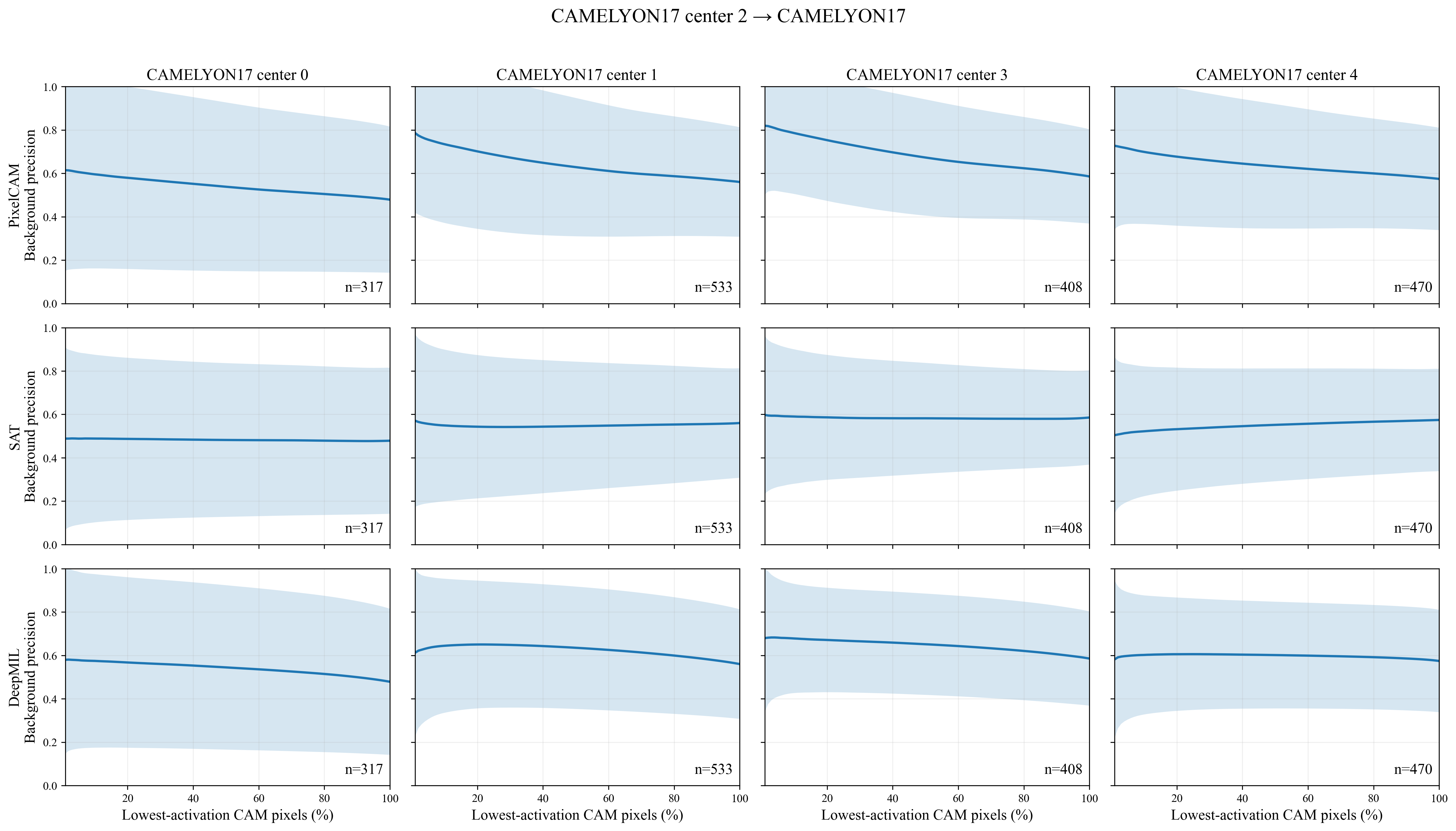}
\caption{
Reliability of low-activation background regions under
\camseventeen inter-center domain shifts using center~2 as the source.
The source models are evaluated on centers~0, 1, 3, and~4.
Results are reported for PixelCAM, SAT, and DeepMIL.
}
\label{fig:bg_precision_c17_2}
\end{figure*}


\section{Additional Visualizations}
\label{app:visualizations}
\begin{figure*}[!t]
    \centering
    \includegraphics[
        width=0.99\linewidth
    ]{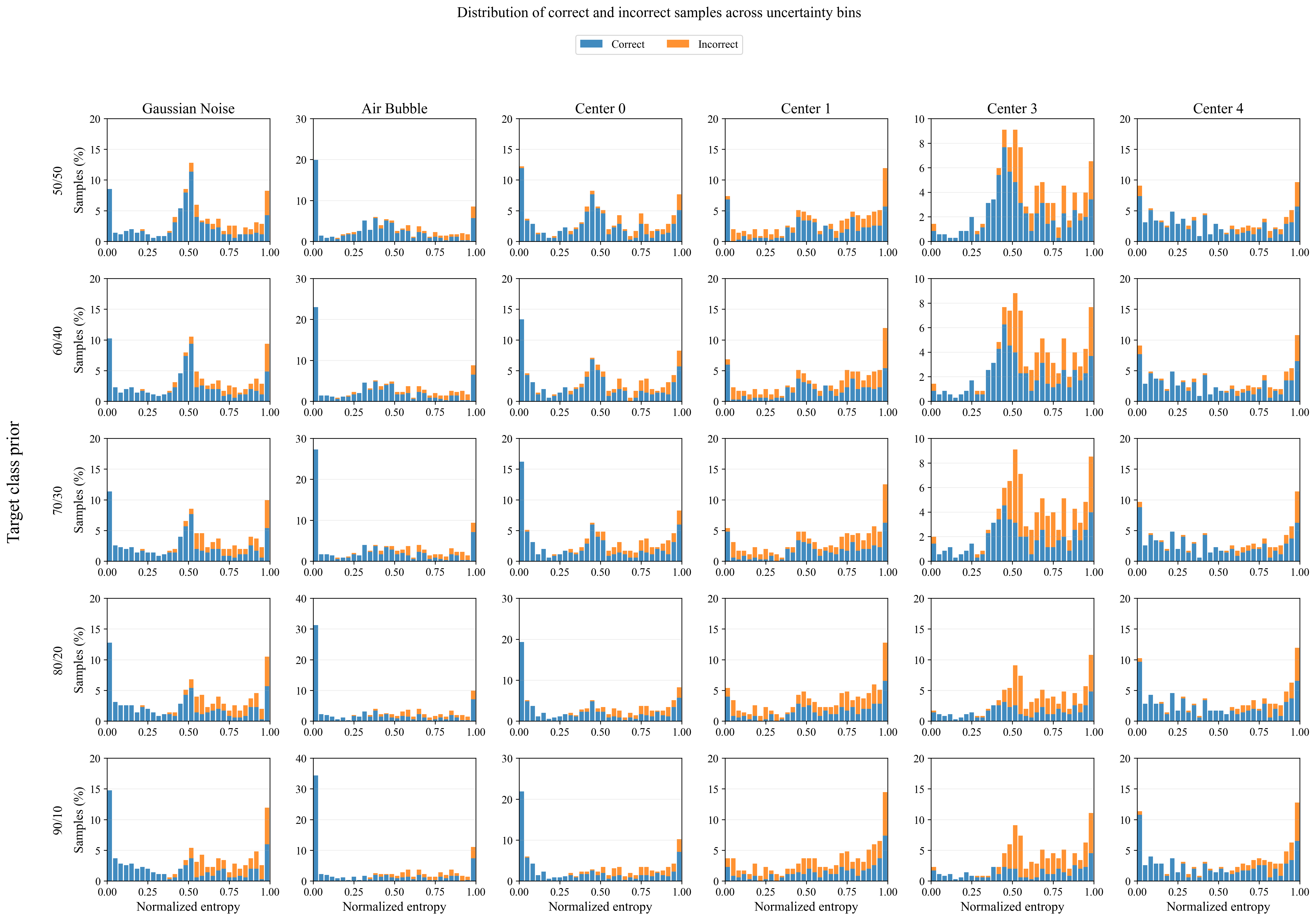}
    \caption{
   Distribution of correct and incorrect predictions across predictive-entropy bins under different domain shifts and target class priors.
    Results are shown for \pixelcam. Bars report the percentage of target samples within each normalized entropy bin,
    decomposed into correctly (blue) and incorrectly (orange) classified samples.
    }
    \label{fig:entropy_distribution_pixelcam}
\end{figure*}

\sfdadep motivation is that uncertain predictions within over-predicted classes are particularly informative for identifying potentially erroneous target predictions. Rather than relying exclusively on confident pseudo-labels,
\sfdadep explicitly identifies uncertain samples from over-predicted classes and encourages the model to forget these predictions. This mechanism allows the decision boundary to progressively move away from source predictions and better separate normal and cancer samples in the target
domain. To empirically support this motivation, Figure~\ref{fig:entropy_distribution_pixelcam} analyzes the distribution of correct and incorrect predictions across predictive-entropy levels under different domain
shifts and target class priors.
\subsection{Entropy Distributions}
\label{app:entropy_distributions}

\subsection{Entropy--Accuracy Relationships}
\label{app:entropy_accuracy}
\begin{figure*}[!t]
    \centering
    \includegraphics[
        width=0.99\linewidth
    ]{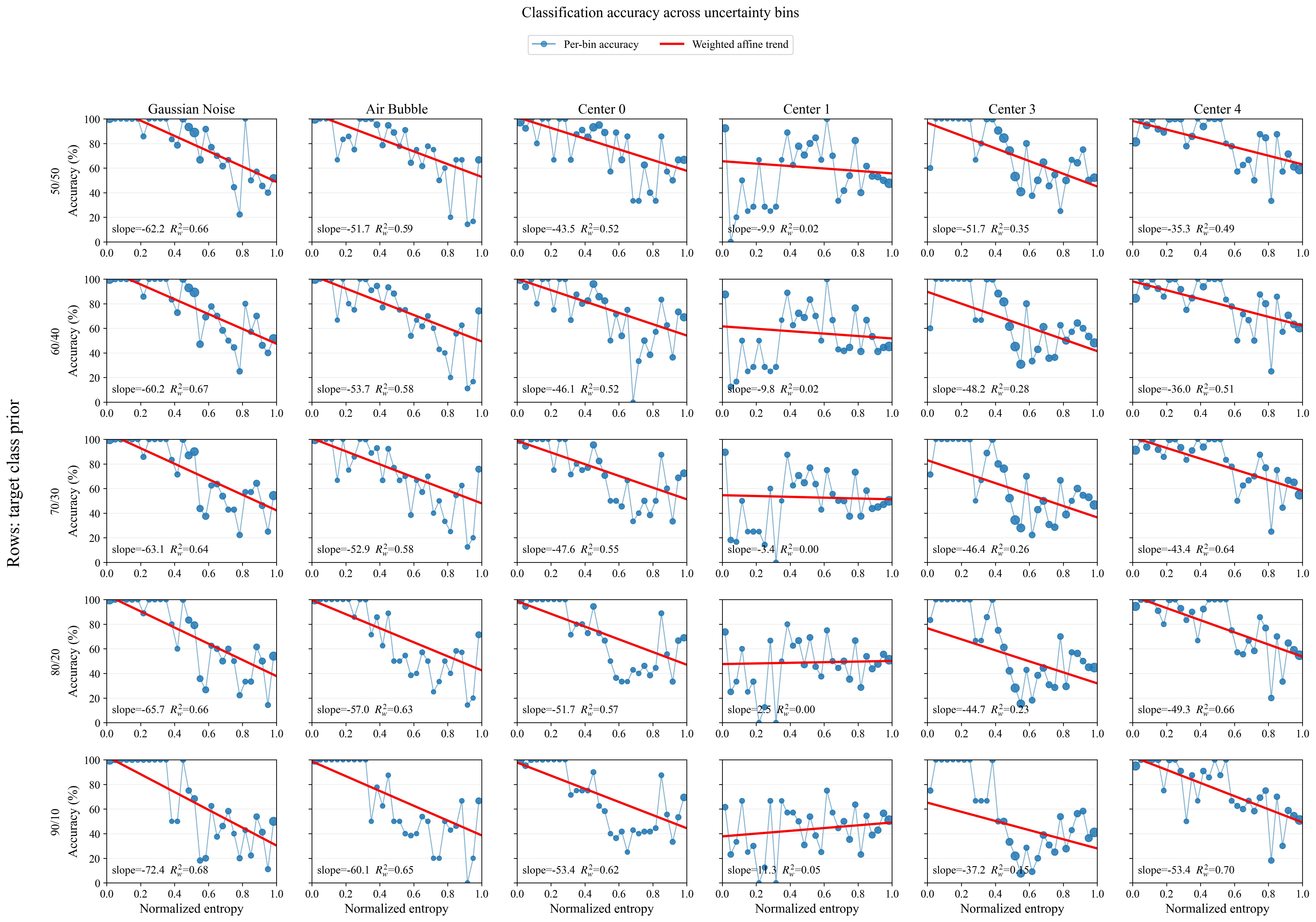}
    \caption{
    Relationship between prediction entropy and classification accuracy under
    \camseventeen inter-center domain shifts using center~2 as the source.
    Results are reported for \pixelcam.
    }
    \label{fig:entropy_accuracy_pixelcam}
\end{figure*}
\begin{figure*}[!t]
    \centering
    \includegraphics[
        width=0.99\linewidth
    ]{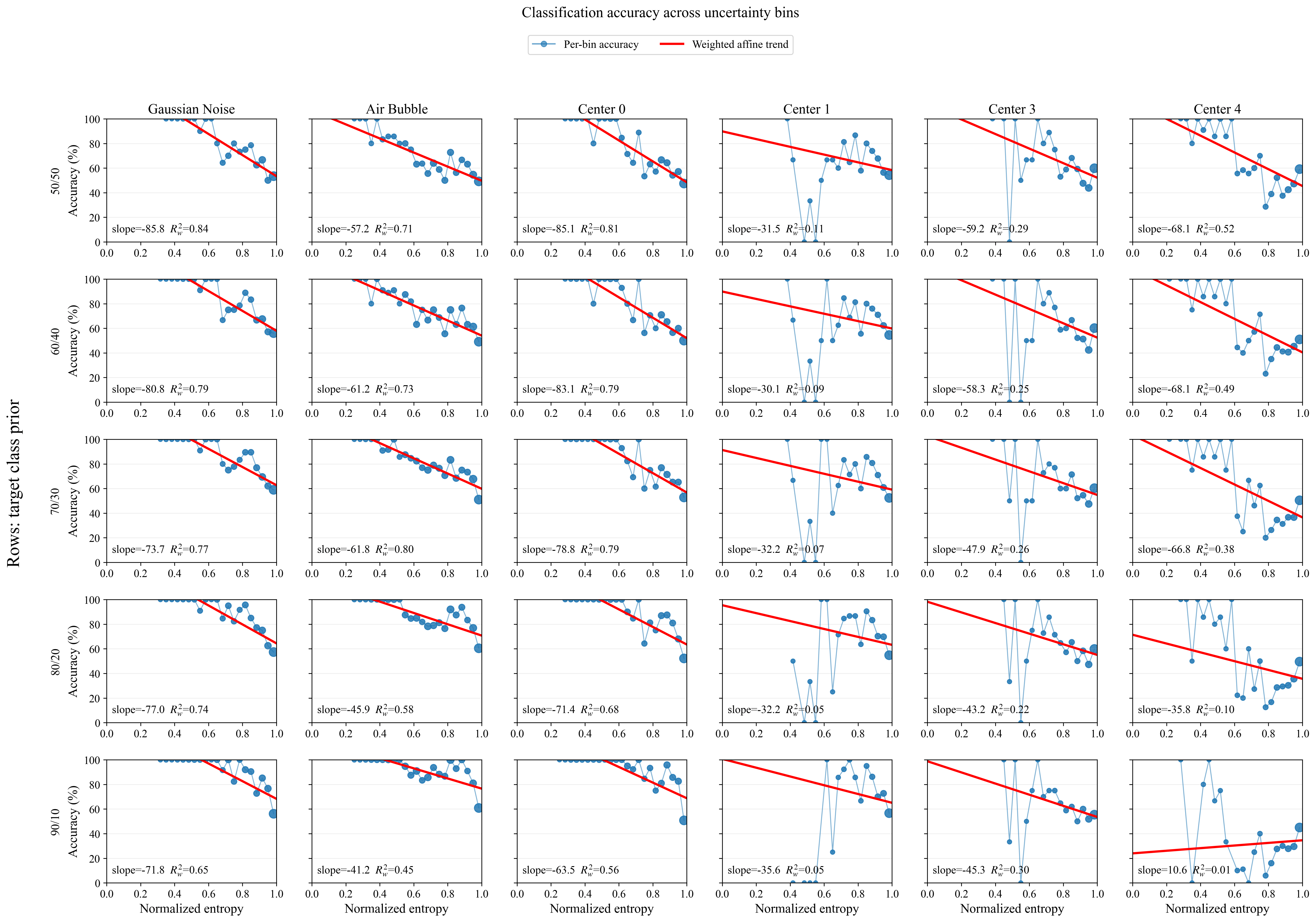}
    \caption{
    Relationship between prediction entropy and classification accuracy under
    \camseventeen inter-center domain shifts using center~2 as the source.
    Results are reported for \sat.
    }
    \label{fig:entropy_accuracy_sat}
\end{figure*}


\subsection{Additional Visualizations}
\label{app:Visualizations}
\begin{figure*}[!t]
    \centering
    \includegraphics[width=0.99\linewidth,trim={0 0 0 0},clip]
    {pxcam_vizu_compressed.png}
    \caption{Localization produced by the \pixelcam WSOL model after \sfdadep and the state-of-the-art SFDA methods (\glas $\rightarrow$ \camseventeenthree).}
    \label{fig:cam-example-pxcam}
\end{figure*}

\begin{figure*}[!t]
    \centering
    \includegraphics[width=0.99\linewidth,trim={0 0 0 0},clip]
    {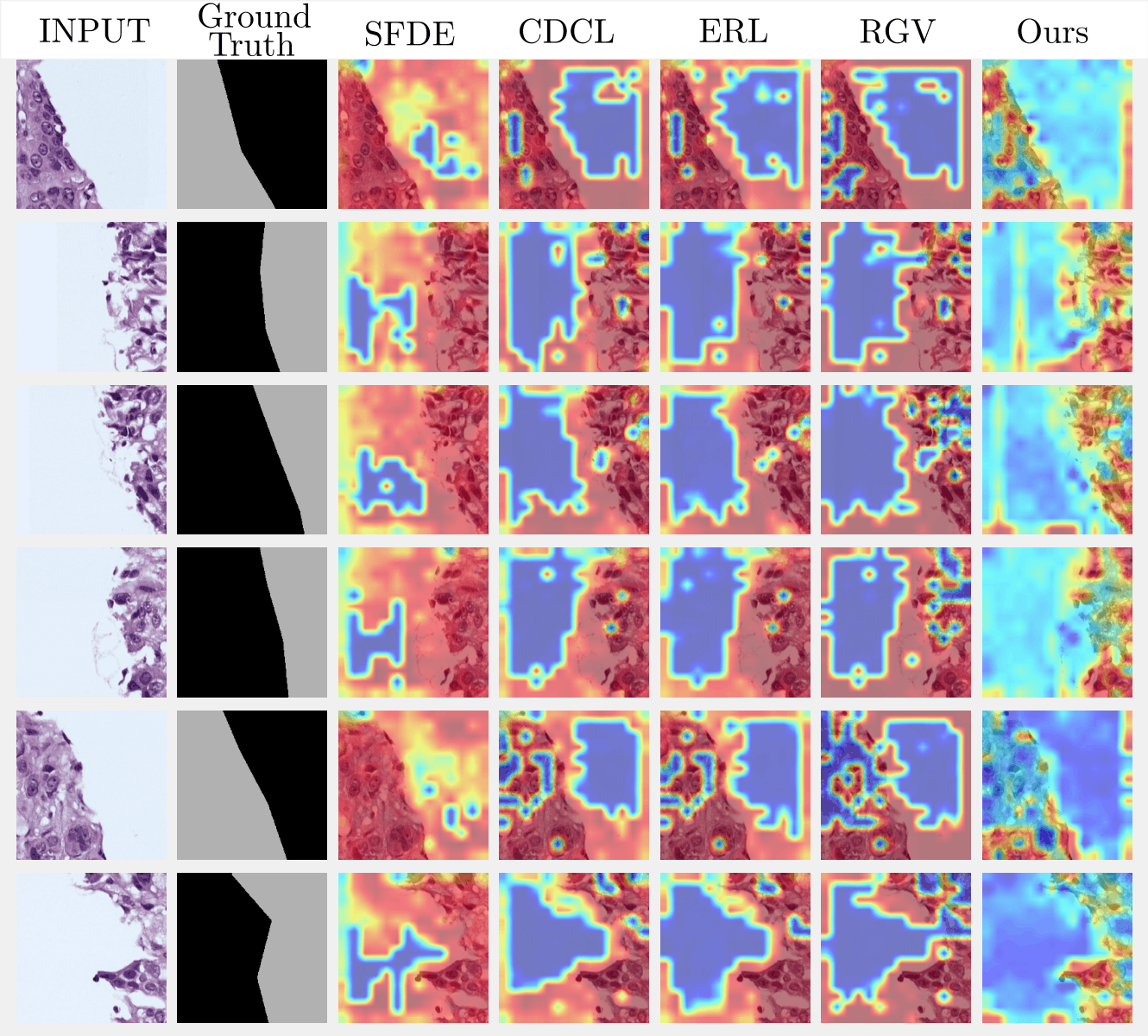}
    \caption{Localization produced by the \sat WSOL model after \sfdadep and the state-of-the-art SFDA methods (\glas $\rightarrow$ \camseventeenthree).}
    \label{fig:cam-example-sat}
\end{figure*}

\begin{figure*}[!t]
    \centering
    \includegraphics[width=0.99\linewidth,trim={0 0 0 0},clip]
    {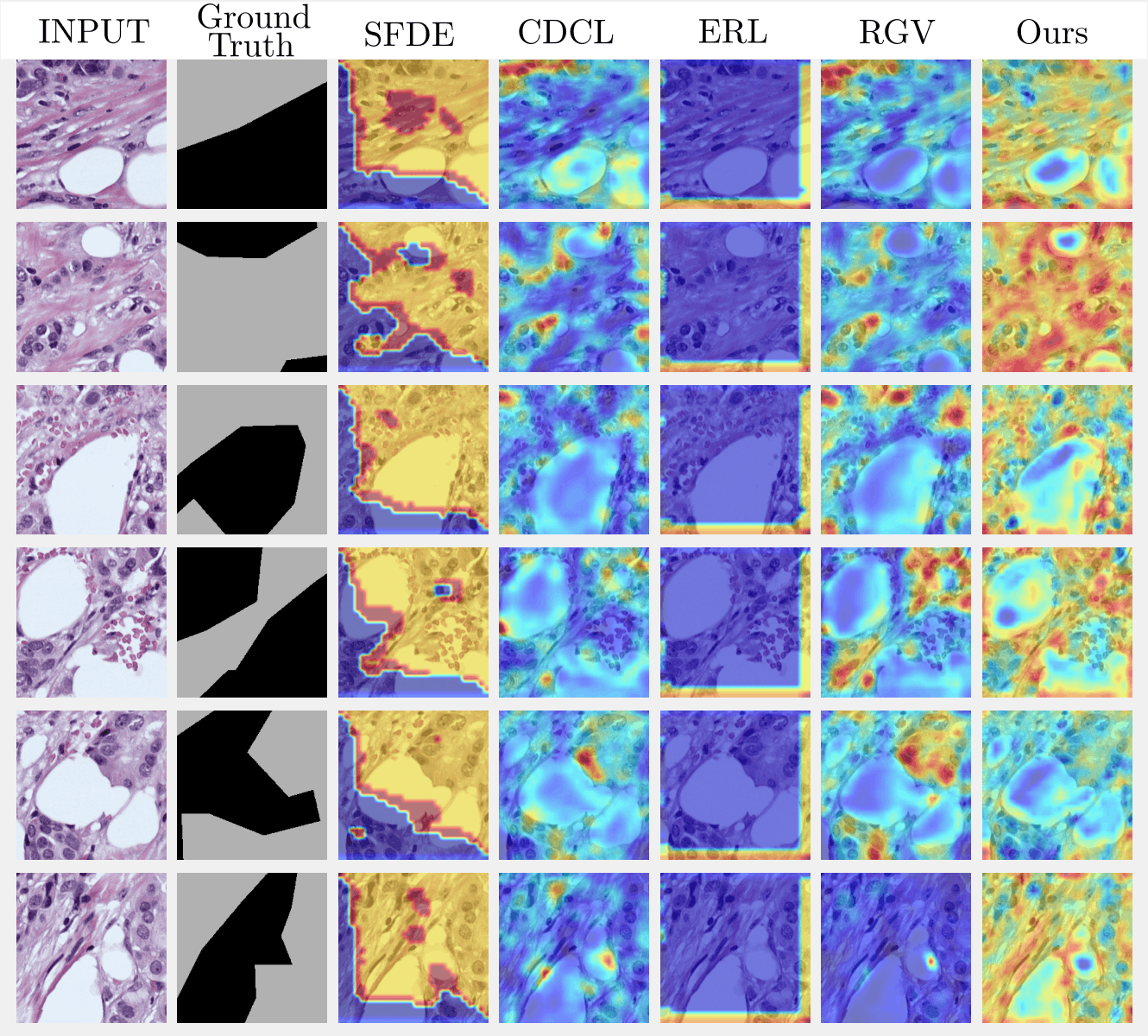}
    \caption{Localization produced by the \deepmil WSOL model after \sfdadep and the state-of-the-art SFDA methods (\glas $\rightarrow$ \camseventeenthree).}
    \label{fig:cam-example-deepmil}
\end{figure*}

\end{document}